\documentclass[letterpaper]{article} 
\usepackage{arxiv} 

\usepackage[hyphens]{url} 
\usepackage{graphicx} 
\usepackage{natbib} 
\usepackage{caption} 
\usepackage{algorithm}
\usepackage{algorithmic}

\usepackage{amsmath}
\usepackage{amssymb}
\usepackage{amsthm}
\usepackage{mathtools}
\usepackage{booktabs}

\newtheorem{theorem}{Theorem}
\newtheorem{lemma}[theorem]{Lemma}

\newtheorem{corollary}[theorem]{Corollary}
\newtheorem{definition}[theorem]{Definition}
\newtheorem{remark}[theorem]{Remark}
\newtheorem{assumption}[theorem]{Assumption}

\DeclareMathOperator*{\argmax}{arg\,max}

\DeclareMathOperator{\BetaCDF}{BetaCDF}

\newif\ifrestates
\restatesfalse
\newcommand{\restatable}[1]{%
  \ifrestates
    #1%
  \fi
}

\usepackage{hyperref}

\title{No-Regret Bayesian Optimization with Finite-Library Input-Warped Kernels\thanks{Code for experiments is available: \url{https://github.com/edvin-ketabati/bogp-paper-experiments}.}}

\author{%
  Edvin Ketabati Augustinsson\corresponding,  
  Robert A. Bridges
}
\affiliations{
\texttt{edvket@gmail.com}, \texttt{robert.bridges@ai.se}\\
AI Sweden, Gothenburg, Sweden, 
}

\begin{document}

\maketitle
\begin{abstract}
Gaussian-process Bayesian optimization (GP-BO) excels at black-box optimization of costly functions, e.g., hyperparameter optimization (HPO) and multi-agent system (MAS) design. Convergence-rate guarantees exist for select methods, notably GP upper confidence bound (GP-UCB), but require a fixed kernel. Critically, the kernel encodes how input proximity affects objective value similarity. When raw coordinates poorly match this geometry---as with log-scaled hyperparameters or localized peaks---input warping can greatly improve sample efficiency, yet known GP-UCB proofs require a fixed kernel. We propose Finite-Library Input-Warped Bayesian Optimization (FLIWBO), which selects warps from a finite library of smooth input maps by any history-dependent rule. It adapts the input geometry to accelerate learning while retaining high-probability convergence guarantees under mild hypotheses, with an explicit \(\sqrt{N_\varepsilon}\) library-size cost. Controlled diagnostics show that finite-library warping repairs planted geometry mismatches and identify FLIWBO failure cases. Across four repeated benchmarks—warped synthetic objectives, a confidence-fence trap, and Fashion-MNIST HPO—FLIWBO-UCB beats raw-coordinate GP-UCB under misspecified geometry, escapes traps that defeat even oracle-warp expected improvement, and recovers much of the gain from manual log scaling, while leading the tested methods that admit a matching regret guarantee. A 20-dimensional MAS design study further shows feasibility under costly noisy evaluations.
Code for experiments is available: \url{https://github.com/edvin-ketabati/bogp-paper-experiments}.

\end{abstract}

\section{Introduction}
\label{sec:introduction}

Many advances in science and technology depend on optimization problems in which only a limited number of candidate designs or configurations can be evaluated.
In many such problems, performance is revealed only by running a costly experiment, simulation, or training procedure, leaving little room for exhaustive search.
Bayesian optimization (BO) is well suited to this black-box setting because it uses a probabilistic surrogate to direct evaluations toward promising or informative inputs.
When evaluations are expensive, however, strong empirical performance is not enough: we also want mathematical assurance that the cost of exploration does not accumulate indefinitely.
Gaussian process upper confidence bound (GP-UCB) is a seminal method that balances predicted performance against uncertainty; under mild hypotheses it enjoys sublinear cumulative regret, so its average loss relative to the optimum vanishes over time \citep{srinivas2010gaussian,chowdhury2017kernelized}.

\begin{figure*}[t]
    \centering
    \begin{minipage}[t]{0.228\textwidth}
        \centering
        \includegraphics[width=\linewidth]{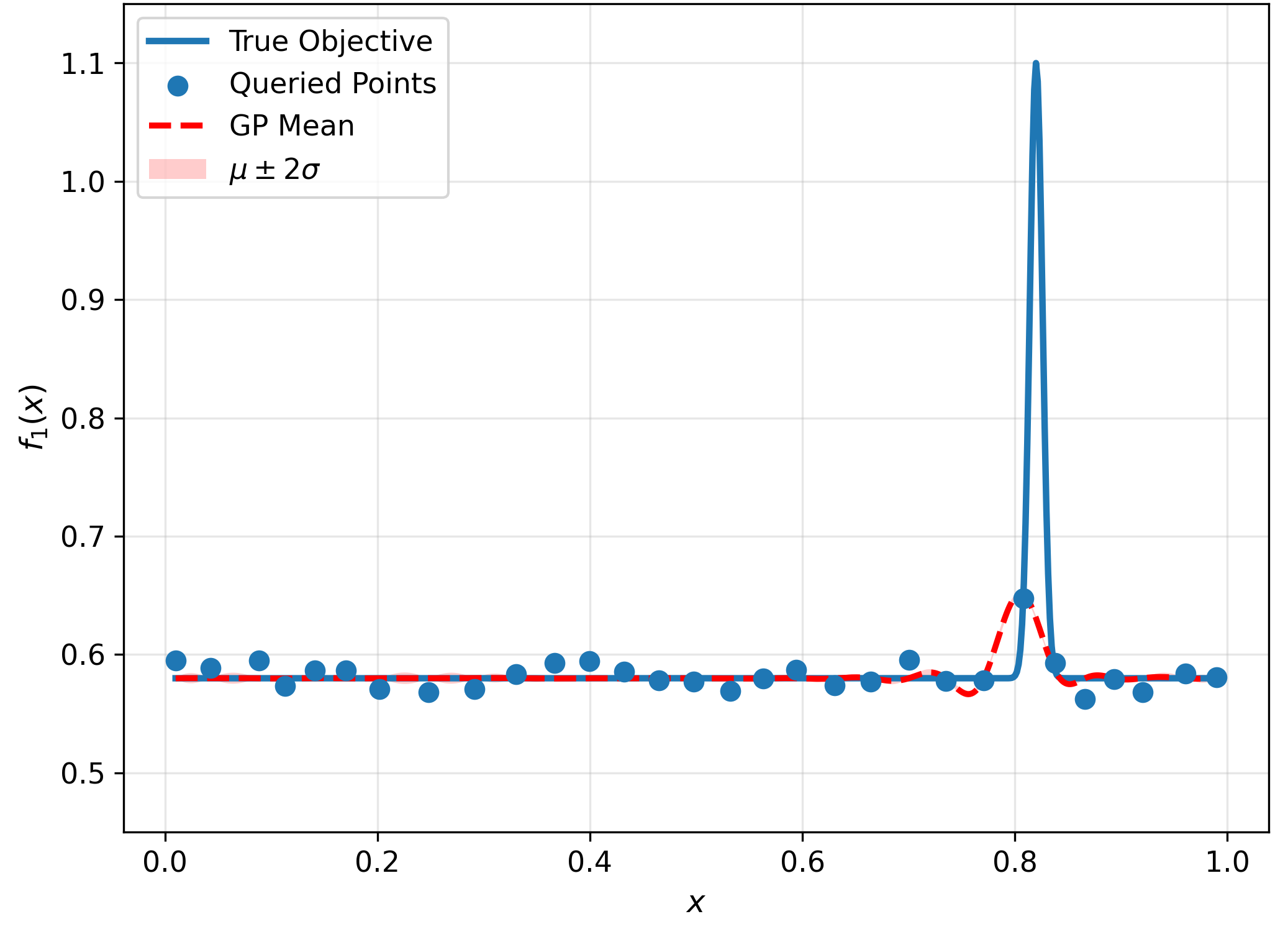}\\[-0.4ex]
        {\scriptsize\textbf{(a) Failure:} stationary GP in $x$}
    \end{minipage}\hfill
    \begin{minipage}[t]{0.228\textwidth}
        \centering
        \includegraphics[width=\linewidth]{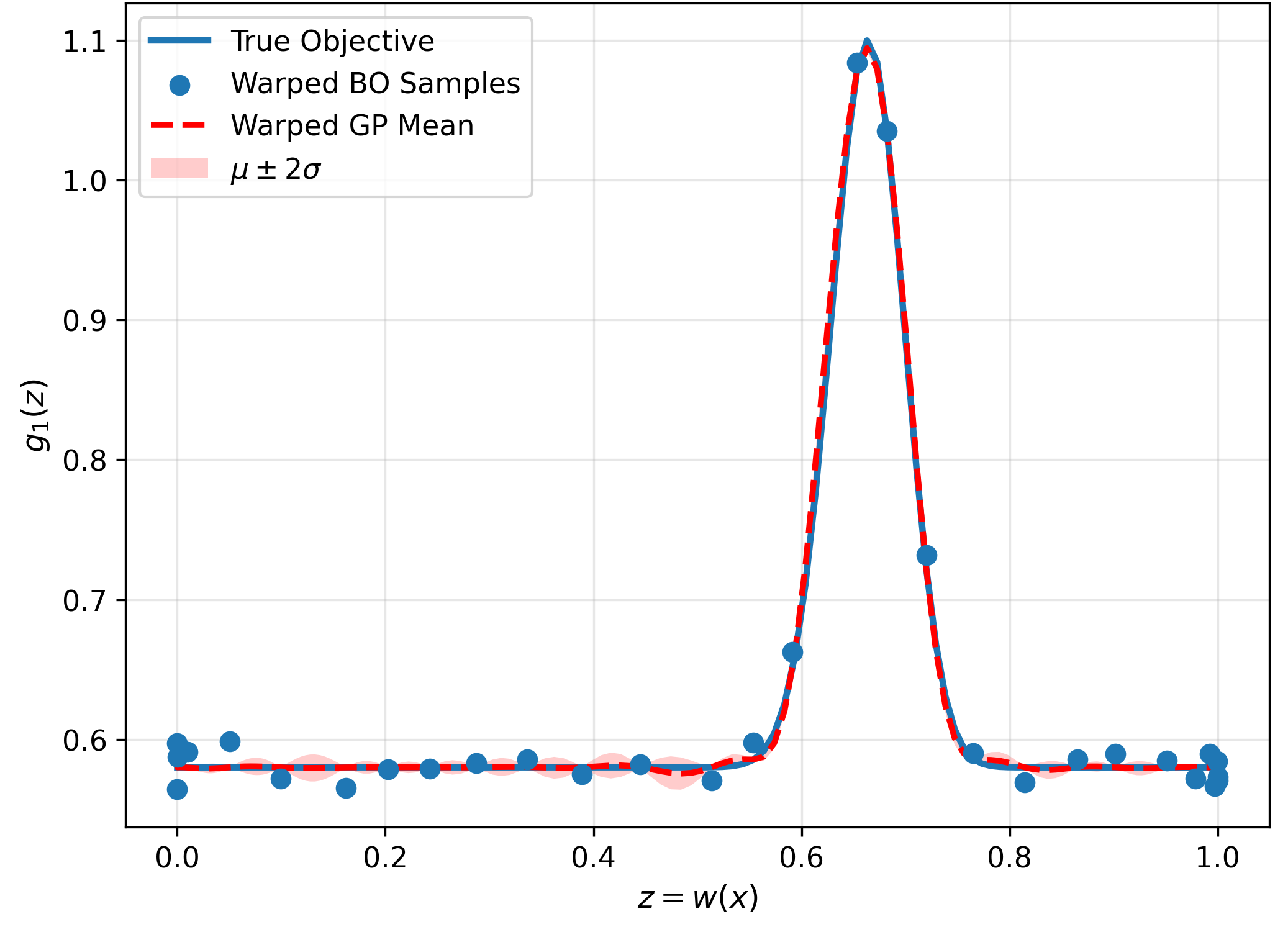}\\[-0.4ex]
        {\scriptsize\textbf{(b) Warp:} fit in $z=w(x)$}
    \end{minipage}\hfill
    \begin{minipage}[t]{0.228\textwidth}
        \centering
        \includegraphics[width=\linewidth]{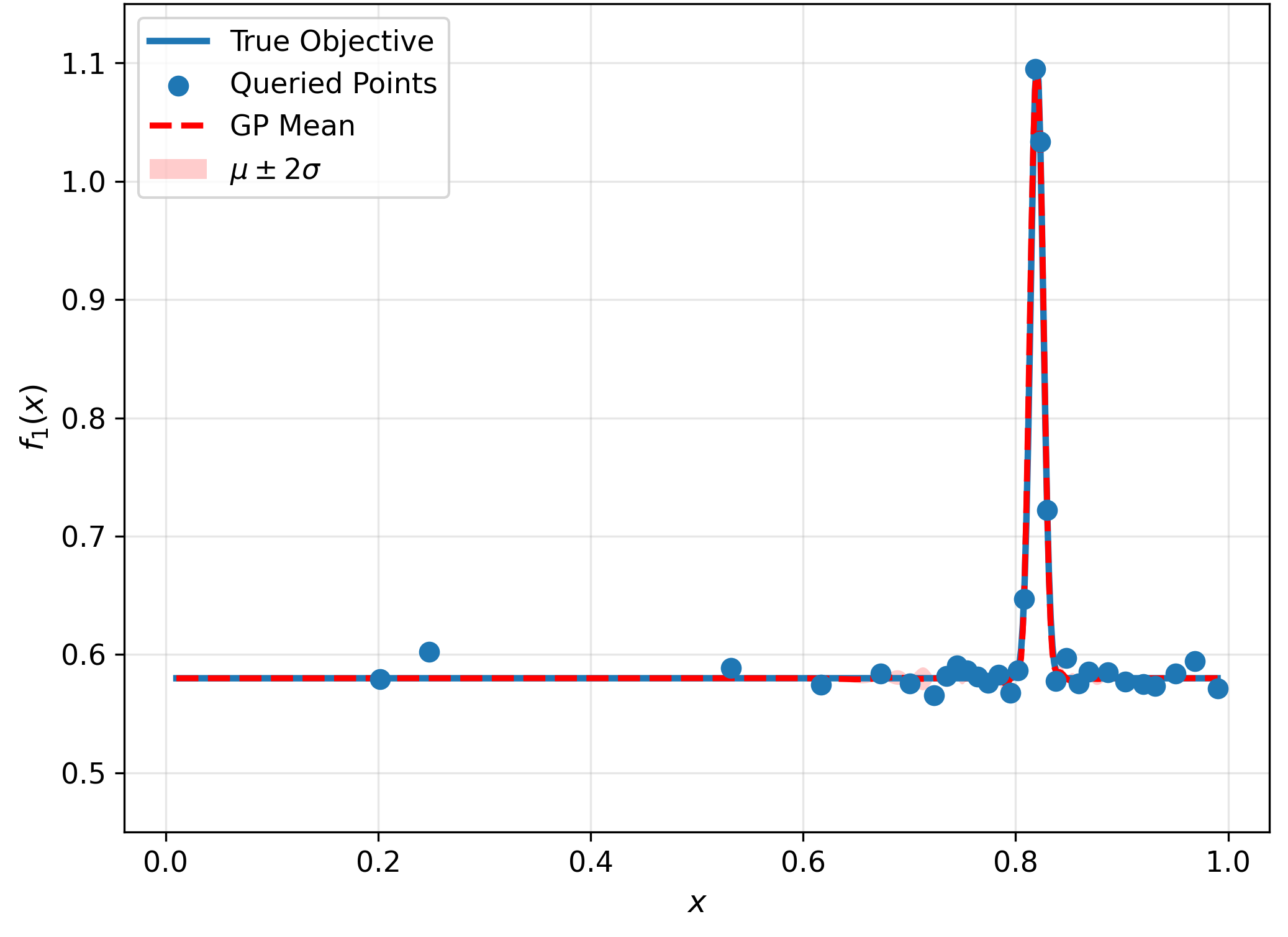}\\[-0.4ex]
        {\scriptsize\textbf{(c) Result:} warped fit back in $x$}
    \end{minipage}\hfill
    \begin{minipage}[t]{0.278\textwidth}
        \centering
        \includegraphics[width=\linewidth]{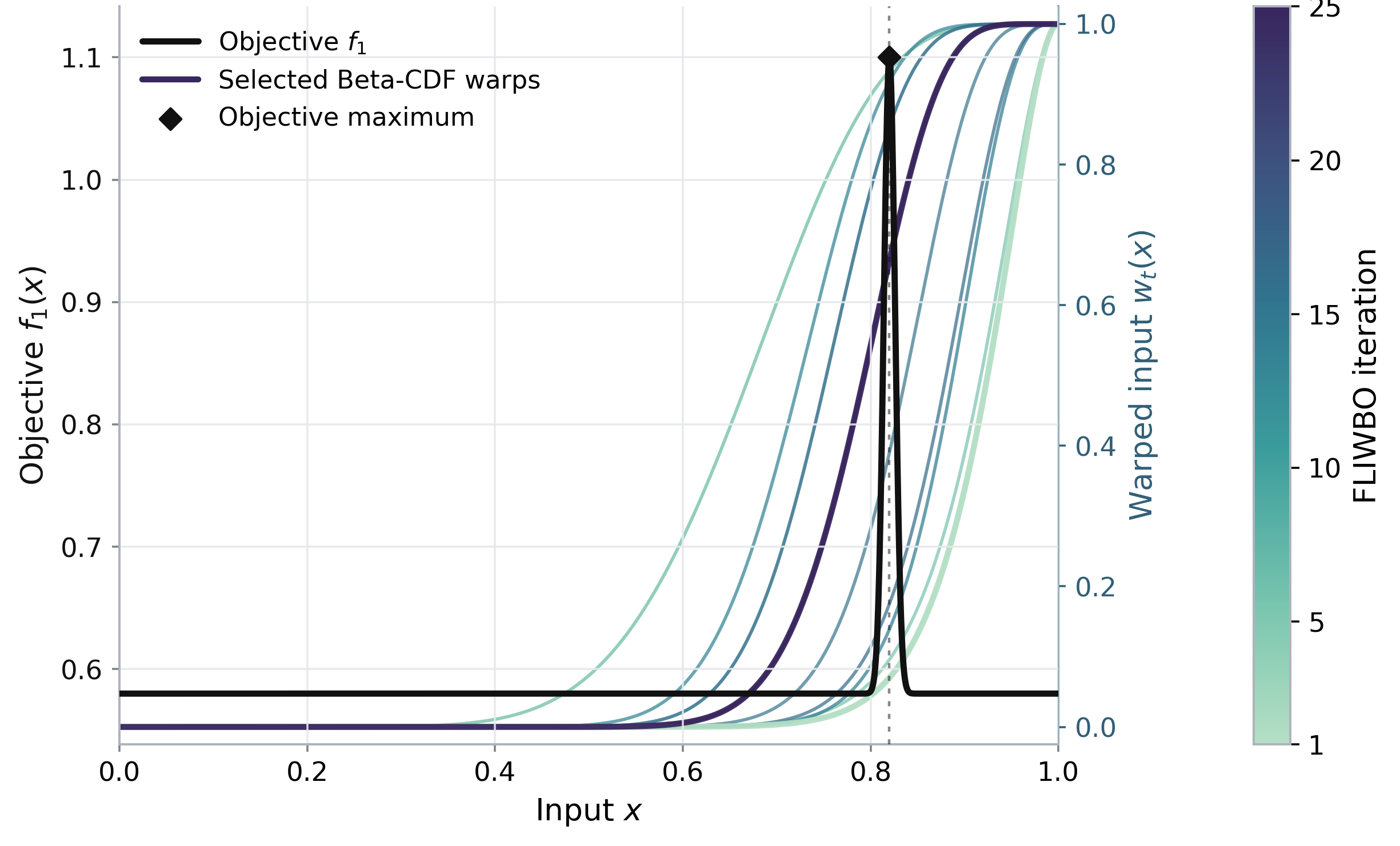}\\[-0.4ex]
        {\scriptsize\textbf{(d) Contribution:} warp learned online}
    \end{minipage}
    \caption{
    All panels use the same one-dimensional P1 objective function and BO budget, as discussed in Section \ref{sec:experiments}.
    \emph{(a) Problem:} a standard fixed-kernel Mat\'ern GP in the original coordinate treats equal $x$-distances as equally informative; it spreads queries over the flat background and oversmooths the narrow maximum.
    \emph{(b) Intuitive fix:} the selected monotone Beta cumulative distribution function (CDF) warp $z=w(x)$ expands the maximum's neighborhood, so the same base kernel instead sees a broad, smooth feature and produces a better fit.
    \emph{(c) Result:} mapping that posterior back to the original $x$-space gives a sharp, well-localized fit; the warp changes input geometry, not the objective.
    \emph{(d) Our contribution:} FLIWBO learns the input geometry from the optimization data rather than requiring the user to specify an oracle warp.
    From light to dark, the selected warps (Beta CDF) adapt toward a representation that expands the narrow high-value region, showing that the algorithm can identify a suitable warp on its own as observations accumulate.
    In (a)--(c), the blue curve is the objective, circles are queried points, the dashed red curve is the GP mean, and the band is $\mu\pm2\sigma$; in (d), the diamond marks the maximizer and color denotes the FLIWBO iteration.}
    \label{fig:stationary-vs-warped}
\end{figure*}

Those guarantees require a fixed kernel, which encodes how input proximity affects objective-value similarity.
Unfortunately, the coordinates used to describe a problem often poorly match how its objective varies.
In hyperparameter optimization (HPO), for example, learning rates are typically handled on a logarithmic scale, whereas other parameters remain linear.
Figure~\ref{fig:stationary-vs-warped} gives a one-dimensional illustration: raw-coordinate geometry can hide the variation that matters, a warp of the input space can expose it, and the resulting optimization concentrates queries on the consequential region.
Input warping can therefore accelerate GP-BO \citep{snoek2014input}, but ideally without sacrificing theoretical guarantees.

Learning such a warp from the observed data makes the kernel history-dependent obstructing standard GP-UCB proofs. 
Existing theory for unknown stationary hyperparameters, misspecification, and finite model selection does not cover this form of data-adaptive nonlinear geometry \citep{wang2014unknown,berkenkamp2019unknown,bogunovic2021misspecified,cutkosky2021dynamic,liu2023adaptation}.

We provide three contributions that reconcile this tension.

\textbf{Method.}
We propose Finite-Library Input-Warped Bayesian Optimization (FLIWBO) \citep{ketabatiAugustinsson2026warping}. The method keeps a finite library of smooth input warps, equiv., candidate GPs, that share one observation history, and admits any (esp., history-dependent) rule to select the active warp each round.
The selected warp determines the acquisition for the next query, adapting input geometry online to accelerate learning.

\textbf{Theory.}
To our knowledge, FLIWBO yields the first high-probability sublinear cumulative-regret guarantee for BO that learns nonlinear input transformations from its own history.
Under common hypotheses (objective function in RHKS \citet{srinivas2010gaussian},  uniform warp regularity \citet{snoek2014input}, and standard Mat\'ern-type Sobolev--RKHS equivalence \citep{kanagawa2018gaussian})---we prove RKHS and information-gain bounds and simultaneous GP-UCB confidence for every branch.
The guarantee holds for any history-dependent selector; library size enters logarithmically in confidence and through a $\sqrt{N_\varepsilon}$ regret factor.

\textbf{Evidence.}
Controlled diagnostics show that finite-library warping repairs planted geometry mismatches and identify failure cases where fixed geometry is preferable.
Across four repeated benchmarks---warped synthetic objectives, a confidence-fence trap, and Fashion-MNIST HPO---FLIWBO-UCB beats raw-coordinate GP-UCB under misspecified geometry, escapes traps that defeat even oracle-warp expected improvement, and recovers much of the gain from manual log scaling, while leading the tested methods that admit a matching frequentist regret guarantee.
A collaborative feasibility study further applies FLIWBO to a 20-dimensional MAS design problem with costly noisy evaluations and compares selected designs to a human engineer's reference. Code, configurations, and data are provided in the supplementary Code and Data Package.

\section{Related Work}
\label{sec:related-work}

Our compatible finite warped-kernel family provides uniform regularity, common information-gain control, and simultaneous GP-UCB confidence under data-adaptive nonlinear geometry.
We know of no comparable cumulative-regret analysis.

\paragraph{Guaranteed GP optimization under adaptation.}
Fixed-kernel GP-UCB bounds cumulative regret via frequentist confidence and maximum information gain \citep{srinivas2010gaussian,chowdhury2017kernelized}.
Unknown-hyperparameter guarantees prescribe expected improvement (EI) estimation or expand the stationary function class \citep{wang2014unknown,berkenkamp2019unknown}; neither adapts nonlinear geometry.
Misspecified GP optimization pays to approximate the objective by a bounded-RKHS function \citep{bogunovic2021misspecified}; we instead assume library-wide realizability, excluding off-library misspecification.

\paragraph{Finite model families.}
Model-selection schemes eliminate candidate algorithms or priors inconsistent with regret bounds \citep{cutkosky2021dynamic}; GP variants eliminate finite priors, sometimes with hyperpriors \citep{ziomek2025unknown,sandberg2025prior}.
RKHS-class adaptation identifies a valid class at a model-selection cost, including under unknown kernel regularity \citep{liu2023adaptation}.
FLIWBO instead exploits compatibility: smooth maps of one base kernel place the objective in every candidate RKHS with uniform norm and information-gain bounds.
Confidence thus precedes arbitrary history-dependent selection; no true or best warp need be identified.

\paragraph{Learning input geometry.}
Beyond length scales, metric and deep-kernel models learn richer representations \citep{Snoek2012,titsias2010variational,kiyohara2025dkl}.
\citet{snoek2014input} most directly infer continuous coordinate-wise Beta-CDF warps.
FLIWBO discretizes their construction and couples it with GP-UCB, trading library size and computation for simultaneous frequentist confidence and finite-time cumulative-regret control under adaptive selection.

\section{FLIWBO Method \& Convergence}
\label{sec:main-result}

We consider a nonempty compact search domain $X\subseteq\mathcal{X}\triangleq[0,1]^D$ and an unknown continuous objective $f:X\to\mathbb{R}$.
At each round $t$, the algorithm chooses $x_t\in X$ as a function of
$\mathcal{D}_{t-1}$ and observes $y_t=f(x_t)+\epsilon_t$, where
$\mathcal{D}_t\triangleq\{(x_s,y_s)\}_{s=1}^t$ and $\mathcal{D}_0\triangleq\emptyset$.
Following \citet{srinivas2010gaussian}, we assume
$\mathbb{E}[\epsilon_t\mid\mathcal{D}_{t-1}]=0$ and
$|\epsilon_t|\leq\sigma$ almost surely (a.s.).
For a base kernel $k$ (later denoted $k_0$) defined on $\mathcal{X}$, we assume $k(x,x)\leq 1$ and
write $K_{t-1}=[k(x_s,x_{s'})]_{s,s'<t}$,
$k_{t-1}(x)=(k(x,x_s))_{s<t}$, and $y_{t-1}=(y_1,\ldots,y_{t-1})$.
The GP surrogate at round $t$ is the usual regularized posterior:
$\mu_{t-1}(x)=k_{t-1}(x)^\top(K_{t-1}+\sigma^2 I)^{-1}y_{t-1}$ and
$\sigma_{t-1}^2(x)=k(x,x)-k_{t-1}(x)^\top(K_{t-1}+\sigma^2 I)^{-1}k_{t-1}(x)$.
The next query maximizes an acquisition function $U_t(x)$ built from
$(\mu_{t-1},\sigma_{t-1})$, e.g.\ UCB or EI.

Let $x^\star\in\arg\max_{x\in X}f(x)$, and define instantaneous regret
$r_t\triangleq f(x^\star)-f(x_t)$ and cumulative regret
$R_T\triangleq\sum_{t=1}^T r_t.$
Sublinear cumulative regret means $R_T=o(T)$, so average regret $R_T/T\to 0$.

\subsubsection{Input warps and finite library}
\label{sec:main-method}
To change the geometry seen by the GP, we use a parameterized input
warp $w_\theta:X\to E_\theta\subseteq\mathcal{X}$, $\theta\in\Theta$.
A warp that expands the domain near a maximizer $x^\star$ makes this high-value region occupy a larger region in warped coordinates, generally accelerating GP-BO. 
A warp that contracts the domain near the maximizer may make the high-value region harder to locate.

We assume there is an integer $m\ge 1$ (matched to the smoothness of
the base kernel) such that each $w_\theta$ is a $C^m$ diffeomorphism
onto its compact image $E_\theta$, with derivatives of $w_\theta$ and
$w_\theta^{-1}$ up to order $m$ uniformly bounded over $\theta\in\Theta$
(formalized below as Assumption~\ref{ass:regular_warp}).
In practice, we follow \citet{snoek2014input}, using coordinate-wise
$w_\theta(x)=(w_{\theta,1}(x_1),\ldots,w_{\theta,D}(x_D))$
with each $w_{\theta,j}$ a Beta CDF on a truncated box $X=[\tau,1-\tau]^D$ (hence $C^\infty$ and strictly
increasing).
Yet (unlike Snoek et al.), we consider a finite library of warp parameters
$\Theta_\varepsilon
    \triangleq
    \{\theta^{(1)},\ldots,\theta^{(N_\varepsilon)}\}
    \subseteq \Theta$; hence the derivative bounds reduce to
$\tfrac1M\le w'_{\theta,j}\le M$.

For a fixed warp $w_\theta$, the warped objective  is
    $g_\theta(z)\triangleq f(w_\theta^{-1}(z)):E_\theta\to\mathbb{R}$,
so
$f(x)=g_\theta(w_\theta(x))$.
From base kernel $k_0$, the warp-induced kernel is
$k_\theta(x,x')\triangleq k_0(w_\theta(x),w_\theta(x')).$
For $\theta^{(i)}\in\Theta_\varepsilon$, we write $k_i\triangleq k_{\theta^{(i)}}$ and GP posterior
$\mu_{i,t-1}$, $\sigma^2_{i,t-1}$  under $k_i$ and common data $\mathcal{D}_{t-1}$.

Algorithm~\ref{alg:main-fliwbo} presents FLIWBO with UCB acquisition.
Under mild assumptions we establish simultaneous confidence for every library branch, so any history-dependent warp selector inherits a sublinear cumulative-regret guarantee.
While other acquisition functions may be used in FLIWBO, our sublinear regret proof is for UCB only.
Practical selectors used in our experiments are specified in Section~\ref{sec:experiments}.

\begin{algorithm}[t]
\caption{FLIWBO with GP-UCB}
\label{alg:main-fliwbo}
\begin{algorithmic}[1]
\REQUIRE $X$, $k_0$, warps $\{w_i\}_{i=1}^{N_\varepsilon}$, selector $s$, confidence sequence $(\beta_t)$
\STATE Define $k_i(x,x')=k_0(w_i(x),w_i(x'))$ \ $\forall \ i$
\STATE Initialize the common dataset $\mathcal{D}_0$
\FOR{$t=1,\ldots,T$}
    \STATE Let $(\mu_{i,t-1},\sigma_{i,t-1})$ be the posterior under $k_i,\mathcal{D}_{t-1}$ for each $i$
    \STATE $j_t\gets s(\mathcal{D}_{t-1})$
    \STATE $x_t\in\argmax_{x\in X}\mu_{j_t,t-1}(x)+\sqrt{\beta_t}\,\sigma_{j_t,t-1}(x)$
    \STATE Observe $y_t=f(x_t)+\epsilon_t$ and update $\mathcal{D}_t\gets\mathcal{D}_{t-1}\cup\{(x_t,y_t)\}$
\ENDFOR
\end{algorithmic}
\end{algorithm}

\subsection{Library-wide RKHS bounds and confidence}
\label{sec:main-guarantees}
Proofs of all results are in the supplement.
We write $\mathcal{H}_{k_\theta}$ for the RKHS of kernel $k_\theta$.
Fix a witness $\theta_0\in\Theta$ with
$f=g_0\circ w_{\theta_0}$ (e.g., $w_{\theta_0}(x)=x$).

\begin{assumption}[Witness RKHS bound]
\label{ass:rkhs}
$g_0\triangleq f\circ w_{\theta_0}^{-1}\in\mathcal{H}_{k_0}$ with $\|g_0\|_{\mathcal{H}_{k_0}}\le B_g<\infty$.
\end{assumption}

\begin{lemma}[Warp preserves RKHS boundedness]
\label{lem:rkhs_warp}
For any $\theta\in\Theta$ and $g\in\mathcal{H}_{k_0}$, the pullback $f\triangleq g\circ w_\theta$ lies in $\mathcal{H}_{k_\theta}$ with $\|f\|_{\mathcal{H}_{k_\theta}}\le\|g\|_{\mathcal{H}_{k_0}}$.
\end{lemma}
It follows that
$\|f\|_{\mathcal{H}_{k_{\theta_0}}}\le B_g$.

Let $m\ge 1$ be the integer Sobolev order of the base kernel $k_0$, e.g.\ Mat\'ern with smoothness $\nu$ when $m=\nu+D/2$ is an integer,
and let $S^m(\Omega)$ denote the Sobolev space of functions on $\Omega$ whose weak derivatives up to order $m$ lie in $L^2(\Omega)$.
\begin{assumption}[Sobolev--RKHS equivalence]
\label{ass:finite_net_base_rkhs}
On every domain $E\in\{E_{\theta_0}\}\cup\{E_\theta:\theta\in\Theta_\varepsilon\}$,
$\mathcal{H}_{k_0|_{E\times E}}=S^m(E)$ as sets, with equivalent norms:
$C_{\mathrm{eq}}^{-1}\|u\|_{S^m(E)}\le\|u\|_{\mathcal{H}_{k_0|_{E\times E}}}\le C_{\mathrm{eq}}\|u\|_{S^m(E)}$.
\end{assumption}
This holds, for example, for Mat\'ern kernels whenever $m=\nu+D/2$ is an integer; see \citet{kanagawa2018gaussian}.

\begin{assumption}[Uniform warp regularity]
\label{ass:regular_warp}
For every $\theta\in\Theta_\varepsilon$, $w_\theta$ is a $C^m$ diffeomorphism, and the $C^m$ norms of $w_\theta$ and $w_\theta^{-1}$ are uniformly bounded over $\theta\in\Theta_\varepsilon$.
\end{assumption}
Define the relative warp $\phi_i\triangleq w_{\theta_0}\circ w_{\theta^{(i)}}^{-1}:E_i\to E_0$.
By the chain rule, each $\phi_i$ is likewise a $C^m$ diffeomorphism with derivatives uniformly bounded in $i$.
\begin{lemma}[Uniform composition bound on $S^m$]
\label{lem:finite_library_composition}
There exists a constant $C_\mathrm{comp}$ such that for all $i\in\{1,\dots,N_\varepsilon\}$ and all $u\in S^m(E_0)$,
    $\|u\circ\phi_i\|_{S^m(E_i)}\le C_\mathrm{comp}\|u\|_{S^m(E_0)}$.
\end{lemma}
Proof sketch: each $\phi_i$ yields a finite composition constant $C_{\mathrm{comp},i}$; set $C_\mathrm{comp}=\max_i C_{\mathrm{comp},i}$.

We can now give a uniform RKHS bound on the library.
\begin{theorem}[Uniform RKHS transfer over the finite library]
\label{thm:finite_library_rkhs_transfer}
\label{thm:main-uniform-compatibility}
There exists a constant $C_{\mathrm{warp}}$ such that for every $i\in\{1,\dots,N_\varepsilon\}$,
$f\in\mathcal{H}_{k_{\theta^{(i)}}}$
and $\|f\|_{\mathcal{H}_{k_{\theta^{(i)}}}}\le C_{\mathrm{warp}}.$
\end{theorem}

\subsubsection{Information gain}
Recall $K_{T}=[k(x_s,x_{s'})]_{s,s'\le T}$ and
define the \textit{maximum information gain} as in \citet{srinivas2010gaussian},
    $\gamma_T(k,X) \triangleq  \sup_{x_1, \dots, x_T \in X} \tfrac{1}{2} \log \det \left( I + \sigma^{-2} K_T \right).$ 
Lemma~5.3 of \citet{srinivas2010gaussian}, gives a fixed-kernel variance bound and information-gain identity:
\begin{lemma}[Variance accumulation for a fixed kernel]
\label{lem:fixed-kernel-variance}
For $k(x,x)\leq1$, let $C_{\mathrm{info}}\triangleq2/\log(1+\sigma^{-2})$.
Then for any adaptively chosen query sequence $x_1,\dots,x_T$ and any fixed kernel $k$,
\[
    \sum_{t=1}^T \sigma_{t-1}^2(x_t) \le C_{\mathrm{info}}\gamma_T(k,X).
\]
\end{lemma}

\begin{lemma}[Maximum information gain under a warp]
\label{lem:gamma_warp_transfer}
For every $\theta \in \Theta$ and every $T \ge 1$,
\begin{equation}
    \gamma_T(k_\theta, X) = \gamma_T \left(k_0, w_\theta(X) \right) \le \gamma_T \left(k_0, \mathcal{X} \right).
\label{eq:main-uniform-ig}
\end{equation}
\end{lemma}

\subsubsection{Uniform confidence}
By Theorem~\ref{thm:finite_library_rkhs_transfer}, $\|f\|_{\mathcal{H}_{k_i}}\le C_{\mathrm{warp}}$ for every library kernel, so the GP-UCB confidence theorem of \citet{srinivas2010gaussian} applies to each fixed $k_i$ separately.
\begin{lemma}[Fixed-kernel GP-UCB confidence; \citet{srinivas2010gaussian}]
\label{lem:fixed-kernel-confidence}
Fix $\delta\in (0,1)$. If $\beta_t \geq 2 B^2 + 300\gamma_t(k,X) \log^3 \left( t / \delta \right)$,
then, with probability at least $1-\delta$,
    $|f(x) - \mu_{t-1}(x)| \le \sqrt{\beta_t}\sigma_{t-1}(x)$
simultaneously for all $t \ge 1$ and all $x \in X$.
\end{lemma}
Applied to branch $i$ with norm bound $C_{\mathrm{warp}}$ and failure probability $\delta/N_\varepsilon$, this yields a per-branch schedule $\beta_{i,t}$ and the UCB acquisition function
\begin{equation}
U_{i,t}(x)
=
\mu_{i,t-1}(x)+\sqrt{\beta_{i,t}}\,\sigma_{i,t-1}(x).
\label{eq:main-ucb}
\end{equation}
A union bound over the $N_\varepsilon$ branches, together with $\gamma_t(k_i,X)\le\Gamma_t:=\gamma_t(k_0,\mathcal{X})$, is controlled by the single shared schedule
\begin{equation}
    \beta_t = 2 C_{\mathrm{warp}}^2 + 300\Gamma_t \log^3 \left( {t N_\varepsilon}/{\delta} \right),
\label{eq:main-beta}
\end{equation}
which satisfies $\beta_t\ge\beta_{i,t}$ for every $i$.
We therefore take $\beta_{i,t}=\beta_t$ for all $i$ in~\eqref{eq:main-ucb}, and obtain simultaneous library-wide confidence:
\begin{lemma}[Uniform confidence over the finite library]
\label{lem:finite_library_confidence}
With $\beta_t$ as in~\eqref{eq:main-beta} and $\delta\in(0,1)$, with probability at least $1-\delta$,
    $|f(x)-\mu_{i,t-1}(x)| \le \sqrt{\beta_t}\,\sigma_{i,t-1}(x)$
 for all $i \in \{1, \dots, N_\varepsilon\}$, all $t \ge 1$, and all $x \in X$.
\end{lemma}

\subsection{FLIWBO-UCB policy and the telescope obstacle}
For the FLIWBO-UCB policy (Algorithm~\ref{alg:main-fliwbo}),
at round $t$ let $j_t\in\{1,\dots,N_\varepsilon\}$ be any history-dependent library index
and query
    $x_t\in\arg\max_{x\in X}U_{j_t,t}(x)$
as in~\eqref{eq:main-ucb} with the shared schedule~\eqref{eq:main-beta}.
Update the common history $\mathcal{D}_t$; posteriors used at round $t$ are computed from $\mathcal{D}_{t-1}$.
Any $\mathcal{D}_{t-1}$-measurable selector is admissible, including marginal likelihood
\citep{snoek2014input} or posterior model probability.
With prior weights $(\pi_i)_{i=1}^{N_\varepsilon}$, a MAP selector is
    $j_t\in\arg\max_{1\le i\le N_\varepsilon}\{\log p(\mathcal{D}_{t-1}\mid\theta^{(i)})+\log\pi_i\}$;
under a uniform prior this is maximum marginal likelihood.
Analogous policies with other acquisition functions are possible, but our regret theorem is for UCB.

\subsubsection{Why history-dependent kernels break fixed-kernel accounting}
\label{sec:main-telescope}
For a fixed kernel $k$, 
$\tfrac{\det(I+\sigma^{-2}K_t^k)}{\det(I+\sigma^{-2}K_{t-1}^k)}
=1+\sigma^{-2}\sigma_{k,t-1}^2(x_t).
$
Intermediate determinants cancel over $t$, yielding the information-gain representation of predictive variance
\citep[Lemma~5.3]{srinivas2010gaussian}.
Under FLIWBO the regret-relevant terms are $\sigma_{j_t,t-1}^2(x_t)$ with history-dependent $j_t$,
so consecutive ratios use different kernels and no longer telescope.
The obstacle is data-dependent switching of the uncertainty model, not adaptive sampling of $x_t$.
Our remedy is Lemma~\ref{lem:finite_library_confidence} (simultaneous confidence for every branch)
together with a branchwise variance bound below.

\subsection{FLIWBO-UCB cumulative regret}
The standard GP-UCB comparison of \citet{srinivas2010gaussian} applies to the selected branch
$k_{j_t}$:
\begin{lemma}[Selected-branch instantaneous regret and variance accumulation]
\label{lem:select-branch-inst-regret}
\label{lem:variance-reduction}
On the event of Lemma~\ref{lem:finite_library_confidence},
$r_t\le 2\sqrt{\beta_t}\,\sigma_{j_t,t-1}(x_t)$ for all~$t$.
If $(\beta_t)$ is nondecreasing, then
$R_T\le 2\sqrt{\beta_T\,T\sum_{t=1}^T\sigma_{j_t,t-1}^2(x_t)}$.
\end{lemma}
Bounding the selected variances by summing over all branches and applying
Lemma~\ref{lem:fixed-kernel-variance} 
gives:
\begin{lemma}[Selected-branch variance bound]
\label{lem:selected-branch-variance}
    $\sum_{t=1}^T \sigma_{j_t,t-1}^2(x_t) \le C_{\mathrm{info}} \sum_{i=1}^{N_\varepsilon} \gamma_T(k_i,X)$.
\end{lemma}

\begin{theorem}[FLIWBO-UCB sublinear cumulative regret]
\label{thm:library-no-regret}
\label{thm:main-no-regret}
Under Assumptions~\ref{ass:rkhs}, \ref{ass:finite_net_base_rkhs}, and \ref{ass:regular_warp}, the FLIWBO-UCB policy with $\beta_t$ as in Lemma~\ref{lem:finite_library_confidence},  has 
\begin{equation}
    R_T \le 2 \sqrt{C_{\mathrm{info}} N_\varepsilon T \beta_T \Gamma_T}
\label{eq:main-regret-bound}
\end{equation}
probability at least $1-\delta$,
where
    $\Gamma_T \triangleq \gamma_T(k_0, \mathcal{X}).$
In particular, if $\beta_T \Gamma_T = o(T)$, then $R_T = o(T)$ and hence $R_T/T\to 0$.
E.g., Mat\'ern kernels satisfy this growth condition when $\nu>D/2$ \citep{vakili2021information}.
\end{theorem}
On the library-wide confidence event, standard GP-UCB controls regret by the selected-branch variances; summing those over branches and transferring information gain through the warp yields an $N_\varepsilon\Gamma_T$ factor, so $R_T=O(\sqrt{T\beta_T N_\varepsilon\Gamma_T})$, which is $o(T)$ whenever $\beta_T\Gamma_T=o(T)$.

\begin{remark}[Selector-adaptive variance refinement]
The bound in Theorem~\ref{thm:library-no-regret} charges every branch for all $T$ rounds.
A tighter bound uses only selected rounds: with
$J_i\triangleq\{t:j_t=i\}$ and $n_i(T)\triangleq|J_i|$,
$\sum_{t=1}^T\sigma_{j_t,t-1}^2(x_t)
\leq
C_{\mathrm{info}}\sum_{i=1}^{N_\varepsilon}\gamma_{n_i(T)}(k_i,X),$ 
hence
$R_T\le 2\sqrt{C_{\mathrm{info}}T\beta_T\sum_i\gamma_{n_i(T)}(k_i,X)}$.
This recovers the $N_\varepsilon\Gamma_T$ bound since $n_i(T)\le T$; if the selector concentrates on few branches it can be much smaller, but the worst-case rate is unchanged.
\end{remark}

\paragraph{What the guarantee costs.}
The library affects the bound in two ways.
Through the union bound it appears only inside the $\log^3(tN_\varepsilon/\delta)$ term of~$\beta_t$ in~\eqref{eq:main-beta}.
Through branchwise variance accumulation it contributes an explicit $\sqrt{N_\varepsilon}$ factor in~\eqref{eq:main-regret-bound}.
There is also a computational cost: exhaustive selection scores one GP per candidate geometry, although selectors may evaluate these posteriors lazily because every branch is defined from the common history.
If $\Theta_\varepsilon$ is an $\varepsilon$-net of a compact $p$-dimensional parameter set, then $N_\varepsilon\le(C_\Theta/\varepsilon)^p$ for some $C_\Theta<\infty$, so $\log N_\varepsilon=\mathcal{O}(p\log(1/\varepsilon))$; this controls library size only, not approximation error to an off-grid or best continuous warp.
FLIWBO extends fixed-kernel GP-UCB to a finite set of warps with any learning mechanism; it does not guarantee that adaptation is  beneficial.

\section{Experimental Design}
\label{sec:experiments}

We use three evidence tiers: controlled single-run diagnostics that isolate warping; repeated comparisons estimate across-run performance on broader tasks; and a costly MAS case study tests feasibility in the motivating application.
These answer different questions, not interchangeable evidence of optimizer superiority.
All FLIWBO experiments select coordinate-wise Beta-CDF warps from finite libraries using shared optimization history; exact constructions, implementation choices, and configurations are in the supplement and code.

\subsection{Controlled Mechanism Diagnostics}
Four constructed problems isolate mechanisms.
P1 tests resolution reallocation toward a narrow high-value region (Figure~\ref{fig:stationary-vs-warped}); P2 tests recovery of planted latent geometry (top row of Figure~\ref{fig:p2-p4-mechanisms}); P3 tests coordinate-specific warp learning in two dimensions (supplement); and P4 is a stationary negative control favoring fixed-kernel GP-UCB (bottom row of Figure~\ref{fig:p2-p4-mechanisms}).

Each paired trajectory shares initialization, noise, kernel, acquisition rule, and budget between fixed-kernel GP-UCB and FLIWBO-UCB.
P1, P2, and P4 use five initial plus 25 BO evaluations; P3 uses five plus 50.
The supplement gives exact objectives, settings, and complete trajectories.

\subsection{Repeated Optimizer Comparisons}
Table~\ref{tab:comparison-design} gives all run and evaluation counts.
The Confidence-Fence Problem tests whether geometry and acquisition jointly escape an initial-design blind spot; the Warped Gaussian-Mixture and Warped Hartmann6 Problems test known latent distortions in two and six dimensions; and the Fashion-MNIST HPO Problem \citep{xiao2017fashion} asks whether learned warping recovers benefits usually supplied by manual log scaling.

Known-function comparisons include fixed GP-UCB, FLIWBO-UCB, continuous warped GP-EI \citep{snoek2014input}, and oracle GP-UCB/EI given the generating warp; the two higher-dimensional problems also include FLIWBO-EI.
The HPO comparison includes FLIWBO-UCB/EI, continuous warped GP-EI, and fixed GP-UCB/EI under linear and manual-log encodings.
Methods share seeds within each run.
We report mean best-so-far simple regret with 95\% confidence intervals for known functions and incumbent validation loss for HPO.
Only UCB configurations have the corresponding regret guarantees; EI variants are practical comparators.
Task definitions, training details, and comparison caveats are in the supplement.

\begin{table}[t]
\centering
{\small
\setlength{\tabcolsep}{2pt}
\begin{tabular}{@{}lrrrr@{}}
\toprule
Problem & $D$ & Init. & BO & Runs \\
\midrule
Confidence-Fence Problem & 1 & 16 & 100 & 50 \\
Warped Gaussian-Mixture Problem & 2 & 5 & 100 & 50 \\
Warped Hartmann6 Problem & 6 & 5 & 150 & 50 \\
Fashion-MNIST HPO Problem & 5 & 8 & 150 & 20 \\
\bottomrule
\end{tabular}
}
\caption{Repeated-comparison design.
``Init.'' and ``BO'' give the initial and sequential evaluations per run.}
\label{tab:comparison-design}
\end{table}

\subsection{MAS Feasibility Case Study}

With a participating organization, we applied FLIWBO-UCB to a multi-agent system (MAS) design problem: Python program-repair workflows evaluated on QuixBugs \citep{lin2017quixbug}.
An organizational engineer provided a manually designed MAS as a practical baseline.
Each workflow is a 20-dimensional categorical configuration (models, prompts, tool subsets, and agent topology; at most five agents), detailed in the supplement.
Full-workflow evaluations are costly in model calls, tokens, and wall-clock time.
The objective balances repairs against token use:
 $f(x)=R(x)-10^{-5}C_{\mathrm{tok}}(x),$
where $R(x)$ is the number of programs repaired (40 possible) and $C_{\mathrm{tok}}(x)$ is total token use.
We run two FLIWBO-UCB trajectories with three initial and 158 sequential evaluations each, then re-evaluate each selected design and the human baseline 58 times.
The second run changes a prior after inspecting the first, no competing optimizer is run at this cost, and the human design need not lie in the same search space.
Thus the study shows application feasibility and a practical human comparison---not a causal benefit from warping or optimizer superiority.
The supplement gives the search-space encoding and full protocol.

\section{Experimental Results and Discussion}
\label{sec:results}

We report results in the three evidence tiers of Section~\ref{sec:experiments}.
Unless stated otherwise, bands are pointwise 95\% confidence intervals for the mean.

\subsection{Controlled Mechanism Diagnostics}

\begin{table}[t]
    \centering
    {\small
    \setlength{\tabcolsep}{5pt}
    \begin{tabular}{@{}lrr@{}}
    \toprule
    Problem & Fixed GP-UCB & FLIWBO-UCB \\
    \midrule
    P1 & $\approx 4\times10^{-1}$ & $\approx\mathbf{2\times10^{-2}}$ \\
    P2 & $\approx 4\times10^{-2}$ & $\approx\mathbf{7\times10^{-3}}$ \\
    P3 & $\approx 3\times10^{-1}$ & $\approx\mathbf{8\times10^{-5}}$ \\
    P4 & $\approx\mathbf{1\times10^{-4}}$ & $\approx 1\times10^{-3}$ \\
    \bottomrule
    \end{tabular}
    }
    \caption{Final instantaneous regret for matched stationary and warped GP-UCB trajectories.
    Lower is better.
    FLIWBO-UCB wins on P1--P3 but not on the stationary control P4.
    The table reports every paired one-trajectory mechanism diagnostic; Figures~\ref{fig:stationary-vs-warped} and~\ref{fig:p2-p4-mechanisms} explain the one-dimensional mechanisms, and the supplement provides the additional P3 surfaces and complete trajectory plots.}
    \label{tab:mechanism-results}
\end{table}

\begin{figure*}[t]
    \centering
    \begin{minipage}[t]{0.326\textwidth}
        \centering
        \includegraphics[width=\linewidth]{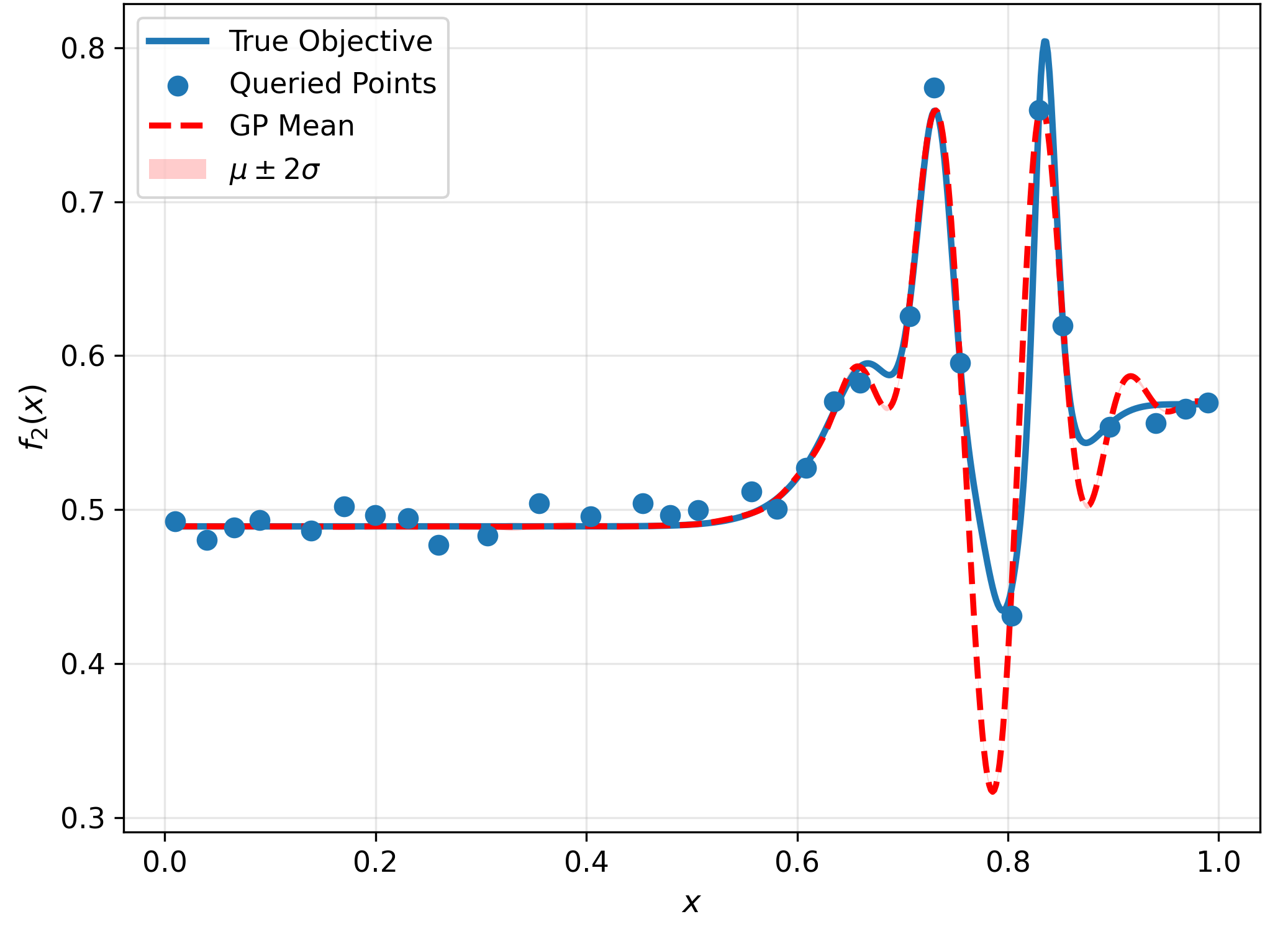}\\[-0.4ex]
        {\scriptsize\textbf{(a) P2:} fixed GP in $x$}
    \end{minipage}\hfill
    \begin{minipage}[t]{0.326\textwidth}
        \centering
        \includegraphics[width=\linewidth]{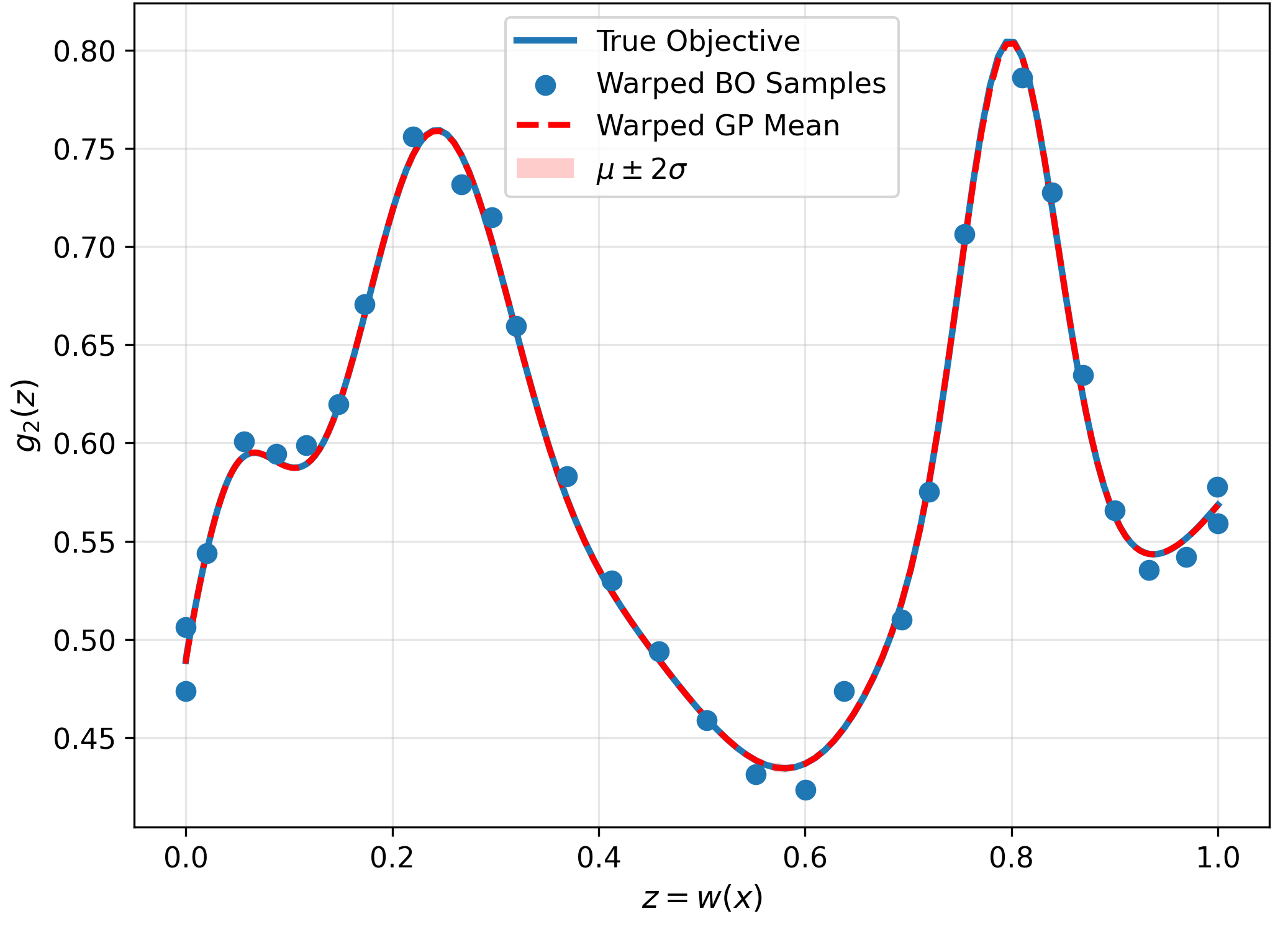}\\[-0.4ex]
        {\scriptsize\textbf{(b) P2:} FLIWBO in $z=w(x)$}
    \end{minipage}\hfill
    \begin{minipage}[t]{0.326\textwidth}
        \centering
        \includegraphics[width=\linewidth]{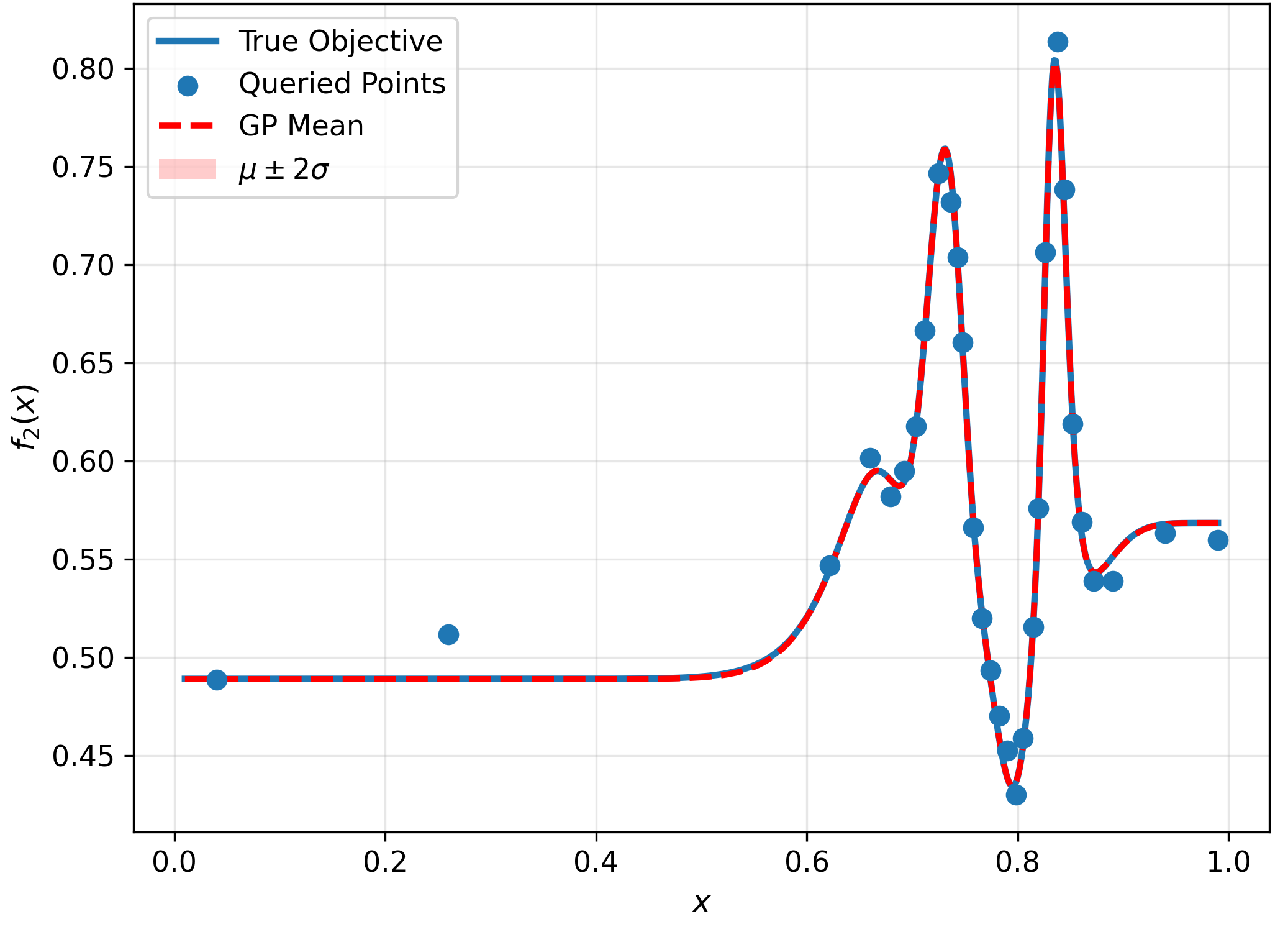}\\[-0.4ex]
        {\scriptsize\textbf{(c) P2:} FLIWBO back in $x$}
    \end{minipage}

    \vspace{1.0ex}
    \begin{minipage}[t]{0.326\textwidth}
        \centering
        \includegraphics[width=\linewidth]{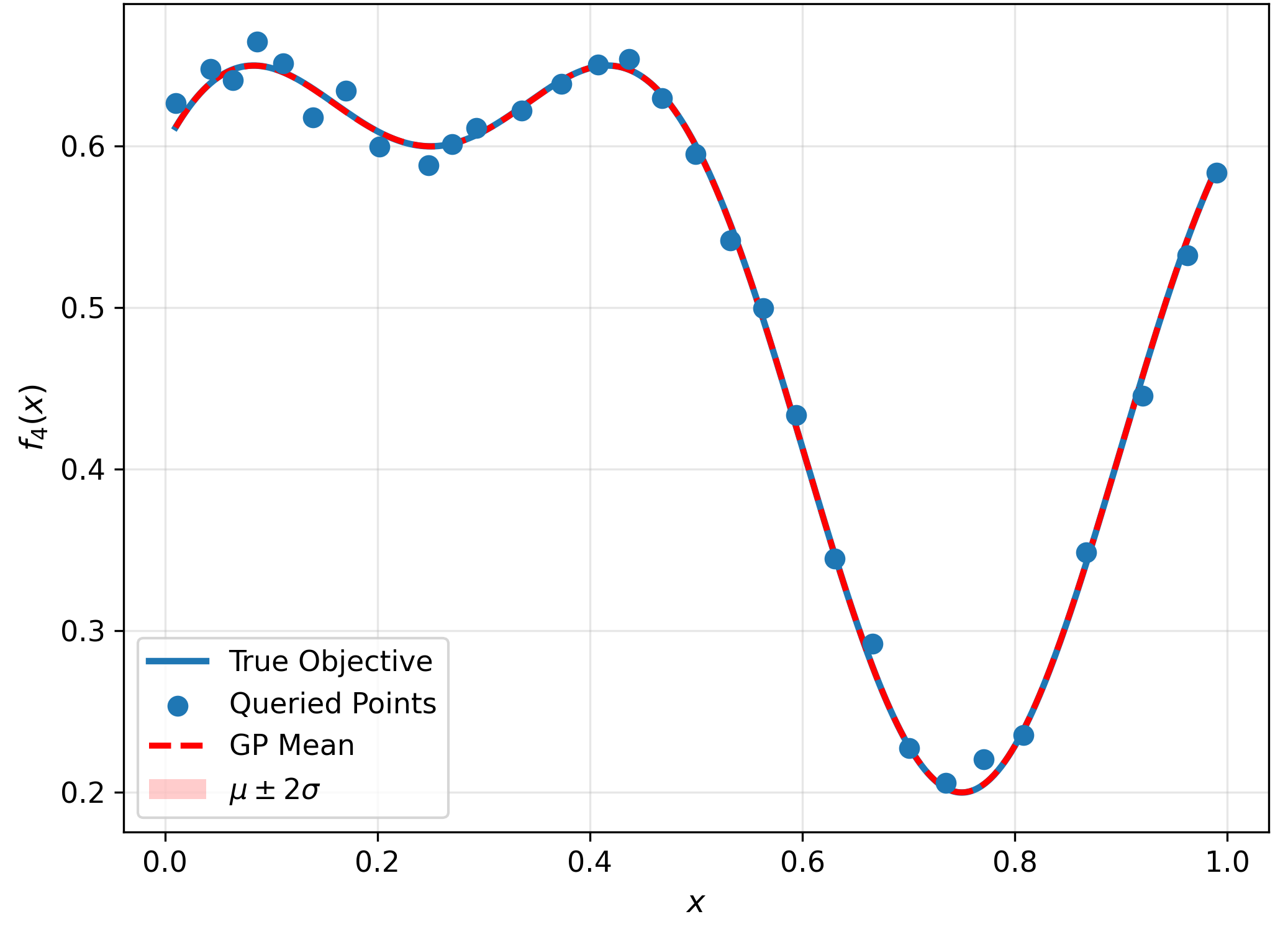}\\[-0.4ex]
        {\scriptsize\textbf{(d) P4:} fixed GP in $x$}
    \end{minipage}\hfill
    \begin{minipage}[t]{0.326\textwidth}
        \centering
        \includegraphics[width=\linewidth]{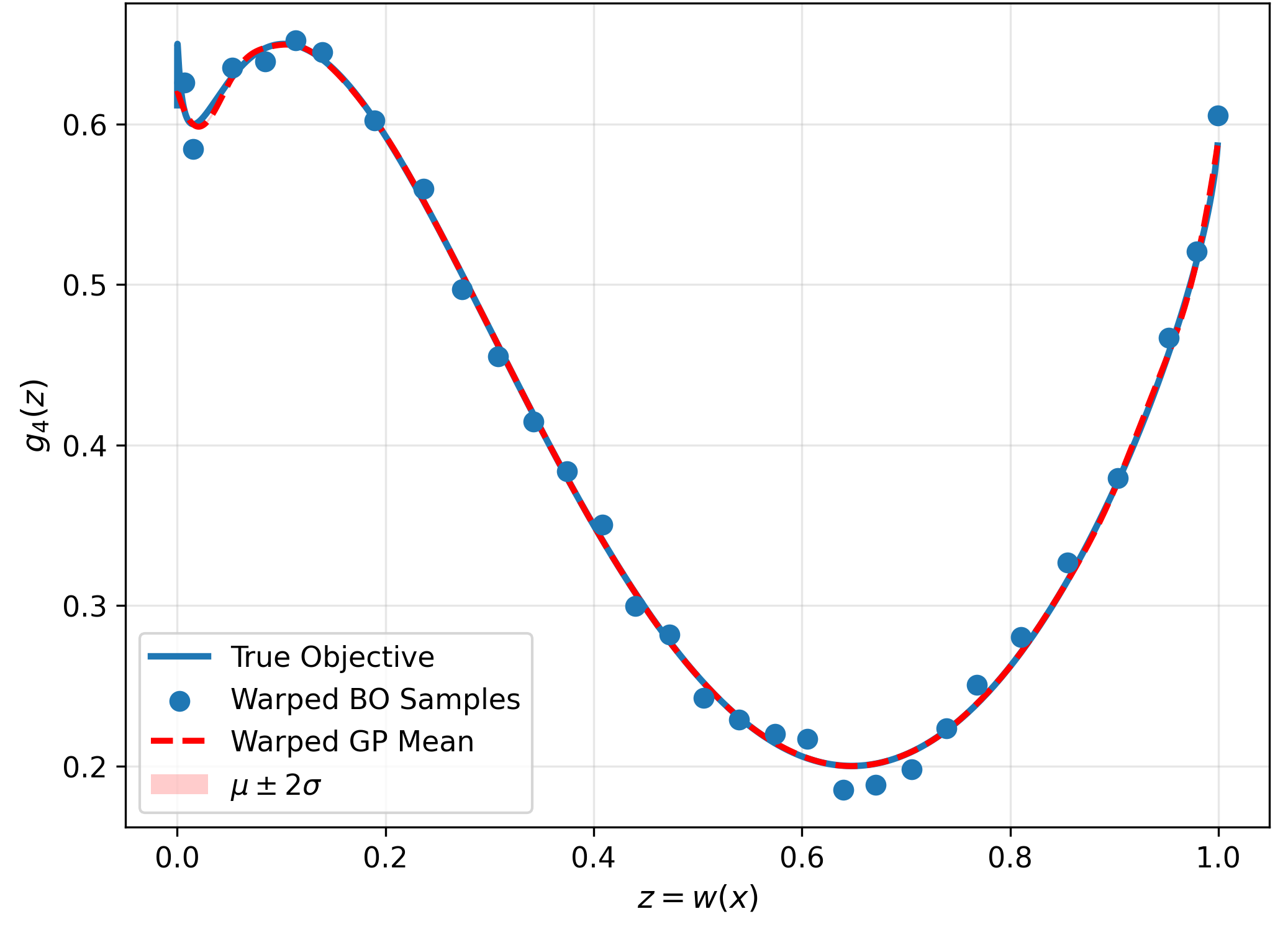}\\[-0.4ex]
        {\scriptsize\textbf{(e) P4:} FLIWBO in $z=w(x)$}
    \end{minipage}\hfill
    \begin{minipage}[t]{0.326\textwidth}
        \centering
        \includegraphics[width=\linewidth]{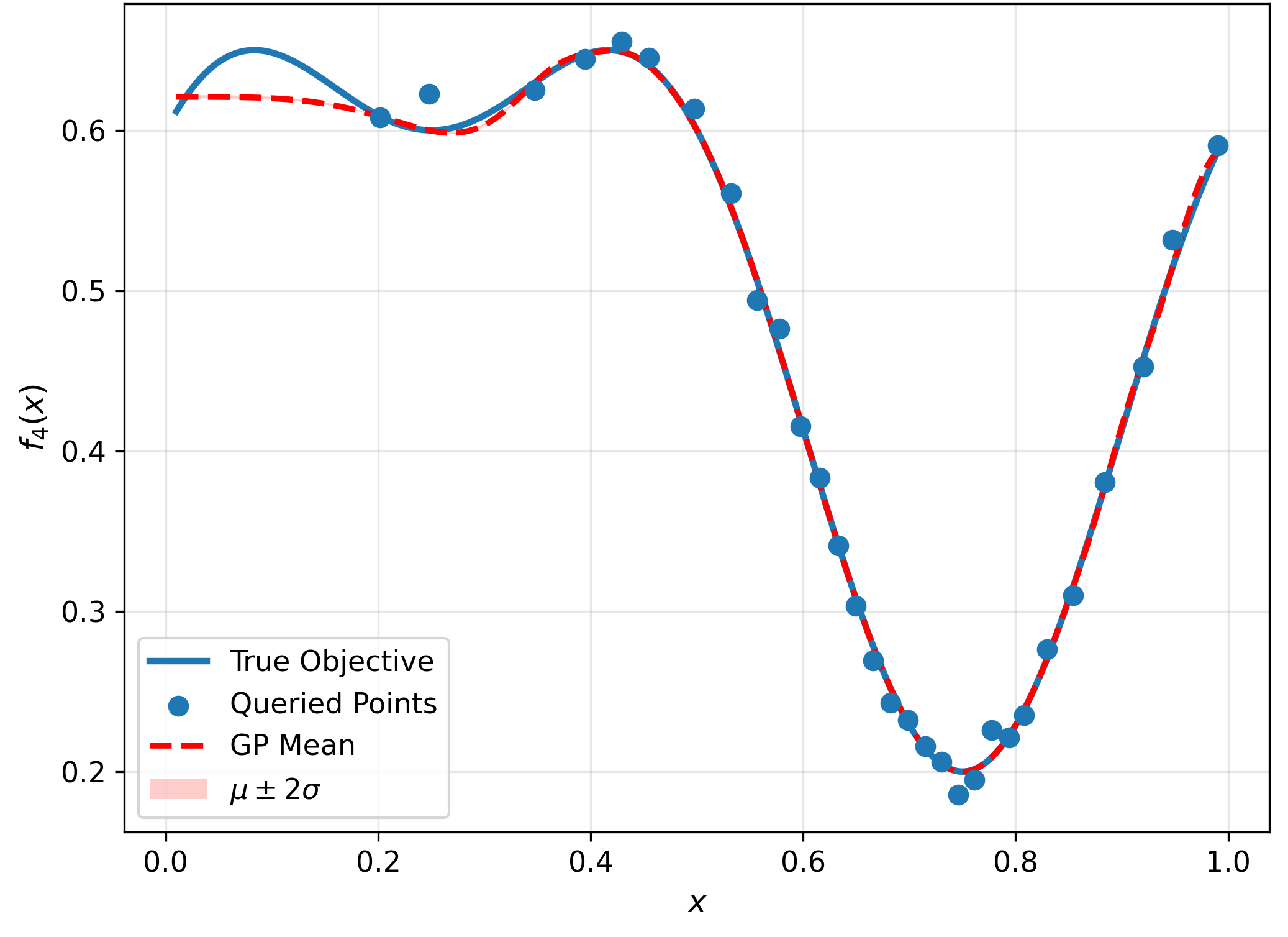}\\[-0.4ex]
        {\scriptsize\textbf{(f) P4:} FLIWBO back in $x$}
    \end{minipage}
    \caption{P2 shows a useful warp repairing planted geometry; P4 shows unnecessary warping making a suitable problem harder.
    In P2, \emph{(a)} one raw-coordinate length scale poorly fits the compressed peak--trough changes.
    \emph{(b)} FLIWBO selects $(\alpha,\beta)=(28.01,8.07)$, close to the planted $(25.093,8.073)$ warp, spreading those changes in $z$ for the stationary base kernel.
    \emph{(c)} Back in $x$, queries concentrate on the compressed structure, tightening the fit and reducing final instantaneous regret by approximately 82.5\%.
    On stationary P4, \emph{(d)} fixed-kernel GP-UCB already fits well.
    \emph{(e)} FLIWBO selects a non-identity warp, $(\alpha,\beta)=(4.09,2.09)$, without simplifying the objective;
    \emph{(f)} poorer left-domain coverage yields higher regret ($10^{-3}$ versus $10^{-4}$).
    Thus adaptive geometry helps by repairing consequential mismatch, not through flexibility alone.}
    \label{fig:p2-p4-mechanisms}
\end{figure*}

Table~\ref{tab:mechanism-results} reports lower final instantaneous regret for FLIWBO-UCB on P1--P3 and for fixed-kernel GP-UCB on stationary P4.
Together, the P1 resolution reallocation (Figure~\ref{fig:stationary-vs-warped}), P2 planted-geometry recovery, and P4 control (Figure~\ref{fig:p2-p4-mechanisms}) show that finite-library warping helps when it expresses consequential mismatch but can distort already appropriate geometry.
The supplement provides P3 surfaces and complete trajectories.

\subsection{Repeated Optimizer Comparisons}

\begin{figure*}[t]
    \centering
    \includegraphics[width=0.495\textwidth]{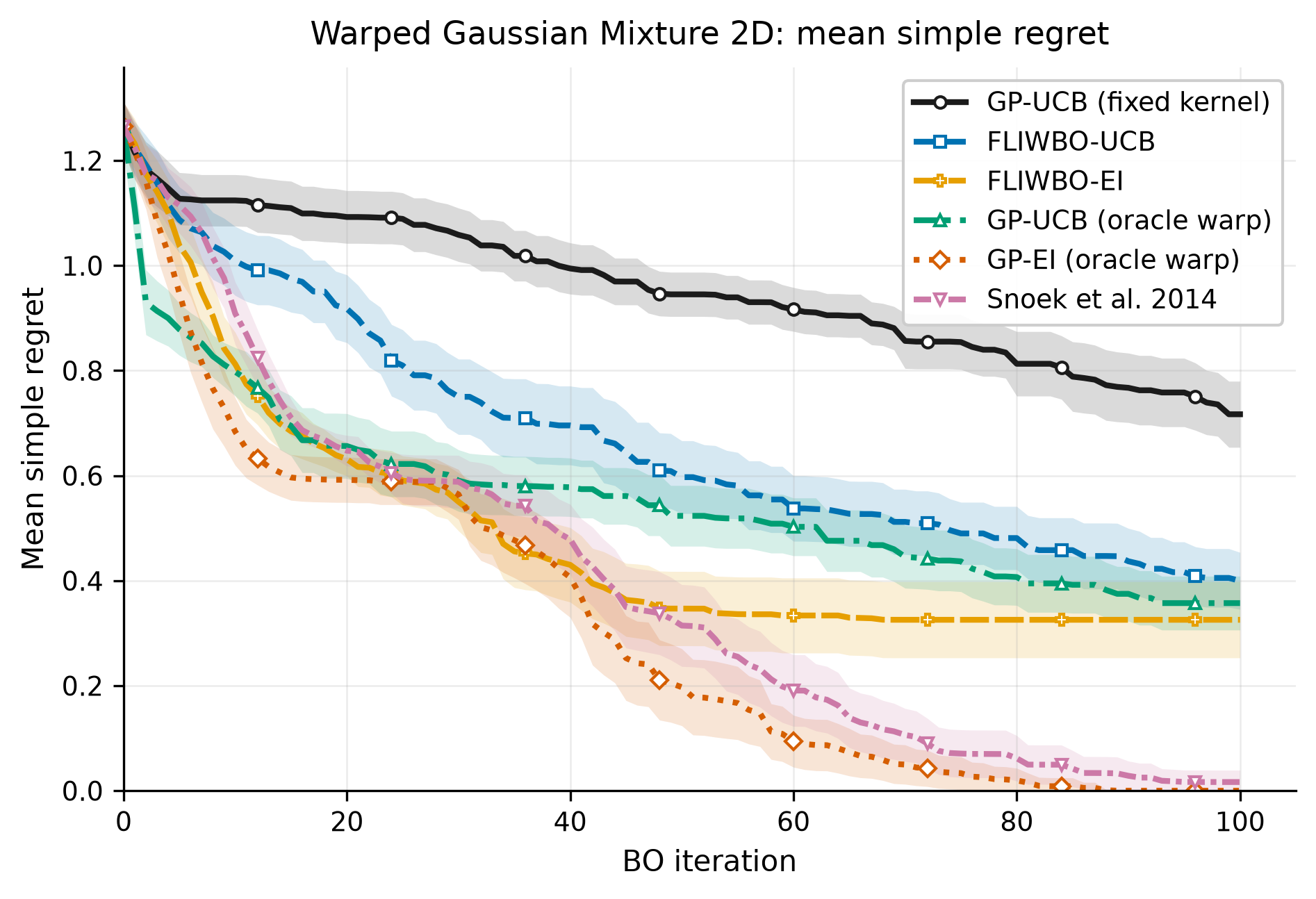}\hfill
    \includegraphics[width=0.495\textwidth]{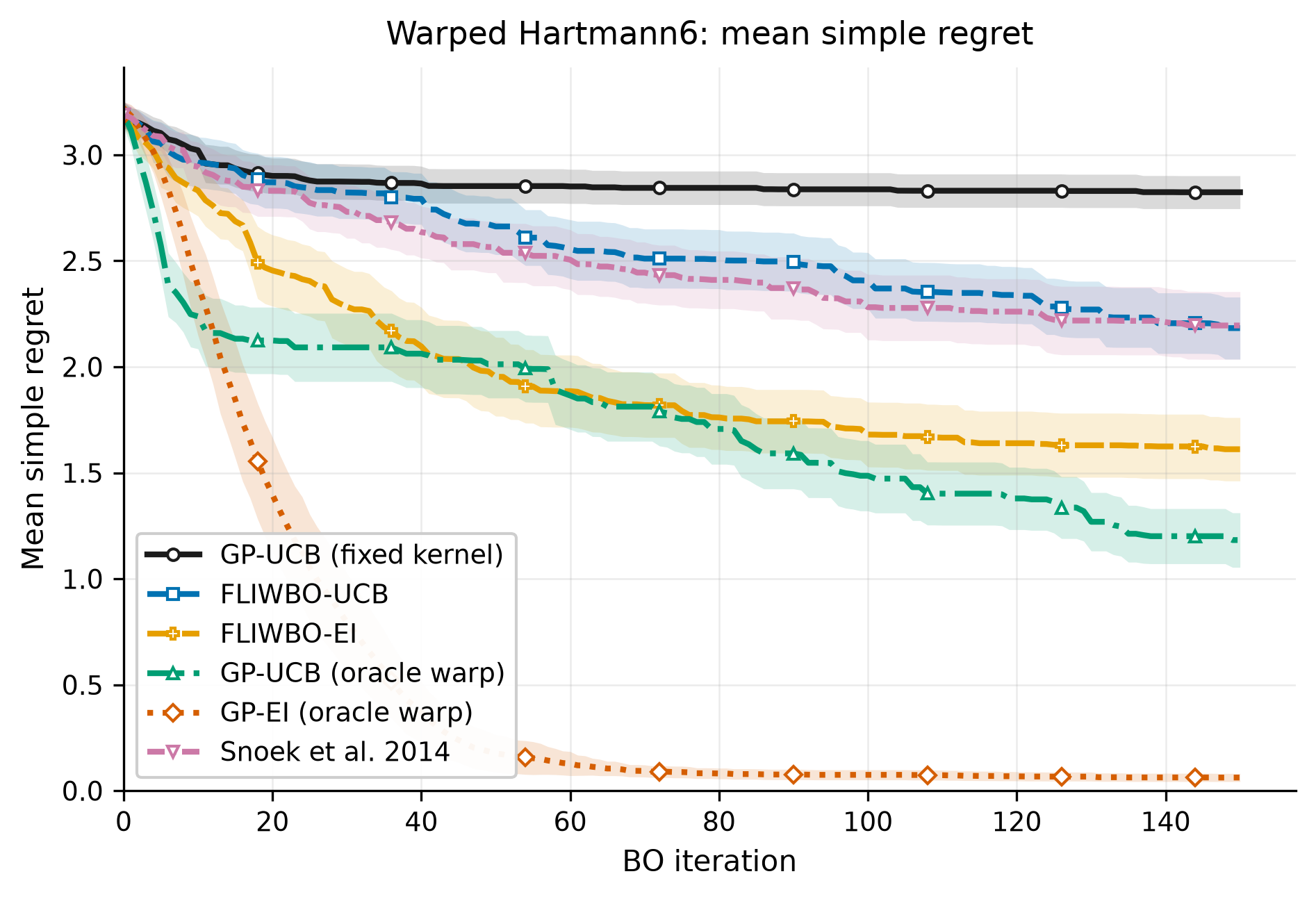}
    \caption{Mean best-so-far simple regret over 50 runs on the Warped Gaussian-Mixture (left) and Warped Hartmann6 (right) Problems; bands are pointwise 95\% confidence intervals.
    FLIWBO-UCB improves on raw-coordinate GP-UCB in both; continuous and oracle curves show more expressive and supplied-geometry references.}
    \label{fig:known-function-results}
\end{figure*}

Figure~\ref{fig:known-function-results} extends the controlled mechanism to two and six dimensions: finite warping improves fixed-kernel UCB when observed coordinates distort the objective.

\begin{figure}[t]
    \centering
    \includegraphics[width=\columnwidth]{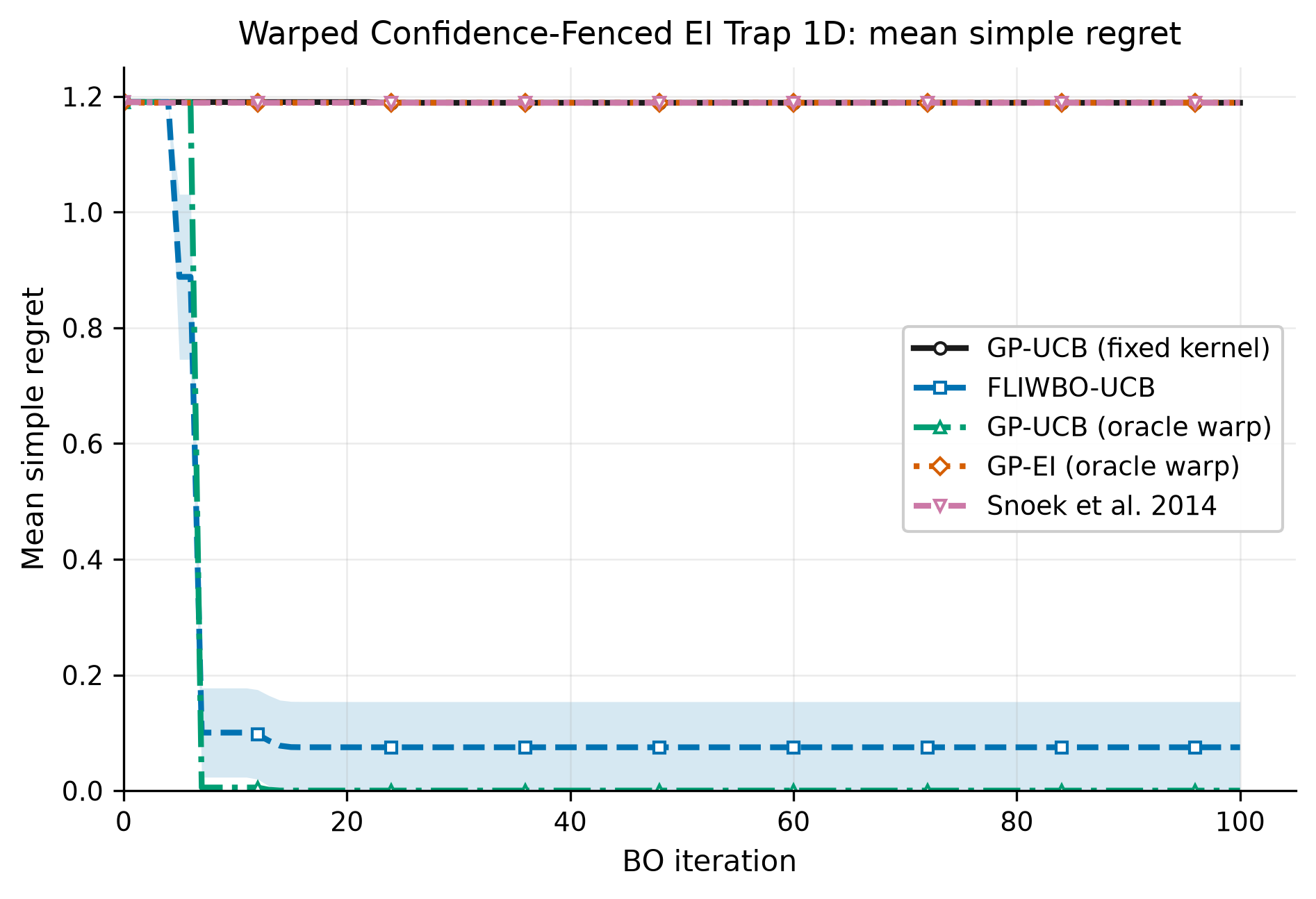}
    \caption{Mean simple regret over 50 Confidence-Fence runs.
    Oracle GP-UCB crosses immediately; fixed GP-UCB and all tested EI variants, including oracle-warp EI, remain trapped.
    FLIWBO-UCB crosses in 47 runs, reaches regret at most $0.01$ in 40, and has zero median final regret, isolating exploration after suitable geometry is available.}
    \label{fig:confidence-fenced-results}
\end{figure}

The Confidence-Fence Problem shows that suitable geometry is insufficient without exploration beyond the local peak; FLIWBO-UCB learns rather than receives that geometry.

\begin{figure}[t]
    \centering
    \includegraphics[width=\columnwidth]{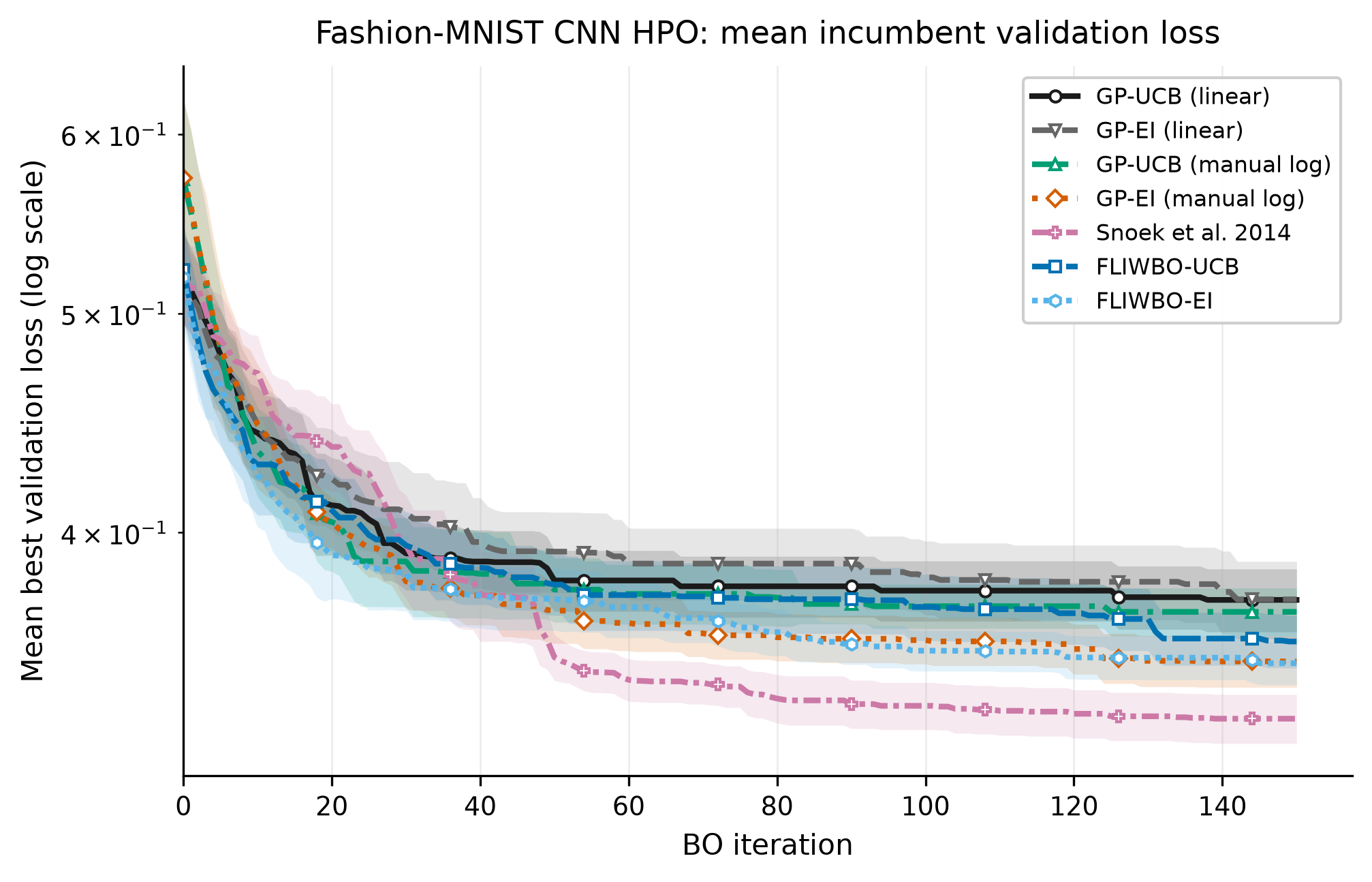}
    \caption{Mean incumbent validation loss over 20 Fashion-MNIST HPO runs; bands are pointwise 95\% confidence intervals.
    FLIWBO improves both acquisitions over linear coordinates and tracks manual log scaling; continuous warped GP-EI has the lowest final mean.
    Unit-cube seed pairing gives linear and log variants different physical initial hyperparameters.}
    \label{fig:hpo-results}
\end{figure}

On Fashion-MNIST HPO, both FLIWBO acquisitions improve on linear coordinates, and FLIWBO-EI tracks manual-log GP-EI without receiving that encoding.
Continuously warped GP-EI provides a target for guarantee-preserving extensions.

Across all four repeated experiments, FLIWBO-UCB has the best final reported mean among tested deployable optimizer families admitting noise-robust frequentist cumulative-regret analysis under their prescribed schedules.
This performance with a matching no-regret guarantee is its central practical strength.

\subsection{MAS Feasibility Case Study}

\begin{table}[t]
\centering
{\small
\setlength{\tabcolsep}{3pt}
\begin{tabular}{lrrc}
\toprule
System & Mean $f(x)$ & SD & 95\% CI \\
\midrule
Human-designed baseline & 24.97 & 2.21 & $[24.38,25.55]$ \\
BO run one & 32.30 & 1.87 & $[31.80,32.79]$ \\
BO run two & 31.31 & 0.82 & $[31.09,31.53]$ \\
\bottomrule
\end{tabular}
}
\caption{MAS objective $f(x)=R(x)-10^{-5}C_{\mathrm{tok}}(x)$ for FLIWBO-selected designs and the human engineer's practical reference ($n=58$ each); higher is better.
Both selected designs outperform the reference under this protocol.}
\label{tab:mas-results}
\end{table}

Under Section~\ref{sec:experiments}'s collaborative protocol, both FLIWBO-selected designs outperform the human engineer's MAS on repeated evaluation, improving mean score by 29.4\% and 25.4\% (Table~\ref{tab:mas-results}).
Together with the completed 161-evaluation trajectories, this shows end-to-end feasibility on a costly, noisy 20-dimensional MAS problem and improvement over the human reference under this protocol.

\paragraph{Overall takeaway.}
The experiments show geometry learning that helps under mismatch but not necessarily under suitable geometry; the best final mean in all four repeated experiments among tested deployable methods with a matching noise-robust frequentist regret guarantee; and MAS feasibility with practical improvement over a human reference.

\section{Conclusion}
\label{sec:conclusion}

FLIWBO extends classical GP-UCB to online selection from a finite library of warped kernels.
Under uniform smoothness and RKHS--Sobolev compatibility, simultaneous confidence across the library yields high-probability sublinear cumulative regret for any history-dependent selector.
The result permits flexible, data-dependent geometry selection while making its cost explicit: logarithmic confidence dependence on $N_\varepsilon$, a $\sqrt{N_\varepsilon}$ regret factor, and one maintained posterior per branch.

The empirical results show that this controlled form of adaptation is practically consequential.
FLIWBO improves optimization under planted coordinate mismatch across one, two, and six dimensions, crosses the Confidence-Fence barrier in 47 of 50 runs while the tested EI variants remain trapped, and recovers much of the benefit of manual log scaling in HPO.
Among the tested deployable optimizer families covered by a matching noise-robust regret guarantee, FLIWBO-UCB achieves the best final mean in all four repeated experiments.
The stationary control identifies the complementary regime in which a fixed geometry is already sufficient, and the MAS collaboration demonstrates end-to-end feasibility under costly, noisy evaluation with a practical comparison against a human-designed reference.

Together, these results establish FLIWBO as a promising route to empirically effective no-regret optimization with learned input geometry.
They also provide a concrete foundation for extending confidence control to continuous or adaptive libraries, branch averaging, and structured domains with explicit approximation control and improved computational scaling.

\section*{Ethical Statement}
Generative AI tools were used to assist with language editing, and LaTeX typesetting.
They were not used as sources of scientific evidence. 
All resulting text, analyses, code, figures, and references were reviewed and verified by the authors, who take full responsibility for the manuscript.

\section*{Acknowledgments}
Research funded by Vinnova Advanced Digitalization (Sweden's Innovation Agency), Försvarsmakten (Swedish Armed Forces), and Saab AB.

\bibliography{references}

\appendix
\section{Problem Formulation and Definitions}
The following section restates the formal problem definition in a concise way for convenience, as well as presenting the suggested algorithm in detail.





\subsection{Objective and Observation Model}
Let
\[
    \mathcal{X} := [0,1]^D
\]
denote the ambient domain, and let $X \subseteq \mathcal{X}$ denote the search domain. 
We assume that $X$ is nonempty and compact. 
The unknown objective function is
\[
    f : X \to \mathbb{R},
\]
and we assume that $f$ is continuous. 
Hence, by compactness of $X$, the set of global maximizers
\[
    X^\star := \argmax_{x \in X} f(x)
\]
is nonempty.

At each round $t$, the algorithm chooses a point $x_t \in X$ and observes
\[
    y_t = f(x_t) + \epsilon_t.
\]
Write
\[
    \mathcal{D}_t := \{(x_s,y_s)\}_{s=1}^t
\]
for the observed data after $t$ evaluations (with $\mathcal{D}_0 := \emptyset$), and let
\[
    \mathcal{F}_t := \sigma(\mathcal{D}_t)
\]
be the $\sigma$-algebra generated by that history (enlarged by any independent algorithmic randomness).
The query $x_t$ is $\mathcal{F}_{t-1}$-measurable.
We assume $\mathbb{E}[\epsilon_t\mid\mathcal{F}_{t-1}]=0$ and $|\epsilon_t|\leq\sigma$ almost surely, and form each GP posterior using regularization (nominal noise variance) $\sigma^2$.

\subsection{Input Warpings}
The purpose of input warping is to change the geometry in which the objective is modelled. 
Rather than changing the objective values themselves, a warp changes how distances and neighborhoods in the input domain are represented by the GP surrogate.

Let $\Theta$ be a parameter set, and for each $\theta \in \Theta$, let
\[
    w_\theta : X \to E_\theta
\]
be a continuously differentiable bijection onto its image
\[
    E_\theta := w_\theta(X) \subseteq \mathcal{X}
\]
We assume that $w_\theta^{-1}:E_\theta \to X$ is well-defined and continuously differentiable. 
In particular, since $X$ is compact and $w_\theta$ is continuous, each warped domain $E_\theta$ is also compact.

In the coordinate-wise case considered later, the warp takes the form
\[
    w_\theta(x) = \left( w_{\theta,1}(x_1), \ldots, w_{\theta,D}(x_D) \right)
\]
where each coordinate map is strictly increasing and continuously differentiable. 
We assume that the warp family is uniformly regular. 
For intuition, in the coordinate-wise case one may think of this regularity as requiring
\[
    \frac{1}{M} \leq w'_{\theta,i}(x_i) \leq M
\]
Equivalently, in the multivariate formulation, the Jacobian of $w_\theta$ and the Jacobian of $w_\theta^{-1}$ are uniformly bounded over the warp family.
The more explicit version needed in the regret proof is stated formally in Assumption~\ref{ass:regular_warp}.

\subsection{Warped Objectives and Local Warp Quality}

For every admissible warp $w_\theta$, we define the corresponding warped objective
\[
    g_\theta : E_\theta \to \mathbb{R}
\]
by
\[
    g_\theta(z) := f(w_\theta^{-1}(z))
\]
Equivalently,
\[
    f(x) = g_\theta(w_\theta(x))
\]
Thus, $g_\theta$ is not a different objective, it is the same objective expressed in the warped coordinates.

The motivation for using input warps is that some coordinate systems may make the optimization problem easier for a GP surrogate. 
In particular, a warp that expands the domain near a maximizer $x^\star \in X^\star$ makes a small high-value region occupy a larger region in warped coordinates. 
Conversely, a warp that contracts the domain near the maximizer may make the high-value region harder to locate.

To formalize this intuition, let
\[
    B_\rho(X^\star) := \{x \in X : \operatorname{dist}(x, X^\star) \leq \rho\}
\]
denote a $\rho$-neighborhood of the maximizer set. 
For a differentiable warp $w_\theta$, define the local expansion score
\[
    q_\rho(\theta) := \inf_{x \in B_\rho(X^\star)} \sigma_{\min}\left(J_{w_\theta}(x)\right)
\]
where $J_{w_\theta}(x)$ is the Jacobian of $w_\theta$ at $x$, and $\sigma_{\min}$ denotes its smallest singular value. 
In the one-dimensional case this reduces to
\[
    q_\rho(\theta)
    =
    \inf_{x \in B_\rho(X^\star)}
    w_\theta'(x)
\]
We say that a warp is locally better near the maximizer if it has a larger local expansion score. 
For fixed $\rho>0$ and $\varepsilon>0$, one may call a warp $\varepsilon$-good if
\[
    q_\rho(\theta) \geq M-\varepsilon
\]
and $\varepsilon$-bad if its local expansion near $X^\star$ is close to the lower regularity bound. 
These scores are not used in the main proof.
The terminology is only meant to capture the geometric intuition that good warps expand neighborhoods of optimizers, while bad warps contract them.

\subsection{Finite Warp Library and Algorithm Outline}

For the theoretical analysis, we restrict attention to a finite library of warp parameters
\[
    \Theta_\varepsilon
    :=
    \{\theta^{(1)},\ldots,\theta^{(N_\varepsilon)}\}
    \subseteq \Theta
\]
Each library element induces a kernel
\begin{definition}[Warp-induced kernel family]
For each $\theta \in \Theta$, we define the kernel on $X$ as
\begin{equation*}
    k_\theta(x,x') := k_0\left(w_\theta(x),w_\theta(x')\right)
\end{equation*}
For a base kernel $k_0$ defined on $\mathcal{X}$ such that $k_0(x,x) \leq 1$ for all $x$.
\end{definition}

The common history defines a GP posterior for each kernel in the finite library.
A selector rule, such as a maximum-a-posteriori selector, may evaluate these posteriors exhaustively or lazily before choosing the model that determines the next GP-UCB query point.

Algorithm \ref{alg:fliw-gpucb} presents the suggested finite-library input-warped BOGP-UCB algorithm that is central to the proposed method in this paper.
\begin{algorithm}[t]
\caption{Selective Finite-Library Input-Warped GP-UCB}
\label{alg:fliw-gpucb}
\begin{algorithmic}[1]
\REQUIRE Search domain $\mathcal{X} \subseteq [0,1]^D$, base kernel $k_0$, finite warp library $\Theta_\varepsilon = \{\theta^{(1)}, \dots, \theta^{(N_\varepsilon)}\}$, selector $s$, noise variance $\sigma^2$, confidence sequence $\beta_t$

\STATE For each $i \in \{1, \dots, N_\varepsilon\}$, define the warped kernel
\begin{equation*}
    k_i(x,x') := k_{\theta^{(i)}}(x,x')
\end{equation*}

\STATE Initialize dataset $\mathcal{D}_0$

\FOR{$t = 1,2,\dots,T$}
    \STATE Let $\mu_{i,t-1},\sigma_{i,t-1}$ denote the posterior under $k_i,\mathcal{D}_{t-1}$ and define $U_{i,t-1}=\mu_{i,t-1}+\sqrt{\beta_t}\,\sigma_{i,t-1}$ for every $i$

    \STATE Select a library index
    \begin{equation*}
        j_t \gets s(\mathcal{D}_{t-1})
    \end{equation*}

    \STATE Select the query point
    \begin{equation*}
        x_t \in \arg\max_{x \in \mathcal{X}} U_{j_t,t-1}(x)
    \end{equation*}

    \STATE Observe noisy feedback
    \begin{equation*}
        y_t = f(x_t) + \epsilon_t
    \end{equation*}

    \STATE Update the dataset
    \begin{equation*}
        \mathcal{D}_t \gets \mathcal{D}_{t-1} \cup \{(x_t,y_t)\}
    \end{equation*}
\ENDFOR

\STATE \textbf{return} $\{(x_t,y_t)\}_{t=1}^T$
\end{algorithmic}
\end{algorithm}

For the coordinate-wise experiments, let the fixed one-coordinate library contain $L$ parameter pairs.
The predeclared family of complete $D$-dimensional warps is its Cartesian product, so $N_\varepsilon=L^D$; $S$ coordinate sweeps implement a lazy selector within that family using $SDL+1$ GP fits per round, each on the full common history.

\section{No-regret Properties for Finite-library Input Warped BOGP}

\restatable{%
\begin{assumption}[Uniform warp regularity]
\label{ass:regular_warp}
For each $\theta \in \Theta$, define the warped domain $E_\theta := w_\theta(X)$.
Assume there exists an integer $m \ge 1$ and a constant $C_w \ge 1$ such that for every $\theta \in \Theta$, the map
\begin{equation*}
    w_\theta : X \to E_\theta
\end{equation*}
is a $C^m$ diffeomorphism, and that the family $\{w_\theta : \theta \in \Theta\}$ has uniformly bounded $C^m$ norms, and that the same holds for their inverses.
\end{assumption}
}

In practice we use the coordinatewise Beta-CDF warp of Snoek et al.~\cite{snoek2014input} which fulfills Assumption~\ref{ass:regular_warp} on the slightly restricted search domain $X = [\tau, 1 - \tau]^D$ for some $\tau \in \left(0,\frac{1}{2}\right)$.
It is possible to apply other warp functions, such as regularized Beta-CDF, logit-affine-sigmoid warps or otherwise, on the full domain $\mathcal{X}$ as long as Assumption~\ref{ass:regular_warp} is fulfilled everywhere. 
Conditioned on $\theta$, regression with kernel $k_\theta$ is equivalent to standard Gaussian process regression applied to the warped inputs, at least on the search domain $X$.

For the regret analysis, we assume that there exists a witness parameter $\theta_0$ in the admissible class and a witness representation $g_0 \in \mathcal{H}_{k_0}$ such that
\[
    f(x) = g_0(w_{\theta_0}(x))
\]
Equivalently, the restriction of $g_0$ to $E_0:=w_{\theta_0}(X)$ is the warped objective associated with the witness coordinate system.
The witness is only an analytical device: it need not be unique, need not belong to the finite library, and is not interpreted as a true warp.
\restatable{%
\begin{assumption}[Witness RKHS bound]
\label{ass:rkhs}
The function $g_0 \in \mathcal{H}_{k_0}$.
Moreover, there exists $B_g < \infty$ such that
\begin{equation*}
    \|g_0\|_{\mathcal{H}_{k_0}} \le B_g
\end{equation*}
\end{assumption}
}
This encodes that the witness representation $g_0$ is assumed to lie in the RKHS $\mathcal{H}_{k_0}$ of the base kernel $k_0$.
The corresponding RKHS boundedness of $f$ under the witness-induced kernel follows from the next lemma.

\restatable{%
\begin{lemma}[Warp preserves RKHS boundedness]
\label{lem:rkhs_warp}
For any $\theta\in\Theta$, if $h \in \mathcal{H}_{k_0}$ and $f(x) := h\left(w_\theta(x)\right)$, then
\begin{equation*}
    f \in \mathcal{H}_{k_\theta} \qquad \text{and} \qquad \|f\|_{\mathcal{H}_{k_\theta}} \le \|h\|_{\mathcal{H}_{k_0}}
\end{equation*}
\end{lemma}
}

\begin{proof}
Let
\begin{equation*}
    E_\theta := w_\theta\left(X\right) \subseteq \mathcal{X}
\end{equation*}
and define the restriction
\begin{equation*}
    u := h|_{E_\theta}
\end{equation*}
By Aronszajn's restriction theorem  \cite{aronszajn1950theory}, $u$ belongs to the RKHS with reproducing kernel $k_0|_{E_\theta \times E_\theta}$ and satisfies
\begin{equation*}
    \|u\|_{\mathcal{H}_{k_0|_{E_\theta \times E_\theta}}} \le \|h\|_{\mathcal{H}_{k_0}}
\end{equation*}
Since
\begin{equation*}
    f(x) = h\left(w_\theta(x)\right) = u\left(w_\theta(x)\right)
\end{equation*}
the class
\begin{equation*}
    \left\{u \circ w_\theta : u \in \mathcal{H}_{k_0|_{E_\theta \times E_\theta}}\right\}
\end{equation*}
is an RKHS on $X$ with reproducing kernel
\begin{equation*}
    k_\theta(x,x') = k_0\left(w_\theta(x),w_\theta(x')\right)
\end{equation*}
Therefore
\begin{equation*}
    \begin{aligned}
        f &\in \mathcal{H}_{k_\theta}, \\
        \|f\|_{\mathcal{H}_{k_\theta}}
        &= \|u\|_{\mathcal{H}_{k_0|_{E_\theta \times E_\theta}}} \\
        &\le \|h\|_{\mathcal{H}_{k_0}}.
    \end{aligned}
\end{equation*}

\end{proof}

\begin{corollary}[Bounded RKHS norm under the witness-induced kernel]
\label{cor:witness-kernel-rkhs}
Under Assumption~\ref{ass:rkhs} and Lemma~\ref{lem:rkhs_warp}, we have
\begin{equation*}
    \|f\|_{\mathcal{H}_{k_{\theta_0}}} \le B_g
\end{equation*}
\end{corollary}

\subsection{RKHS Transfer}
%
\begin{remark}[Size of the $\varepsilon$-net]
If the finite library $\Theta_\varepsilon = \{ \theta^{(1)}, \dots, \theta^{(N_\varepsilon)} \} \subset \Theta$ is chosen as an $\varepsilon$-net of the compact parameter set $\Theta \subset \mathbb R^p$, then its cardinality can be bounded:
\begin{equation*}
    N_\varepsilon \le \left( \frac{ C_\Theta}{\varepsilon} \right)^p
\end{equation*}
for some constant $C_\Theta < \infty$ depending only on $\Theta$.
Hence
\begin{equation*}
    \log N_\varepsilon = \mathcal O\left( p\log(1 / \varepsilon) \right)
\end{equation*}
\end{remark}

Let $S^m(\Omega)$ denote the Sobolev space of functions on the domain $\Omega$ whose weak derivatives up to order $m$ lie in $L^2(\Omega)$.
Intuitively, $S^m$ contains functions that are smooth up to $m$ derivatives.
In our analysis, the integer $m$ is not arbitrary, it is tied to the smoothness level implied by the base kernel, e.g. a Matérn kernel with sufficient $\nu$ through $m = \nu + D/2$. 
For this type of kernel, the associated RKHS is equivalent to a Sobolev space, which is useful for analyzing composition with smooth maps such as $w_\theta(x)$.
\restatable{%
\begin{assumption}[Sobolev--RKHS equivalence]
\label{ass:finite_net_base_rkhs}
There exist an integer $m \ge 1$ and a constant $C_{\mathrm{eq}} \ge 1$ such that, with
\begin{equation*}
    E_0 := w_{\theta_0}(X), \qquad E_i := w_{\theta^{(i)}}(X), \quad i=1,\dots,N_\varepsilon,
\end{equation*}
for every set $E \in \left\{E_0, E_1, \dots, E_{N_\varepsilon}\right\}$,
\begin{equation*}
    \mathcal{H}_{k_0|_{E \times E}} = S^m(E)
\end{equation*}
as a set of functions, and for every $u \in \mathcal{H}_{k_0|_{E \times E}} = S^m(E)$,
\begin{equation*}
    C_{\mathrm{eq}}^{-1}\|u\|_{S^m(E)} \le \|u\|_{\mathcal{H}_{k_0|_{E \times E}}} \le C_{\mathrm{eq}}\|u\|_{S^m(E)}
\end{equation*}
\end{assumption}
}

Assumption~\ref{ass:finite_net_base_rkhs} applies to the base kernel $k_0$ and is used only to move between RKHS norms and Sobolev norms on the finitely many warped domains appearing in the library $\Theta_\varepsilon$.
This assumption holds, for example, for Matérn kernels whenever the induced Sobolev order is an integer. 
See \cite{kanagawa2018gaussian} for details.

\begin{remark}[Regularity of the relative warps]
\label{rem:finite_library_relative_warp}
For each $i = 1, \dots, N_\varepsilon$, define the relative warp
\begin{equation*}
    \phi_i := w_{\theta_0}\circ w_{\theta^{(i)}}^{-1} : E_i \to E_0
\end{equation*}
Under the uniform regularity Assumption~\ref{ass:regular_warp} above, each $\phi_i$ is a $C^m$ diffeomorphism. 
Moreover, its derivatives and the derivatives of its inverse up to order $m$ are bounded uniformly across $i$.
This follows immediately from the corresponding bounds for $w_{\theta_0}$, $w_{\theta^{(i)}}$, and $w_{\theta^{(i)}}^{-1}$ by composition and the chain rule.
\end{remark}

\restatable{%
\begin{lemma}[Uniform composition bound on $S^m$]
\label{lem:finite_library_composition}
Under Assumption~\ref{ass:regular_warp}, there exists a constant $C_\mathrm{comp} < \infty$ such that for all $i \in \left\{1,\dots,N_\varepsilon\right\}$ and all $u \in S^m(E_0)$,
\begin{equation*}
    \|u \circ \phi_i\|_{S^m(E_i)} \le C_\mathrm{comp}\|u\|_{S^m(E_0)}
\end{equation*}
\end{lemma}
}

\begin{proof}
As stated in Remark~\ref{rem:finite_library_relative_warp}, each $\phi_i$ is a $C^m$ diffeomorphism between $E_i$ and $E_0$, and the derivatives of $\phi_i$ and $\phi_i^{-1}$ up to order $m$ are uniformly bounded over $i$.
Therefore, we can apply a Sobolev composition theorem for $C^m$ diffeomorphisms \cite{wendl_sobolev, adams2003sobolev}.
Consequently, for each $i$, there exists a constant $C_i < \infty$ depending only on $m$, $D$, and the uniform bounds such that
\begin{equation*}
    \|u \circ \phi_i\|_{S^m(E_i)} \le C_i \|u\|_{S^m(E_0)}
\end{equation*}
Since the library is finite, one constant may be chosen uniformly over $i$, $C_\mathrm{comp} := \underset{i}{\max}\ C_i < \infty$.
\end{proof}

\restatable{%
\begin{theorem}[Uniform RKHS transfer over the finite library]
\label{thm:finite_library_rkhs_transfer}
\label{thm:main-uniform-compatibility}
Let
\begin{equation*}
    f(x) = g_0\left( w_{\theta_0}(x) \right), \qquad g_0 \in \mathcal{H}_{k_0}, \qquad \|g_0\|_{\mathcal{H}_{k_0}} \le B_g
\end{equation*}
Under Assumptions~\ref{ass:regular_warp} and~\ref{ass:finite_net_base_rkhs}, there exists a constant $C_{\mathrm{warp}}$ such that for every $i \in \left\{1, \dots, N_\varepsilon\right\}$,
\begin{equation*}
    f \in \mathcal{H}_{k_{\theta^{(i)}}} \qquad \text{and} \qquad \|f\|_{\mathcal{H}_{k_{\theta^{(i)}}}} \le C_{\mathrm{warp}}
\end{equation*}
\end{theorem}
}

\begin{proof}
Fix $i \in \left\{1, \dots,N_\varepsilon \right\}$ and write
\begin{equation*}
    \begin{aligned}
        E_0 &= w_{\theta_0}(X),
        & E_i &= w_{\theta^{(i)}}(X), \\
        \phi_i &= w_{\theta_0} \circ w_{\theta^{(i)}}^{-1}
        &&: E_i \to E_0.
    \end{aligned}
\end{equation*}
Let
\begin{equation*}
    u_0 := g_0|_{E_0}
\end{equation*}
By Aronszajn's restriction theorem \cite{aronszajn1950theory},
\begin{equation*}
    \begin{aligned}
        u_0 &\in \mathcal{H}_{k_0|_{E_0 \times E_0}}, \\
        \|u_0\|_{\mathcal{H}_{k_0|_{E_0 \times E_0}}}
        &\le \|g_0\|_{\mathcal{H}_{k_0}} \le B_g.
    \end{aligned}
\end{equation*}

Assumption~\ref{ass:finite_net_base_rkhs} then gives
\begin{equation*}
    \|u_0\|_{S^m(E_0)} \le C_{\mathrm{eq}} B_g
\end{equation*}

Define
\begin{equation*}
    u_i := u_0 \circ \phi_i \qquad \text{on } E_i
\end{equation*}

By Lemma~\ref{lem:finite_library_composition},
\begin{equation*}
    \|u_i\|_{S^m(E_i)} \le C_{\mathrm{comp}} \|u_0\|_{S^m(E_0)} \le C_{\mathrm{comp}} C_{\mathrm{eq}} B_g
\end{equation*}

Applying Assumption~\ref{ass:finite_net_base_rkhs} once more,
\begin{equation*}
    \|u_i\|_{\mathcal{H}_{k_0|_{E_i \times E_i}}} \le C_{\mathrm{eq}} \|u_i\|_{S^m(E_i)} \le C_{\mathrm{eq}}^2 C_{\mathrm{comp}} B_g
\end{equation*}

For every $x \in X$,
\begin{equation*}
    \begin{aligned}
        u_i\left(w_{\theta^{(i)}}(x)\right)
        &= u_0\left(\phi_i\left(w_{\theta^{(i)}}(x)\right)\right) \\
        &= u_0\left(w_{\theta_0}(x)\right) \\
        &= g_0\left(w_{\theta_0}(x)\right) = f(x).
    \end{aligned}
\end{equation*}

Thus
\begin{equation*}
    f(x) = u_i\left( w_{\theta^{(i)}}(x) \right)
\end{equation*}

By Aronszajn's restriction theorem, since $u_i \in \mathcal H_{k_0|E_i \times E_i}$ there exists an extension $\tilde{u}_i \in \mathcal{H}_{k_0}$ such that
\begin{equation*}
    \tilde{u}_i |_{E_i} = u_i \quad \text{and} \quad \| \tilde{u}_i \|_{\mathcal{H}_{k_0}} = \|u_i\|_{\mathcal{H}_{k_0 |E_i \times E_i}}
\end{equation*}
Applying Lemma~\ref{lem:rkhs_warp} to $\tilde{u}_i$, we obtain that $\tilde{u}_i \circ w_{\theta^{(i)}}$ belongs to $\mathcal{H}_{k_{\theta^{(i)}}}$. 
Since $w_{\theta^{(i)}}(x) \in E_i$, we have
\begin{equation*}
    \tilde{u}_i \left( w_{\theta^{(i)}}(x) \right) = u_i \left( w_{\theta^{(i)}}(x) \right) = f(x)
\end{equation*}
Therefore, $f \in \mathcal{H}_{k_{\theta^{(i)}}}$.

Define $C_{\mathrm{warp}} := C_{\mathrm{eq}}^2 C_{\mathrm{comp}} B_g$, which proves the claim.
\end{proof}

\subsection{Information Gain}

\begin{definition}[Maximum information gain]
\label{def:information-gain}
For any sequence $x_1, \dots,x_T \in X$, and a kernel $k$, let
\begin{equation*}
    K_T := \left[ k(x_i,x_j) \right]_{i,j=1}^T
\end{equation*}
denote the $T \times T$ kernel matrix.
Define the maximum information gain as in ~\cite{srinivas2010gaussian},
\begin{equation*}
    \gamma_T(k,X) := \sup_{x_1, \dots, x_T \in X} \frac{1}{2} \log \det \left( I + \sigma^{-2} K_T \right)
\end{equation*}
where $I$ is the $T \times T$ identity matrix and $\sigma^2$ is the noise variance parameter.
\end{definition}

\restatable{%
\begin{lemma}[Variance accumulation for a fixed kernel]
\label{lem:fixed-kernel-variance}
For $k(x,x)\leq1$, let $C_{\mathrm{info}}:=2/\log(1+\sigma^{-2})$.
Then for any adaptively chosen query sequence $x_1,\dots,x_T$ and any fixed kernel $k$,
\begin{equation*}
    \sum_{t=1}^T \sigma_{t-1}^2(x_t) \le C_{\mathrm{info}}\gamma_T(k,X)
\end{equation*}
\end{lemma}
}

\begin{proof}
For the realized sequence $x_1,\dots,x_T$, let
\begin{equation*}
    K_T := [k(x_i,x_j)]_{i,j=1}^T
\end{equation*}
By Lemma 5.3 of Srinivas et al.~\cite{srinivas2010gaussian},
\begin{equation*}
\frac{1}{2} \log \det \left( I + \sigma^{-2} K_T\right) = \frac{1}{2} \sum_{t=1}^T \log \left( 1 + \sigma^{-2} \sigma_{t-1}^2(x_t) \right)
\end{equation*}
Since $u_t:=\sigma_{t-1}^2(x_t)\in[0,1]$, concavity of $u\mapsto\log(1+\sigma^{-2}u)$ gives
\begin{equation*}
    u_t\log(1+\sigma^{-2})
    \leq \log(1+\sigma^{-2}u_t).
\end{equation*}
Summing over $t$ and using the preceding identity gives
\begin{equation*}
    \sum_{t=1}^T \sigma_{t-1}^2(x_t) \le C_{\mathrm{info}} \frac{1}{2} \log \det \left( I + \sigma^{-2} K_T \right)
\end{equation*}
By Definition 2,
\begin{equation*}
    \frac{1}{2} \log \det \left( I + \sigma^{-2} K_T \right) \le \gamma_T(k,X)
\end{equation*}
and therefore
\begin{equation*}
    \sum_{t=1}^T \sigma_{t-1}^2(x_t) \le C_{\mathrm{info}} \gamma_T(k,X)
\end{equation*}
\end{proof}

\restatable{%
\begin{lemma}[Maximum information gain under a warp]
\label{lem:gamma_warp_transfer}
For every $\theta \in \Theta$ and every $T \ge 1$,
\begin{equation*}
    \gamma_T(k_\theta, X) = \gamma_T \left(k_0, w_\theta(X) \right) \le \gamma_T \left(k_0, \mathcal{X} \right)
\end{equation*}
\end{lemma}
}

\begin{proof}
Fix $\theta \in \Theta$ and $T \ge 1$.
Let $x_1, \dots, x_T \in X$ be an arbitrary sequence, and define
\begin{equation*}
    z_t := w_\theta(x_t) \in w_\theta(X) \qquad t = 1, \dots, T
\end{equation*}

Let the kernel matrices along the sequence be
\begin{equation*}
    \begin{aligned}
        K_{\theta,T} &:= \left[k_\theta(x_p,x_q)\right]_{p,q=1}^T, \\
        K_{0,T} &:= \left[k_0(z_p,z_q)\right]_{p,q=1}^T.
    \end{aligned}
\end{equation*}

Then $K_{\theta,T}=K_{0,T}$, and therefore
\begin{equation*}
    \frac{1}{2}\log\det\left(I+\sigma^{-2}K_{\theta,T}\right)
    = \frac{1}{2}\log\det\left(I+\sigma^{-2}K_{0,T}\right).
\end{equation*}
Hence the log-det quantity in Definition~\ref{def:information-gain} for the sequence $x_1, \dots, x_T$ under $k_\theta$ is exactly equal to the log-det quantity in Definition~\ref{def:information-gain} for the sequence $z_1, \dots, z_T$ under $k_0$ on the domain $w_\theta(X)$.

Taking the supremum over all sequences $x_1, \dots, x_T \in X$ is thus equivalent to taking the supremum over all sequences $z_1, \dots, z_T \in w_\theta(X)$, which yields
\begin{equation*}
    \gamma_T(k_\theta, X) = \gamma_T\left(k_0, w_\theta(X) \right)
\end{equation*}

Since $w_\theta(X) \subseteq \mathcal{X}$, the supremum over sequences in $w_\theta(X)$ is bounded by the supremum over sequences in $\mathcal{X}$, and hence
\begin{equation*}
    \gamma_T\left(k_0, w_\theta(X)\right) \le \gamma_T\left(k_0, \mathcal{X}\right)
\end{equation*}
\end{proof}

\subsection{Uniform Confidence Bound}

\restatable{%
\begin{lemma}[Fixed-kernel GP-UCB confidence; \citet{srinivas2010gaussian}]
\label{lem:fixed-kernel-confidence}
Let $k$ be a fixed kernel on $X$, and suppose $f \in \mathcal{H}_k$ with
\begin{equation*}
    \|f\|_{\mathcal{H}_k} \le B
\end{equation*}
Assume $k(x,x)\leq1$, $\mathbb{E}[\epsilon_t\mid\mathcal{F}_{t-1}]=0$, and $|\epsilon_t|\leq\sigma$ almost surely.
Let $\mu_{t-1}$ and $\sigma_{t-1}$ denote the posterior mean and posterior standard deviation under kernel $k$ and regularization $\sigma^2$.
Choose $\beta_t$ large enough, e.g. according to the fixed-kernel GP-UCB confidence theorem for RKHS-bounded functions with failure probability $\delta$ from Srinivas et al.~\cite{srinivas2010gaussian}, i.e.
\begin{equation*}
    \beta_t = 2 B^2 + 300 \cdot \gamma_t(k,X) \log^3 \left( t / \delta \right)
\end{equation*}
Then, with probability at least $1-\delta$,
\begin{equation*}
    |f(x) - \mu_{t-1}(x)| \le \sqrt{\beta_t}\sigma_{t-1}(x)
\end{equation*}
simultaneously for all $t \ge 1$ and all $x \in X$.
\end{lemma}
}

\begin{proof}
This is the GP-UCB confidence event in the bounded-RKHS setting.
See the analysis by Srinivas et al.~\cite{srinivas2010gaussian}.
\end{proof}

\restatable{%
\begin{lemma}[Uniform confidence over the finite library]
\label{lem:finite_library_confidence}
For each $i \in \{1, \dots, N_\varepsilon\}$, define
\begin{equation*}
    k_i := k_{\theta^{(i)}}
\end{equation*}
and let $\mu_{i,t-1}$ and $\sigma_{i,t-1}$ denote the posterior mean and posterior standard deviation obtained from kernel $k_i$ and the common data $\mathcal{D}_{t-1}$.

Define
\begin{equation*}
    \Gamma_t := \gamma_t \left(k_0, \mathcal{X} \right)
\end{equation*}

Choose
\begin{equation*}
    \beta_t = 2 C_{\mathrm{warp}}^2 + 300\cdot \Gamma_t \log^3 \left( \frac{t N_\varepsilon}{\delta} \right)
\end{equation*}

Then, with probability at least $1 - \delta$,
\begin{equation*}
    |f(x)-\mu_{i,t-1}(x)| \le \sqrt{\beta_t}\,\sigma_{i,t-1}(x)
\end{equation*}
simultaneously for all $i \in \{1, \dots, N_\varepsilon\}$, all $t \ge 1$, and all $x \in X$.
\end{lemma}
}

\begin{proof}
Fix $i \in \{1, \dots, N_\varepsilon\}$. 
By Theorem~\ref{thm:finite_library_rkhs_transfer},
\begin{equation*}
    f \in \mathcal{H}_{k_i} \qquad \text{and} \qquad \|f\|_{\mathcal{H}_{k_i}} \le C_{\mathrm{warp}}
\end{equation*}
Moreover, by Lemma~\ref{lem:gamma_warp_transfer}, for every $t \ge 1$,
\begin{equation*}
    \gamma_t(k_i,X) \le \gamma_t \left(k_0, \mathcal{X} \right) =: \Gamma_t
\end{equation*}

The chosen $\beta_t$ upper-bounds confidence parameter required by Lemma~\ref{lem:fixed-kernel-confidence} for all $i$, and therefore the confidence bound remains valid.
Therefore Lemma~\ref{lem:fixed-kernel-confidence} applies to the single kernel $k_i$ with failure probability $\delta / N_\varepsilon$ and the above choice of $\beta_t$, yielding
\begin{equation*}
    |f(x) - \mu_{i,t-1}(x)| \le \sqrt{\beta_t} \sigma_{i,t-1}(x)
\end{equation*}
for all $t \ge 1$ and all $x \in X$, except on an event of probability at most $\delta / N_\varepsilon$.

Taking a union bound over $i = 1, \dots, N_\varepsilon$ proves the claim.
\end{proof}

\subsection{No-Regret Proof}

The following policy is a finite-library analogue of GP-UCB, as presented in \cite{srinivas2010gaussian}.

\begin{definition}[Finite-library warped GP-UCB policy]
\label{def:library-gpucb}
For each round $t \ge 1$ and each $i \in \left\{1, \dots, N_\varepsilon \right\}$, define
\begin{equation*}
    U_{i,t-1}(x) := \mu_{i,t-1}(x) + \sqrt{\beta_t} \sigma_{i,t-1}(x)
\end{equation*}

At round $t$, let
\begin{equation*}
    j_t \in \{1, \dots, N_\varepsilon \}
\end{equation*}
be any index chosen from the common data $\mathcal{D}_{t-1}$ as defined earlier. 
Then choose
\begin{equation*}
    x_t \in \arg\max_{x \in X} U_{j_t,t-1}(x)
\end{equation*}

Finally, we recall the standard definitions of instantaneous ($r_t$) and cumulative ($R_T$) regret
\begin{equation*}
    \begin{aligned}
        x^\star &\in \arg\max_{x \in X} f(x), \\
        r_t &:= f(x^\star) - f(x_t), \\
        R_T &:= \sum_{t=1}^T r_t.
    \end{aligned}
\end{equation*}
\end{definition}

\begin{remark}
The policy of Definition~\ref{def:library-gpucb} includes, as special cases, selectors based on for example marginal likelihood, posterior model probability, or any other score computed from $\mathcal{D}_{t-1}$. 
In particular, using the prior weights $(\pi_i)_{i=1}^{N_\varepsilon}$, a MAP selector is 
\begin{equation*}
    j_t \in \arg\max_{1 \leq i \leq N_\varepsilon} \left\{ \log p(\mathcal{D}_{t-1} \mid \theta^{(i)}) + \log \pi_i \right\}
\end{equation*}
Under a uniform prior, this reduces to maximization of the log marginal likelihood.
\end{remark}


(The main text combines the next two lemmas under one ``Selected-branch instantaneous regret and variance accumulation'' statement.)
\restatable{%
\begin{lemma}[Selected-branch instantaneous regret]
\label{lem:select-branch-inst-regret}
On the event of Lemma~\ref{lem:finite_library_confidence}, for each $t \ge 1$,
\begin{equation*}
    r_t \le 2 \sqrt{\beta_t} \sigma_{j_t,t-1}(x_t)
\end{equation*}
\end{lemma}
}

\begin{proof}
Fix $t \ge 1$.
By Lemma~\ref{lem:finite_library_confidence}, for every $i \in \left\{1, \dots, N_\varepsilon \right\}$ and every $x \in X$,
\begin{equation*}
    f(x) \le \mu_{i,t-1}(x) + \sqrt{\beta_t} \sigma_{i,t-1}(x) = U_{i,t-1}(x)
\end{equation*}

In particular, for the selected index $j_t$,
\begin{equation*}
    f(x^\star) \leq U_{j_t, t-1}(x^\star)
\end{equation*}
Since $x_t$ is chosen to maximize $U_{j_t,t-1}(x)$, we have
\begin{equation*}
    U_{j_t, t-1}(x^\star) \leq U_{j_t, t-1}(x_t) = \mu_{j_t,t-1}(x_t) + \sqrt{\beta_t} \sigma_{j_t,t-1}(x_t)
\end{equation*}
Combining gives
\begin{equation*}
    f(x^\star) \le \mu_{j_t,t-1}(x_t) + \sqrt{\beta_t} \sigma_{j_t,t-1}(x_t)
\end{equation*}

Applying Lemma~\ref{lem:finite_library_confidence} again at $(j_t,x_t)$ yields
\begin{equation*}
    \mu_{j_t,t-1}(x_t) \le f(x_t) + \sqrt{\beta_t}\sigma_{j_t,t-1}(x_t)
\end{equation*}

Hence
\begin{equation*}
    f(x^\star) - f(x_t) \le 2 \sqrt{\beta_t} \sigma_{j_t,t-1}(x_t)
\end{equation*}
which proves the claim.
\end{proof}

\restatable{%
\begin{lemma}[Reduction to selected-branch variance accumulation]
\label{lem:variance-reduction}
Assume $\beta_t$ is non-decreasing.
On the event of Lemma~\ref{lem:finite_library_confidence},
\begin{equation*}
    R_T \le 2 \sqrt{\beta_T} \sum_{t=1}^T \sigma_{j_t,t-1}(x_t)
\end{equation*}
and therefore
\begin{equation*}
    R_T \le 2 \sqrt{\beta_T} \sqrt{T \sum_{t=1}^T \sigma_{j_t,t-1}^2(x_t) }
\end{equation*}
\end{lemma}
}

\begin{proof}
Summing Lemma~\ref{lem:select-branch-inst-regret} over $t$ gives
\begin{equation*}
    R_T = \sum_{t=1}^T r_t \le 2 \sum_{t=1}^T \sqrt{\beta_t} \sigma_{j_t,t-1}(x_t)
\end{equation*}
Since $\beta_t \le \beta_T$ for all $t \le T$,
\begin{equation*}
    R_T \le 2 \sqrt{\beta_T} \sum_{t=1}^T \sigma_{j_t,t-1}(x_t)
\end{equation*}

Applying Cauchy-Schwarz to the variance term yields
\begin{equation*}
    \sum_{t=1}^T \sigma_{j_t,t-1}(x_t) \le \sqrt{T \sum_{t=1}^T \sigma_{j_t,t-1}^2 (x_t)}
\end{equation*}

Substituting this into the previous expression proves the result.
\end{proof}

\restatable{%
\begin{lemma}[Selected-branch variance bound]
\label{lem:selected-branch-variance}
Under the policy of Definition~\ref{def:library-gpucb},
\begin{equation*}
    \sum_{t=1}^T \sigma_{j_t,t-1}^2(x_t) \le C_{\mathrm{info}} \sum_{i=1}^{N_\varepsilon} \gamma_T(k_i,X)
\end{equation*}
\end{lemma}
}

\begin{proof}
For each $i \in \left\{1,\dots,N_\varepsilon\right\}$, define
\begin{equation*}
    J_i := \left\{t \in \left\{1,\dots,T\right\} : j_t = i\right\}
\end{equation*}
Then $\left\{J_i\right\}_{i=1}^{N_\varepsilon}$ forms a partition of $\left\{1,\dots,T\right\}$, so
\begin{equation*}
    \sum_{t=1}^T \sigma_{j_t,t-1}^2(x_t) 
    = \sum_{i=1}^{N_\varepsilon} \sum_{t \in J_i} \sigma_{i,t-1}^2(x_t)
    \le \sum_{i=1}^{N_\varepsilon} \sum_{t=1}^T \sigma_{i,t-1}^2(x_t)
\end{equation*}

Now fix $i$.
Since each $k_i$ is a fixed kernel, Lemma~\ref{lem:fixed-kernel-variance} gives
\begin{equation*}
    \sum_{t=1}^T \sigma_{i,t-1}^2(x_t) \le C_{\mathrm{info}}\gamma_T(k_i,X)
\end{equation*}
Summing over $i$ proves the claim.
\end{proof}

\restatable{%
\begin{theorem}[FLIWBO-UCB sublinear cumulative regret]
\label{thm:library-no-regret}
\label{thm:main-no-regret}
Under Assumptions~\ref{ass:regular_warp}, \ref{ass:rkhs}, and \ref{ass:finite_net_base_rkhs} and the policy of Definition~\ref{def:library-gpucb}, using $\beta_t$ as specified in Lemma~\ref{lem:finite_library_confidence}, then with probability at least $1 - \delta$,
\begin{equation*}
    R_T \le 2 \sqrt{C_{\mathrm{info}} N_\varepsilon T \beta_T \Gamma_T}
\end{equation*}
Where
\begin{equation*}
    \Gamma_T := \gamma_T(k_0, \mathcal{X})
\end{equation*}
In particular, if
\begin{equation*}
    \beta_T \Gamma_T = o(T)
\end{equation*}
then
\begin{equation*}
    R_T = o(T) \qquad \text{and hence} \qquad \frac{R_T}{T} \underset{T \to \infty}{\longrightarrow} 0
\end{equation*}
\end{theorem}
}

\begin{proof}
Let $\mathcal{E}$ denote the event of Lemma~\ref{lem:finite_library_confidence}.
Then $\mathbb{P}(\mathcal{E}) \ge 1 - \delta$.
From here, we work on $\mathcal{E}$.
By Lemma~\ref{lem:variance-reduction} and Lemma~\ref{lem:selected-branch-variance} combined with Lemma~\ref{lem:gamma_warp_transfer},
\begin{equation*}
    R_T \le 2 \sqrt{ T \beta_T C_{\mathrm{info}} N_\varepsilon \gamma_T(k_0, \mathcal{X}) }
\end{equation*}
Using $\Gamma_T = \gamma_T(k_0, \mathcal{X})$ as in Lemma~\ref{lem:finite_library_confidence} gives the stated regret bound.

If
\begin{equation*}
    \beta_T \Gamma_T = o(T)
\end{equation*}
then the right-hand side is $o(T)$, so $R_T/T \underset{T \to \infty}{\longrightarrow} 0$.
\end{proof}


\begin{remark}[Selector-adaptive variance refinement]
A crude bound ``charges'' every branch for all $T$ rounds,
$\sum_{t=1}^T\sigma_{j_t,t-1}^2(x_t)
\leq
\sum_{i=1}^{N_\varepsilon}\sum_{t=1}^T\sigma_{i,t-1}^2(x_t)
\leq
C_{\mathrm{info}}N_\varepsilon\Gamma_T$,
even on rounds when that branch is not selected.
A tighter bound uses only the rounds on which that branch is selected.
For each $i\in\{1,\ldots,N_\varepsilon\}$, let
$J_i\triangleq\{t\in\{1,\ldots,T\}:j_t=i\}$ and $n_i(T)\triangleq|J_i|$.
Because each branch-$i$ posterior is conditioned on the full common history,
its predictive variance at any $t\in J_i$ is at most the variance obtained by
conditioning only on the earlier points in $J_i$.
Applying Lemma~\ref{lem:fixed-kernel-variance} to each subsequence $(x_t)_{t\in J_i}$ therefore yields
$
\sum_{t=1}^T\sigma_{j_t,t-1}^2(x_t)
\leq
C_{\mathrm{info}}\sum_{i=1}^{N_\varepsilon}\gamma_{n_i(T)}(k_i,X)
$
(with the convention $\gamma_0\equiv0$), and hence by Lemma~\ref{lem:variance-reduction}
\[
R_T
\leq
2\sqrt{C_{\mathrm{info}}T\beta_T\sum_{i=1}^{N_\varepsilon}\gamma_{n_i(T)}(k_i,X)}.
\]
Since $n_i(T)\le T$ and $\gamma_{n_i(T)}(k_i,X)\le\Gamma_T$, this recovers the
uniform $N_\varepsilon\Gamma_T$ bound of Theorem~\ref{thm:library-no-regret}.
When the selector concentrates on few branches the refined bound can be much
smaller, but the worst-case no-regret rate is unchanged.
\end{remark}

\paragraph{Probability-weighted acquisition averaging.}
The preceding analysis extends directly to policies that average the
library acquisition functions rather than selecting a single branch.

\begin{definition}[Probability-weighted finite-library warped GP-UCB policy]
\label{def:probability-weighted-library-gpucb}
For each round $t \geq 1$, let
\[
q_t = (q_{t,1},\ldots,q_{t,N_\varepsilon})
\]
be an $\mathcal{F}_{t-1}$-measurable probability vector, so that
\[
q_{t,i} \geq 0,
\qquad
\sum_{i=1}^{N_\varepsilon} q_{t,i} = 1.
\]
Define the probability-weighted posterior mean
\[
\bar{\mu}_{t-1}(x)
:=
\sum_{i=1}^{N_\varepsilon}
q_{t,i}\mu_{i,t-1}(x),
\]
and the probability-weighted acquisition function
\[
\begin{aligned}
\bar{U}_{t-1}(x)
&:=
\sum_{i=1}^{N_\varepsilon}
q_{t,i}U_{i,t-1}(x) \\
&=
\bar{\mu}_{t-1}(x)
+
\sqrt{\beta_t}
\sum_{i=1}^{N_\varepsilon}
q_{t,i}\sigma_{i,t-1}(x).
\end{aligned}
\]
At round $t$, choose
\[
x_t \in
\arg\max_{x\in\mathcal{X}}
\bar{U}_{t-1}(x).
\]
\end{definition}

\begin{theorem}[No-regret for probability-weighted finite-library
input-warped GP-UCB]
\label{thm:probability-weighted-library-no-regret}
Under Assumptions 2, 3, and 7 and the probability-weighted policy above,
using $\beta_t$ as specified in Lemma 15, with probability at least
$1-\delta$,
\[
R_T
\leq
2\sqrt{
C_{\mathrm{info}}N_\varepsilon
T\beta_T\Gamma_T
},
\]
where
\[
\Gamma_T := \gamma_T(k_0,\mathcal{X}).
\]
In particular, if
\[
\beta_T\Gamma_T=o(T),
\]
then
\[
R_T=o(T)
\qquad\text{and hence}\qquad
\frac{R_T}{T}\longrightarrow 0
\quad\text{as }T\to\infty.
\]
\end{theorem}

\begin{proof}
Let $\mathcal{E}$ denote the simultaneous confidence event of Lemma 15,
so that $\mathbb{P}(\mathcal{E})\geq 1-\delta$. We work on
$\mathcal{E}$.

For every $t\geq 1$ and $x\in\mathcal{X}$,
\[
\begin{aligned}
\left|f(x)-\bar{\mu}_{t-1}(x)\right|
&=
\left|
\sum_{i=1}^{N_\varepsilon}
q_{t,i}
\bigl(f(x)-\mu_{i,t-1}(x)\bigr)
\right| \\
&\leq
\sum_{i=1}^{N_\varepsilon}
q_{t,i}
\left|f(x)-\mu_{i,t-1}(x)\right| \\
&\leq
\sqrt{\beta_t}
\sum_{i=1}^{N_\varepsilon}
q_{t,i}\sigma_{i,t-1}(x).
\end{aligned}
\]
Consequently,
\[
f(x)\leq \bar{U}_{t-1}(x)
\]
for every $x\in\mathcal{X}$. Since $x_t$ maximizes
$\bar{U}_{t-1}$,
\[
f(x^\star)
\leq
\bar{U}_{t-1}(x^\star)
\leq
\bar{U}_{t-1}(x_t).
\]
Applying the preceding confidence inequality once more at $x_t$ gives
\[
r_t
=
f(x^\star)-f(x_t)
\leq
2\sqrt{\beta_t}
\sum_{i=1}^{N_\varepsilon}
q_{t,i}\sigma_{i,t-1}(x_t).
\]

Assuming $\beta_t$ is non-decreasing and summing over $t$,
\[
R_T
\leq
2\sqrt{\beta_T}
\sum_{t=1}^{T}
\sum_{i=1}^{N_\varepsilon}
q_{t,i}\sigma_{i,t-1}(x_t).
\]
By Cauchy--Schwarz,
\[
R_T
\leq
2\sqrt{
T\beta_T
\sum_{t=1}^{T}
\left(
\sum_{i=1}^{N_\varepsilon}
q_{t,i}\sigma_{i,t-1}(x_t)
\right)^2
}.
\]
Since $z\mapsto z^2$ is convex, Jensen's inequality yields
\[
\left(
\sum_{i=1}^{N_\varepsilon}
q_{t,i}\sigma_{i,t-1}(x_t)
\right)^2
\leq
\sum_{i=1}^{N_\varepsilon}
q_{t,i}\sigma_{i,t-1}^2(x_t).
\]
Therefore,
\[
R_T
\leq
2\sqrt{
T\beta_T
\sum_{t=1}^{T}
\sum_{i=1}^{N_\varepsilon}
q_{t,i}\sigma_{i,t-1}^2(x_t)
}.
\]
Using $q_{t,i}\leq 1$ and then Lemma 12 for each fixed kernel $k_i$,
\[
\begin{aligned}
\sum_{t=1}^{T}
\sum_{i=1}^{N_\varepsilon}
q_{t,i}\sigma_{i,t-1}^2(x_t)
&\leq
\sum_{i=1}^{N_\varepsilon}
\sum_{t=1}^{T}
\sigma_{i,t-1}^2(x_t) \\
&\leq
C_{\mathrm{info}}
\sum_{i=1}^{N_\varepsilon}
\gamma_T(k_i,\mathcal{X}).
\end{aligned}
\]
Lemma 13 gives
\[
\gamma_T(k_i,\mathcal{X})
\leq
\gamma_T(k_0,\mathcal{X})
=
\Gamma_T
\]
for every $i$. Hence
\[
R_T
\leq
2\sqrt{
C_{\mathrm{info}}N_\varepsilon
T\beta_T\Gamma_T
}.
\]
If $\beta_T\Gamma_T=o(T)$, dividing the preceding bound by $T$
shows that $R_T/T\to 0$.
\end{proof}

\begin{remark}
No additional confidence penalty is required for the probability
weights. Lemma 15 holds simultaneously over every library member,
round, and point in $\mathcal{X}$; conditional on this event, the
weighted inequalities above hold for any
$\mathcal{F}_{t-1}$-measurable probability vector $q_t$.

The selective policy of Definition 16 is recovered by taking
\[
q_{t,j_t}=1,
\qquad
q_{t,i}=0 \quad\text{for }i\neq j_t.
\]
Thus, probability-weighted acquisition averaging strictly generalizes
the selected-branch policy while retaining the same uniform
no-regret bound.
\end{remark}

\section{Controlled Mechanism Diagnostics}
\label{app:controlled-diagnostics}

This section expands the compact diagnostic summary in the main paper.
All four comparisons use matched initial observations, noise, base kernels, acquisition rules, and evaluation budgets; only the input geometry differs.
Each problem is one constructed trajectory, so the figures diagnose behavior rather than estimate average performance.
For each problem, we show the objective, the final fixed and warped surrogate fits, the same fit in learned coordinates, and the complete instantaneous-regret trajectory.

\subsection{Exact Objectives and Shared Protocol}

Each problem $P_i$ optimizes an objective $f_i$ on $[0.01,0.99]^D$.
Define the common bump function
\begin{equation*}
 q(u,c,s)=\exp[-\tfrac12((u-c)/s)^2].
\end{equation*}
For P2, define the latent objective
\begin{align*}
 g(z)&=0.50+0.08z+0.055\sin(2\pi z+0.2)\\
     &\quad+0.035\sin(8\pi z-0.7)\\
     &\quad+0.16q(z,0.36,0.11)-0.09q(z,0.60,0.07)\\
     &\quad+0.24q(z,0.86,0.035).
\end{align*}
The four diagnostic objectives are
\begin{align*}
 f_1(x)&=0.58+0.52q(x,0.82,0.006),\\
 f_2(x)&=g\bigl(\BetaCDF(x,25.093,8.073)\bigr),\\
 f_3(x)&=0.45+0.30q(x_1,0.28,0.05)\\
       &\quad+0.50q(x_2,0.78,0.02),\\
 f_4(x)&=0.5+0.2\sin(2\pi x)+0.1\cos(4\pi x).
\end{align*}
The one-dimensional tasks P1, P2, and P4 use a Mat\'ern-$5/2$ base kernel, five fixed initial points, a 256-warp library, and 25 sequential BO evaluations.
P3 uses a Mat\'ern-$3$ base kernel, five initial points, a 1296-warp library, and 50 sequential BO evaluations.
We report instantaneous regret against a dense-grid optimum.

\subsection{P1: Narrow High-Value Region}

P1 is nearly flat except for a very narrow peak near the maximizer.
It tests whether a selected monotone warp can expand the consequential part of the domain enough for a stationary base kernel to resolve it.

\begin{figure}[t]
    \centering
    \includegraphics[width=\columnwidth]{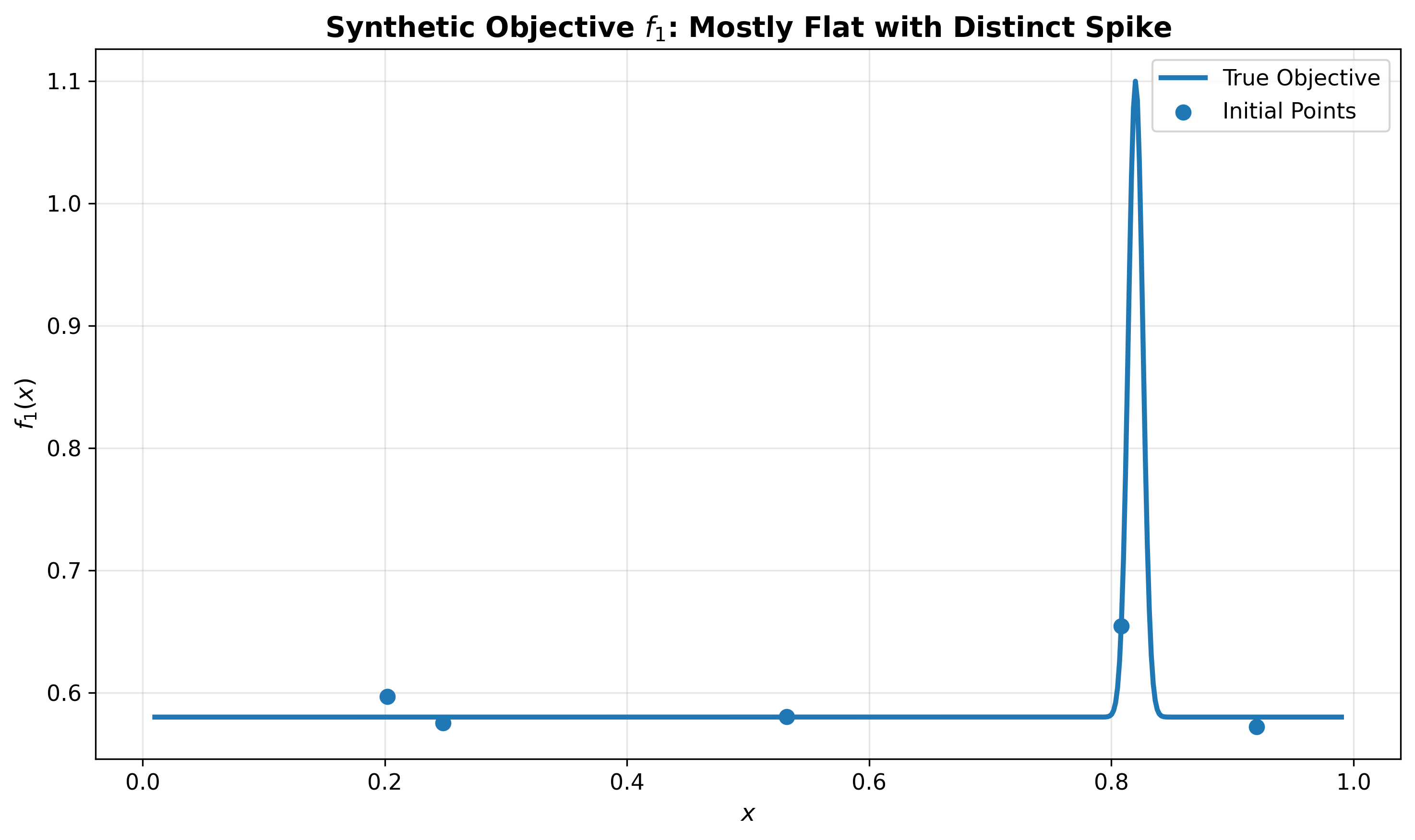}
    \caption{P1 objective and shared initial design.
    The objective's useful variation is confined to a narrow peak on an otherwise nearly constant background.
    A stationary kernel in the raw coordinate must model both scales with one notion of distance.}
    \label{fig:supp-p1-objective}
\end{figure}

\begin{figure*}[p]
    \centering
    \includegraphics[width=\textwidth]{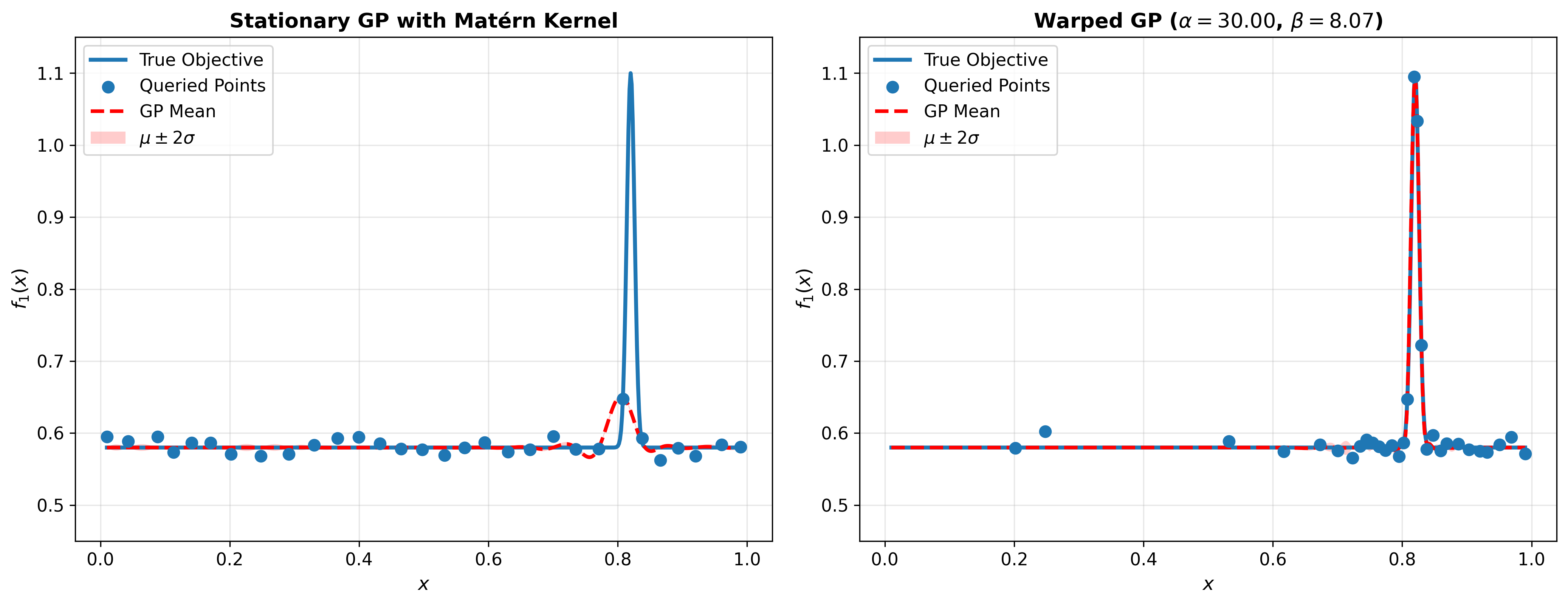}
    \caption{P1 final fits in the original coordinate.
    The fixed model (left) spreads its queries across the flat background and smooths the spike into a modest bump.
    FLIWBO-UCB (right) selects $(\alpha,\beta)=(30.00,8.07)$, concentrates observations around the peak, and recovers its height.
    The warp changes the geometry seen by the kernel; it does not add observations.}
    \label{fig:supp-p1-fits}
\end{figure*}

\begin{figure*}[p]
    \centering
    \includegraphics[width=\textwidth]{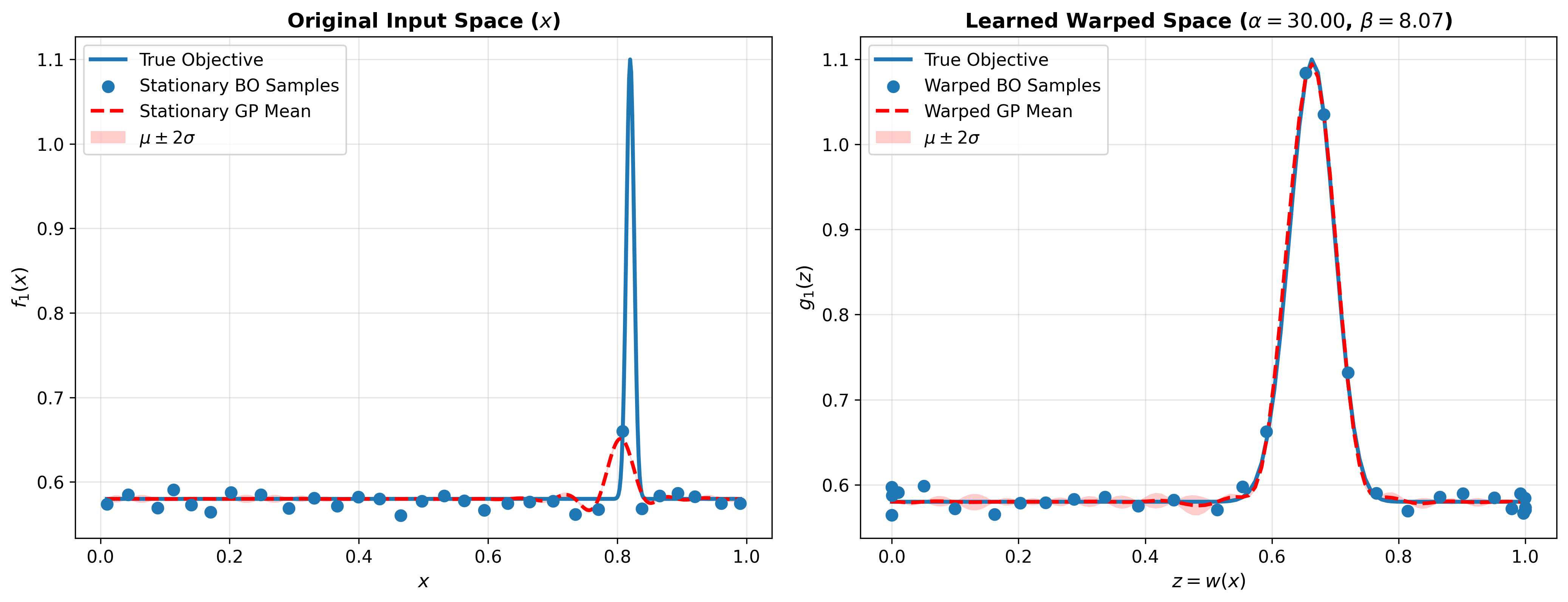}
    \caption{P1 before and after the learned coordinate change.
    In raw $x$-space (left), the high-value region is easy to smooth away.
    In learned $z=w(x)$ coordinates (right), that neighborhood occupies substantially more modeling resolution while the uninformative background is compressed.}
    \label{fig:supp-p1-warped-space}
\end{figure*}

\begin{figure*}[p]
    \centering
    \includegraphics[width=0.76\textwidth]{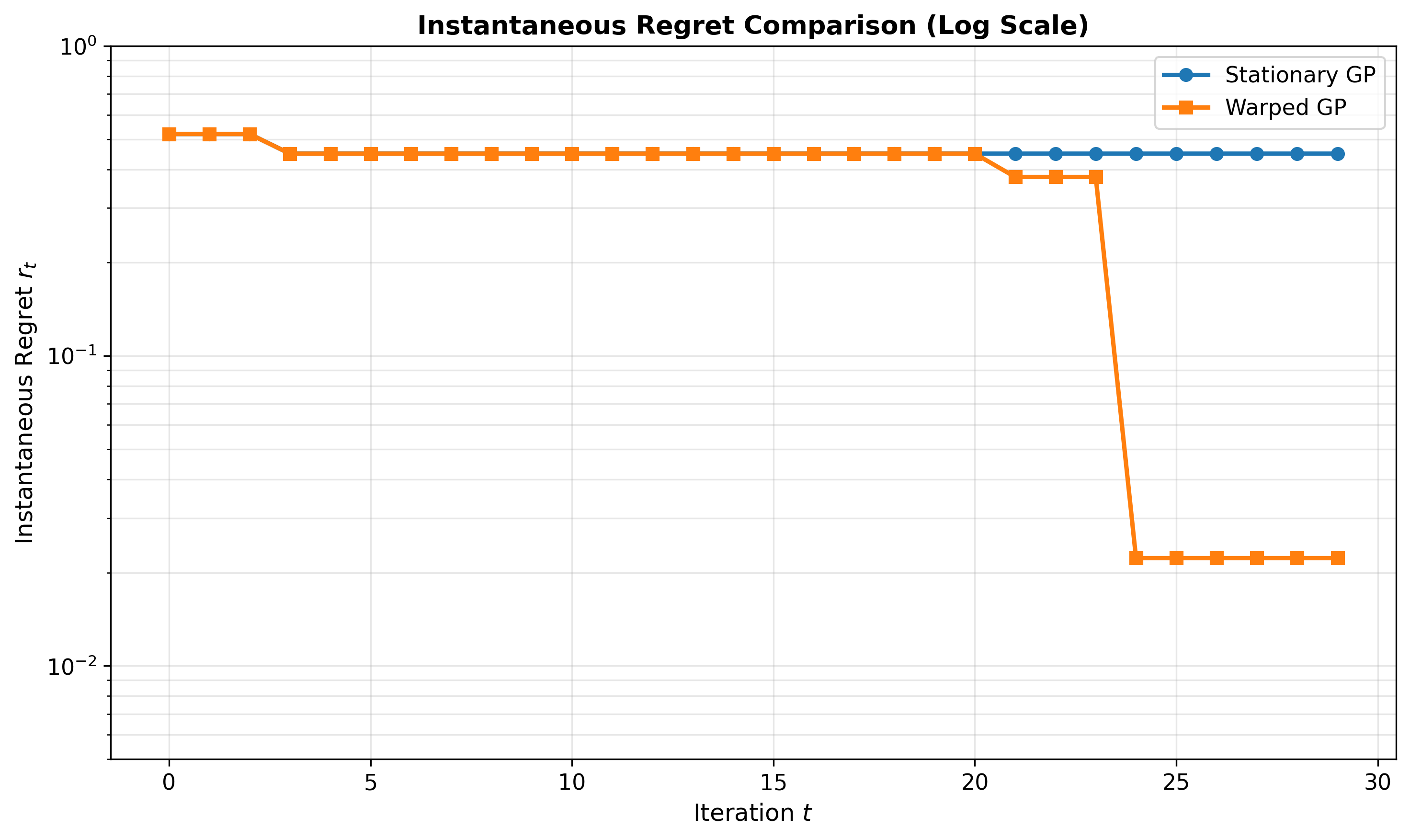}
    \caption{P1 instantaneous regret.
    The two methods initially behave similarly, but the warped trajectory drops once it resolves the narrow optimum; the fixed model continues querying the broad background.
    Lower is better and the vertical axis is logarithmic.}
    \label{fig:supp-p1-regret}
\end{figure*}

\subsection{P2: Objective Generated by a Latent Warp}

P2 is generated from a smooth latent objective through a known Beta-CDF transformation with $(\alpha^\star,\beta^\star)=(25.093,8.073)$.
It is the cleanest recovery diagnostic: useful selection should move toward the planted coordinate system, although the finite library need not contain the exact generator.

\begin{figure}[t]
    \centering
    \includegraphics[width=\columnwidth]{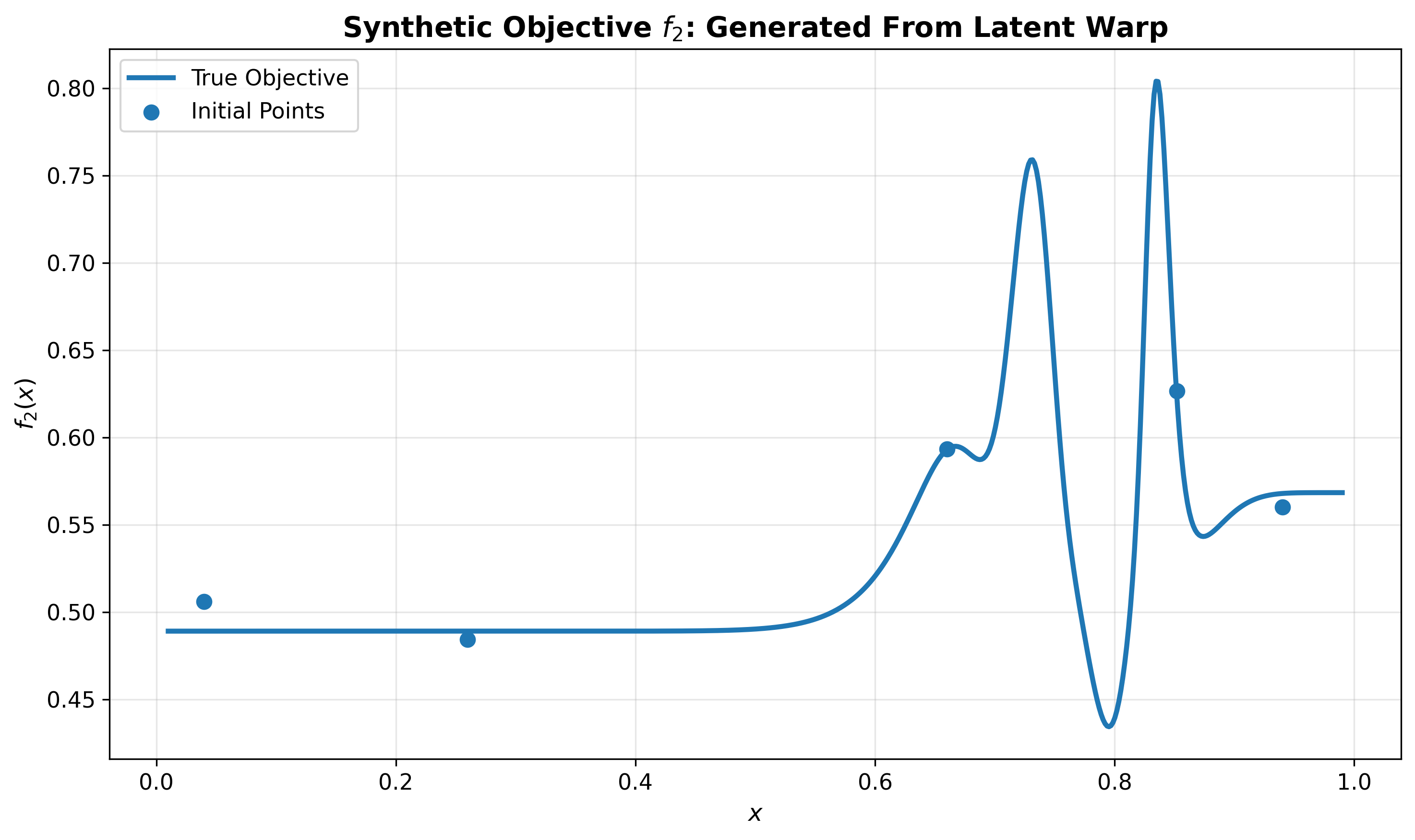}
    \caption{P2 objective and shared initial design.
    A smooth latent function becomes strongly non-uniform when expressed in the observed coordinate, concentrating most visible structure toward one side of the domain.}
    \label{fig:supp-p2-objective}
\end{figure}

\begin{figure*}[p]
    \centering
    \includegraphics[width=\textwidth]{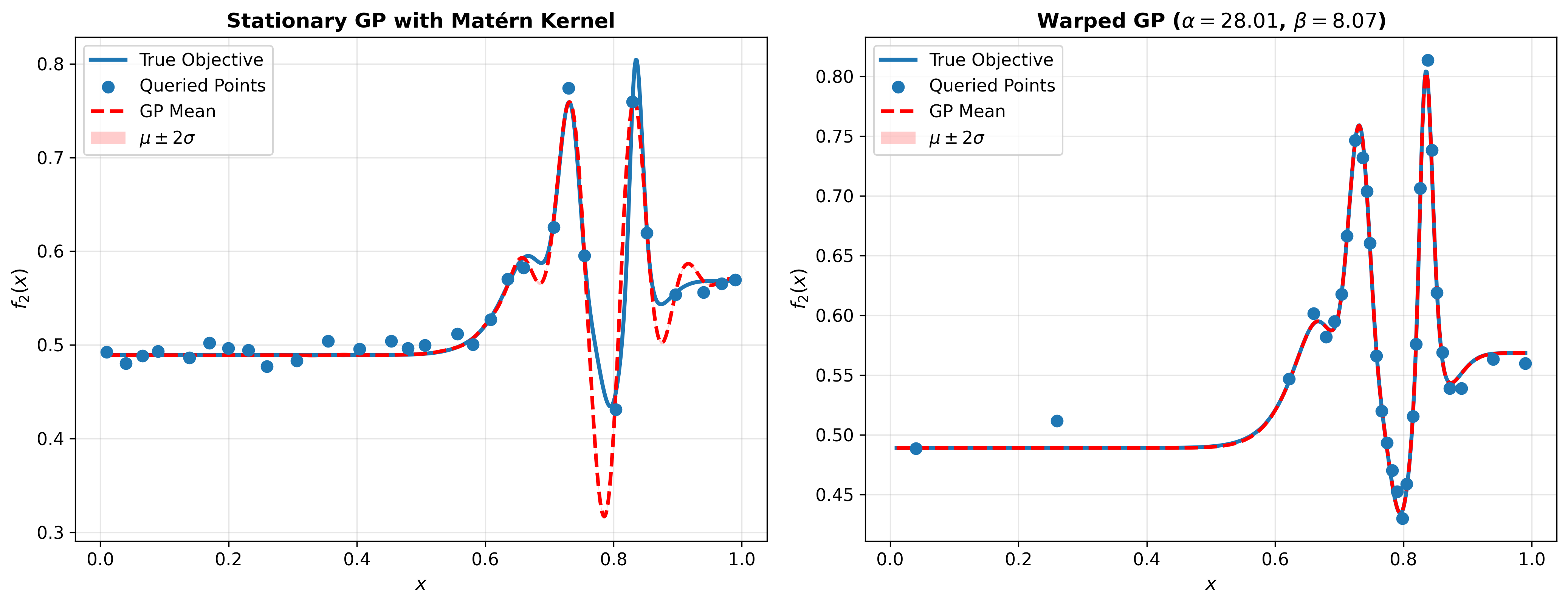}
    \caption{P2 final fits in the original coordinate.
    The fixed model (left) allocates observations broadly.
    FLIWBO-UCB (right) selects $(\alpha,\beta)=(28.01,8.07)$, close to the planted warp, and concentrates resolution where the objective changes most rapidly.}
    \label{fig:supp-p2-fits}
\end{figure*}

\begin{figure*}[p]
    \centering
    \includegraphics[width=\textwidth]{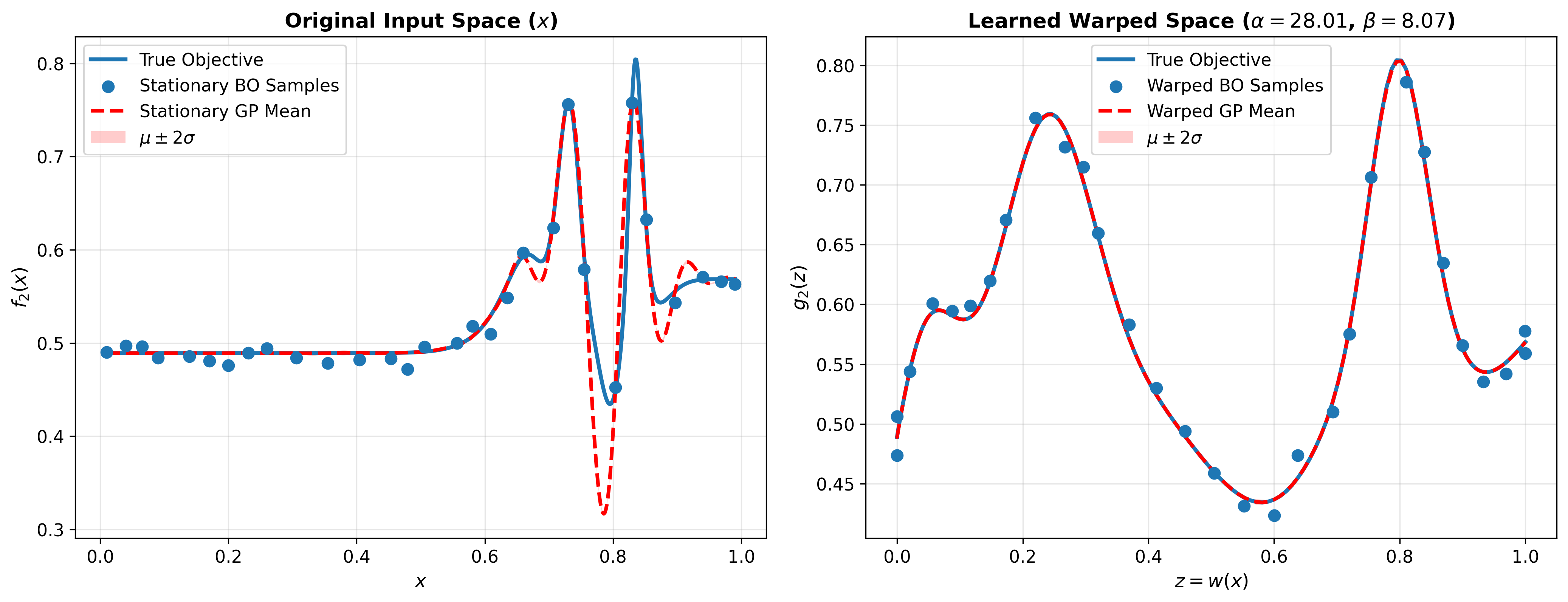}
    \caption{P2 before and after the learned coordinate change.
    The raw-coordinate fit (left) must represent compressed variation near the right boundary.
    In learned coordinates (right), the same structure is spread over the domain and is better matched to the stationary base kernel.}
    \label{fig:supp-p2-warped-space}
\end{figure*}

\begin{figure*}[p]
    \centering
    \includegraphics[width=0.76\textwidth]{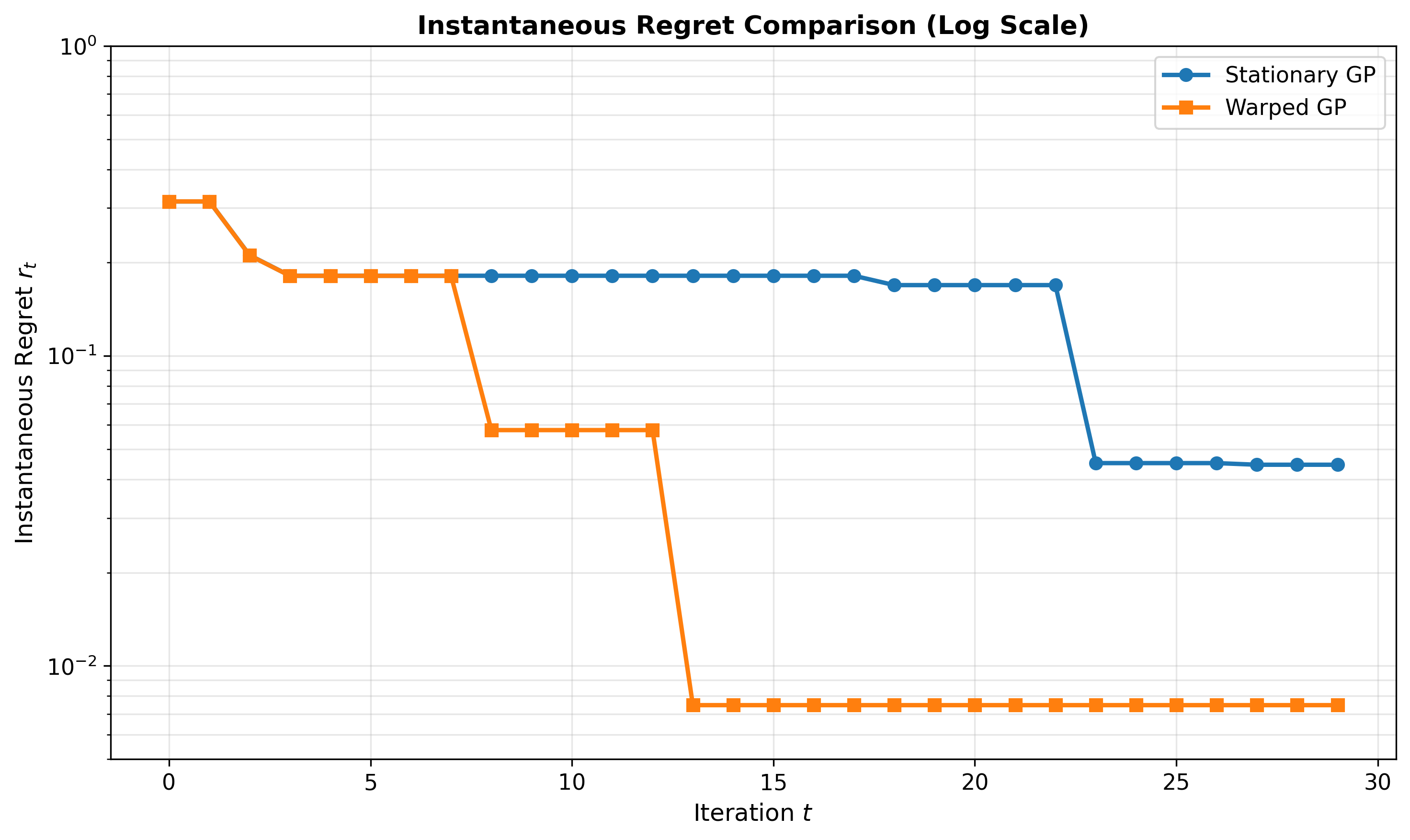}
    \caption{P2 instantaneous regret.
    The warped trajectory makes its decisive improvements earlier and finishes below the fixed model.
    Together with the recovered parameters, the curve connects the optimization gain to the planted latent geometry.
    Lower is better and the vertical axis is logarithmic.}
    \label{fig:supp-p2-regret}
\end{figure*}

\subsection{P3: Coordinate-Specific Feature Widths}

P3 extends the mechanism check to two dimensions.
Its additive features vary at different locations and widths along the two coordinates, allowing coordinate-wise warps to allocate resolution separately in each dimension.

\begin{figure}[t]
    \centering
    \includegraphics[width=\columnwidth]{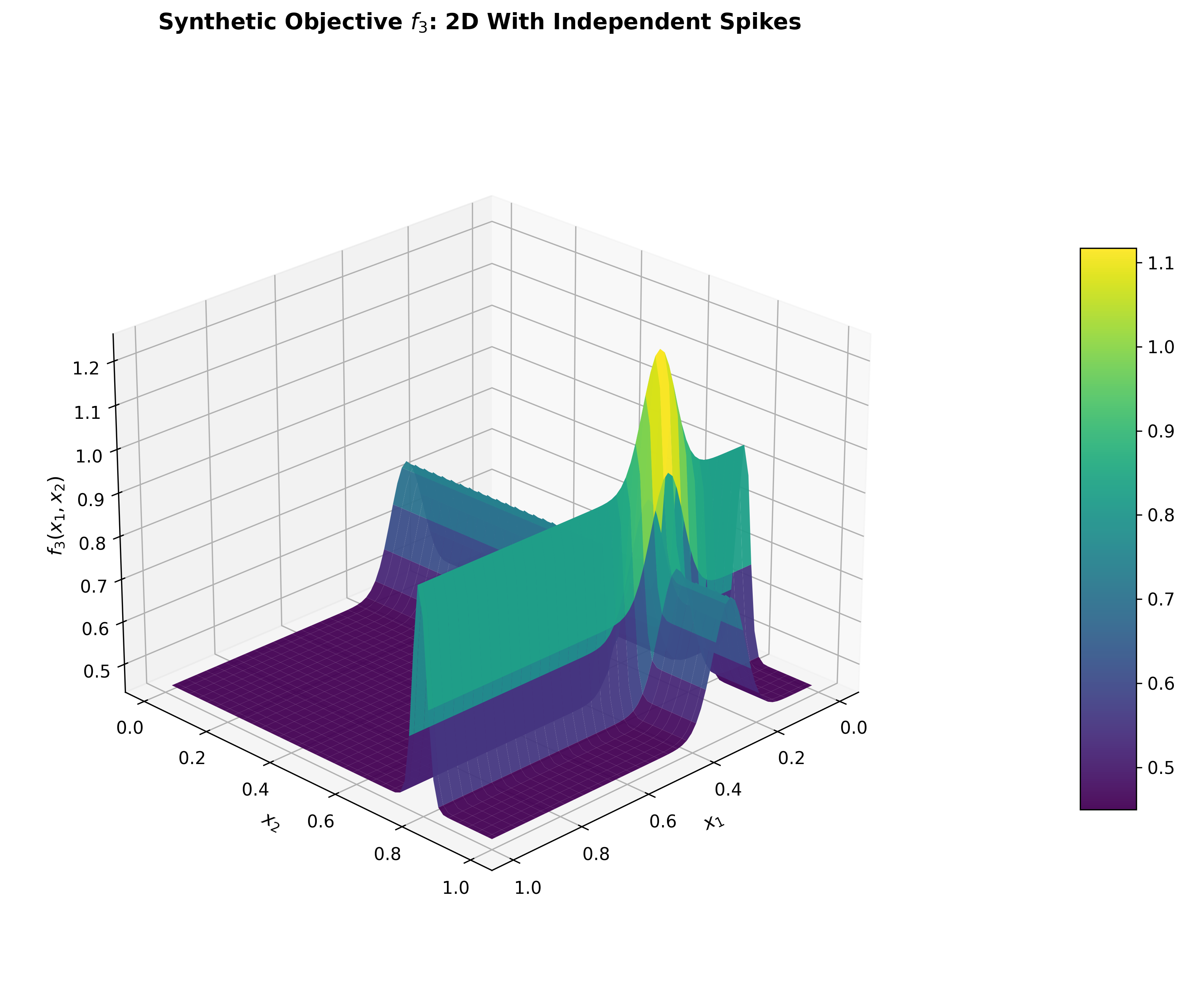}
    \caption{P3 objective.
    Two narrow coordinate-aligned structures intersect at the maximizer.
    The initial design is omitted from this surface to keep the different feature scales visible.}
    \label{fig:supp-p3-objective}
\end{figure}

\begin{figure*}[p]
    \centering
    \includegraphics[width=\textwidth]{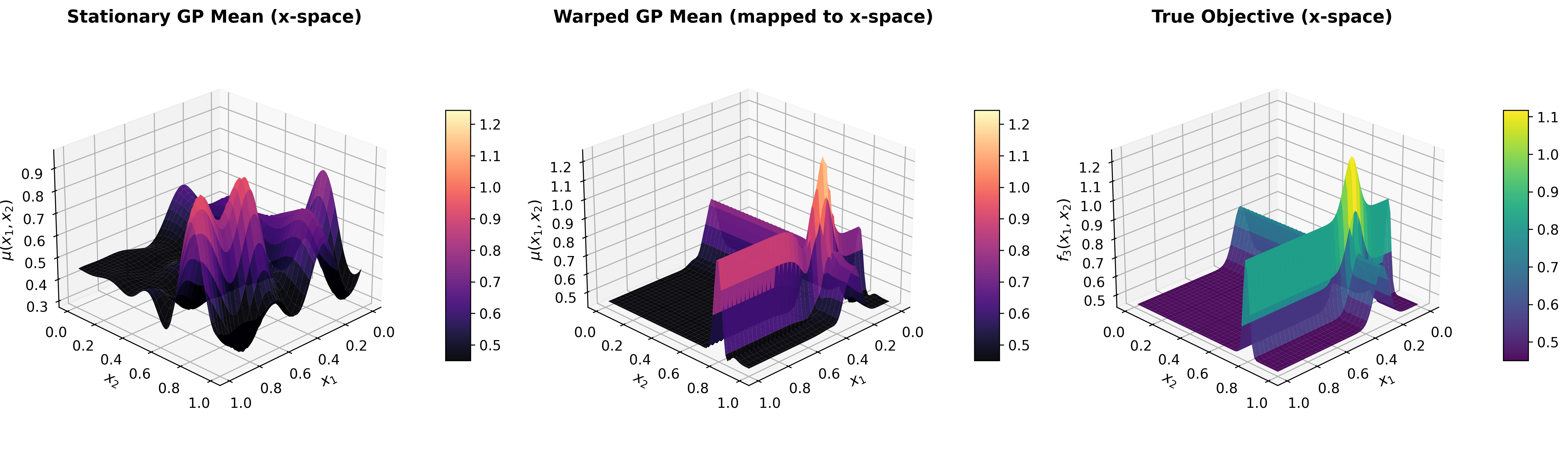}
    \caption{P3 final surfaces in the original coordinates.
    From left to right: the fixed GP posterior mean, the warped GP posterior mean mapped back to $x$-space, and the true objective.
    The fixed fit varies across much of the domain, whereas the warped fit concentrates its structure around the narrow ridges that matter.}
    \label{fig:supp-p3-fits}
\end{figure*}

\begin{figure*}[p]
    \centering
    \includegraphics[width=0.92\textwidth]{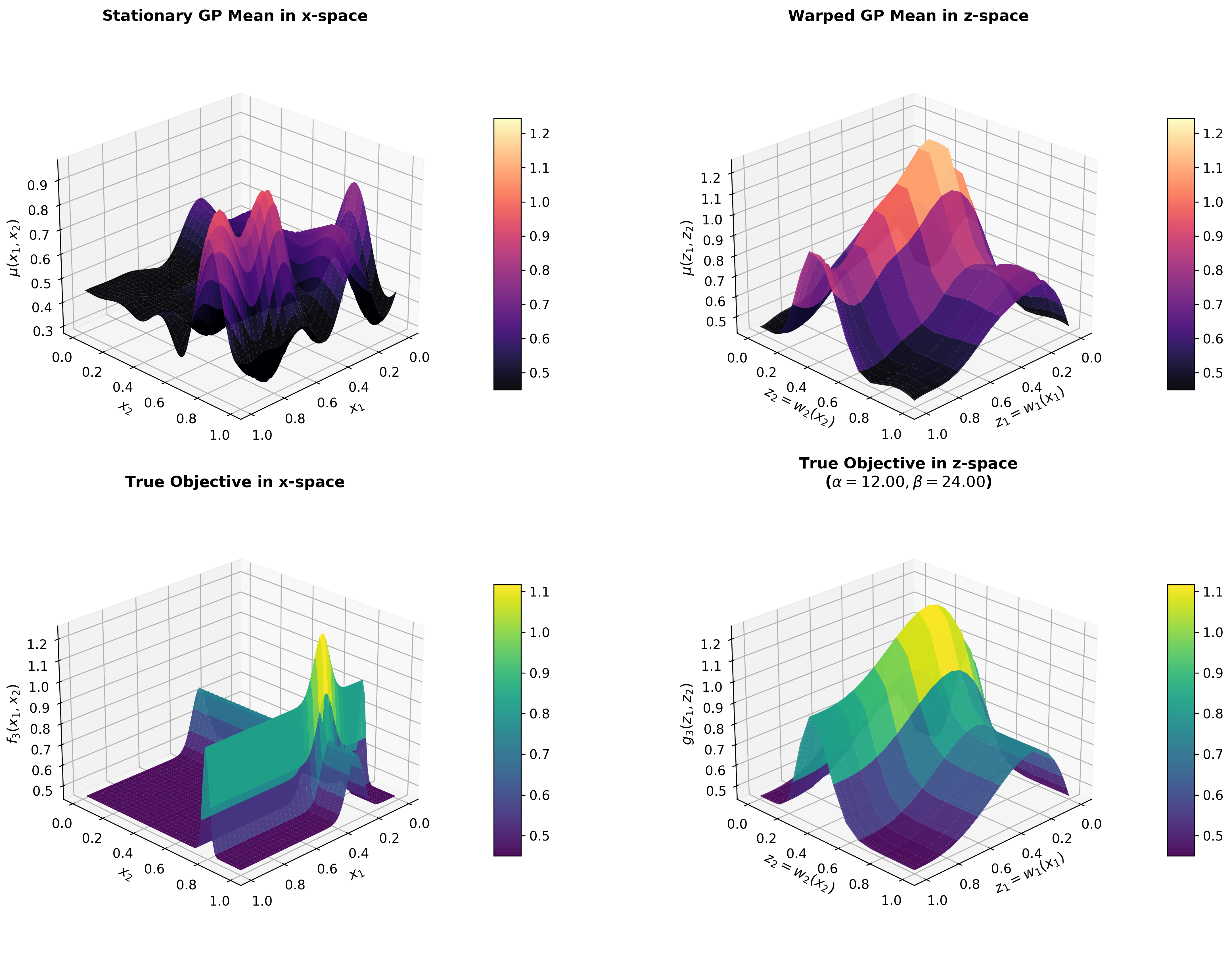}
    \caption{P3 in original and learned coordinates.
    The upper row compares the fixed posterior in $x$-space with the warped posterior in $z$-space; the lower row shows the true objective in the corresponding coordinates.
    The learned representation broadens the high-value structures seen by the stationary base kernel.}
    \label{fig:supp-p3-warped-space}
\end{figure*}

\begin{figure*}[p]
    \centering
    \includegraphics[width=0.76\textwidth]{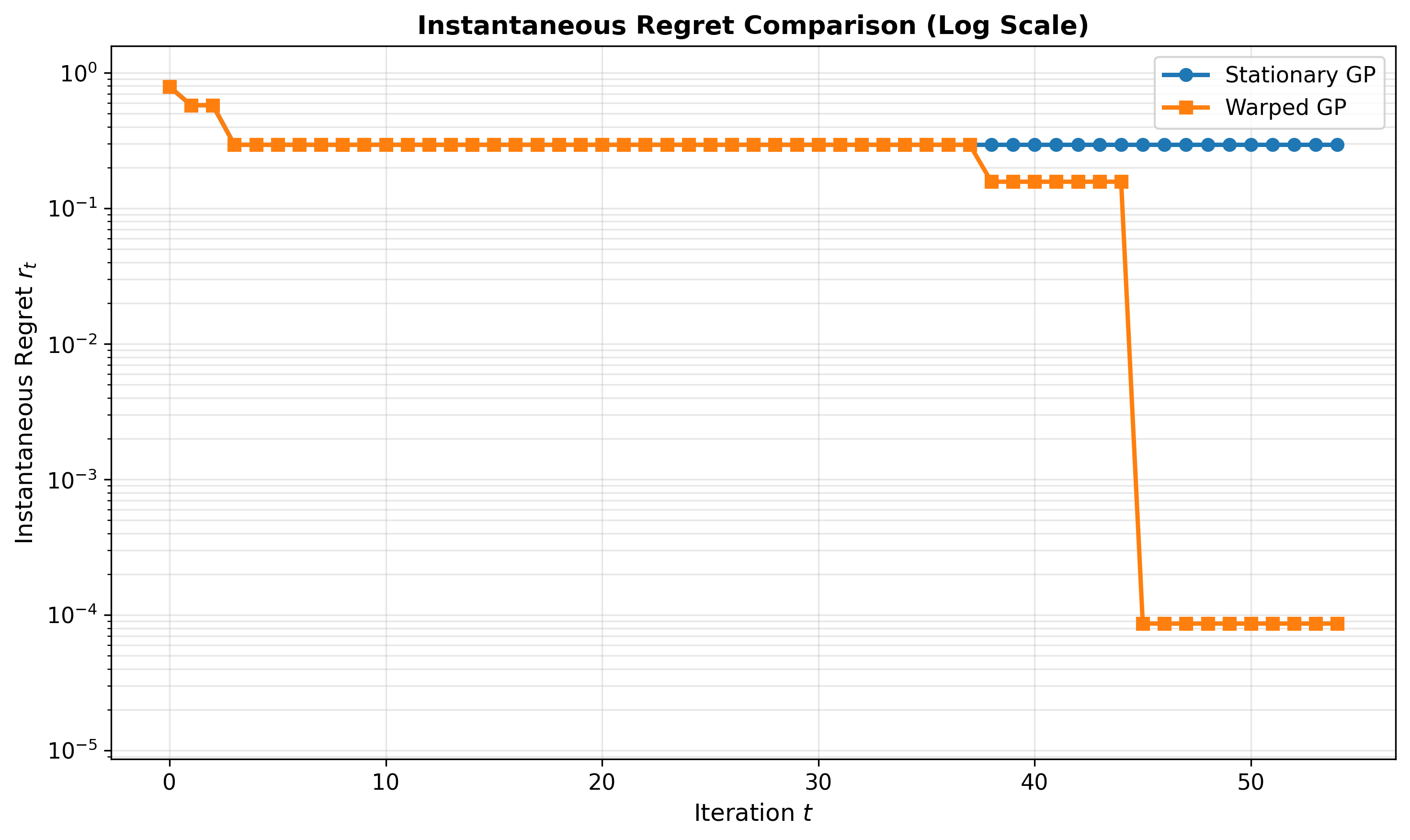}
    \caption{P3 instantaneous regret.
    The fixed trajectory remains on the same plateau, while FLIWBO-UCB makes several improvements and ultimately resolves the maximizer near the end of the budget.
    Lower is better and the vertical axis is logarithmic.}
    \label{fig:supp-p3-regret}
\end{figure*}

\subsection{P4: Stationary Control}

P4 is smooth in the observed coordinate and requires no latent transformation.
It tests the cost of adapting the geometry when the stationary base model is already appropriate.

\begin{figure}[t]
    \centering
    \includegraphics[width=\columnwidth]{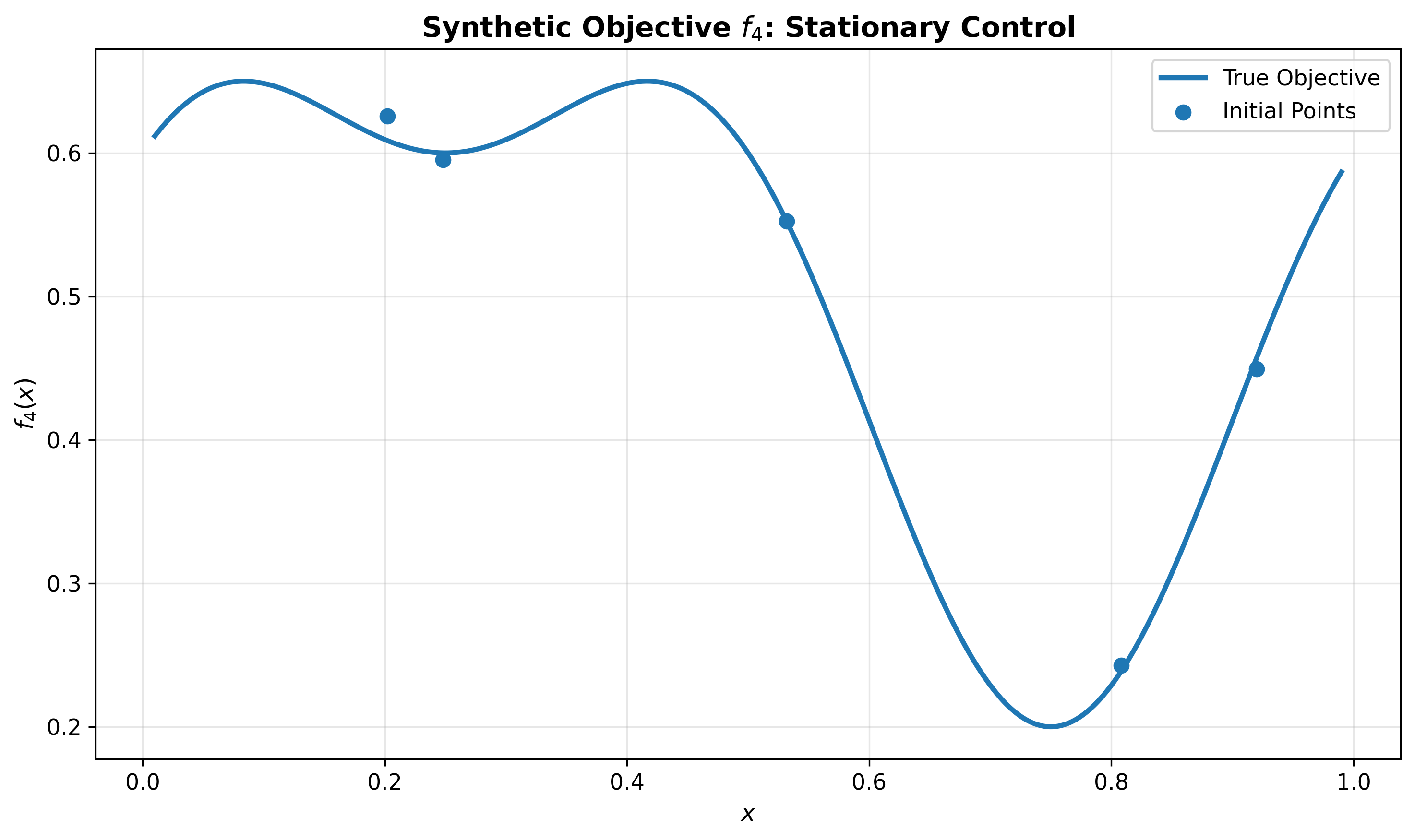}
    \caption{P4 objective and shared initial design.
    Unlike P1--P3, the objective varies smoothly over the raw coordinate and presents no narrow hidden region for a warp to reveal.}
    \label{fig:supp-p4-objective}
\end{figure}

\begin{figure*}[p]
    \centering
    \includegraphics[width=\textwidth]{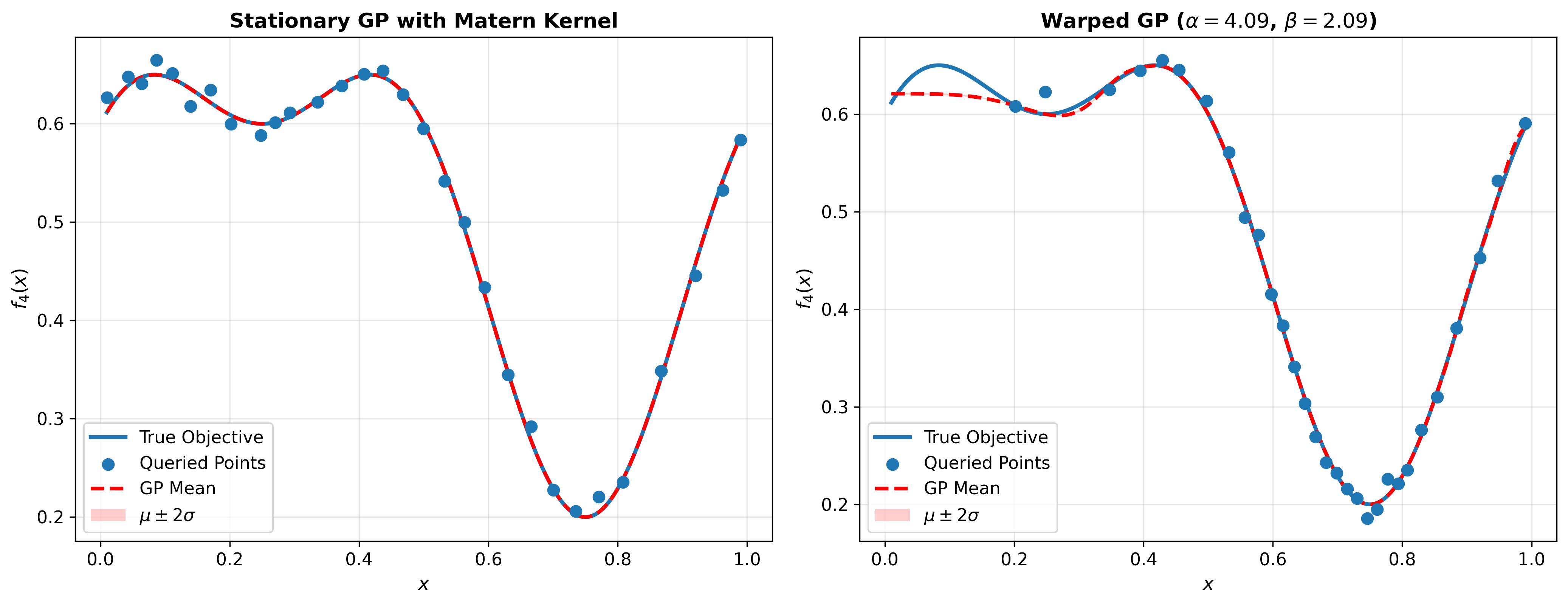}
    \caption{P4 final fits in the original coordinate.
    Both surrogates fit the objective well, but FLIWBO-UCB (right) selects the unnecessary non-identity warp $(\alpha,\beta)=(4.09,2.09)$.
    The added flexibility changes where queries are spent without repairing a genuine mismatch.}
    \label{fig:supp-p4-fits}
\end{figure*}

\begin{figure*}[p]
    \centering
    \includegraphics[width=\textwidth]{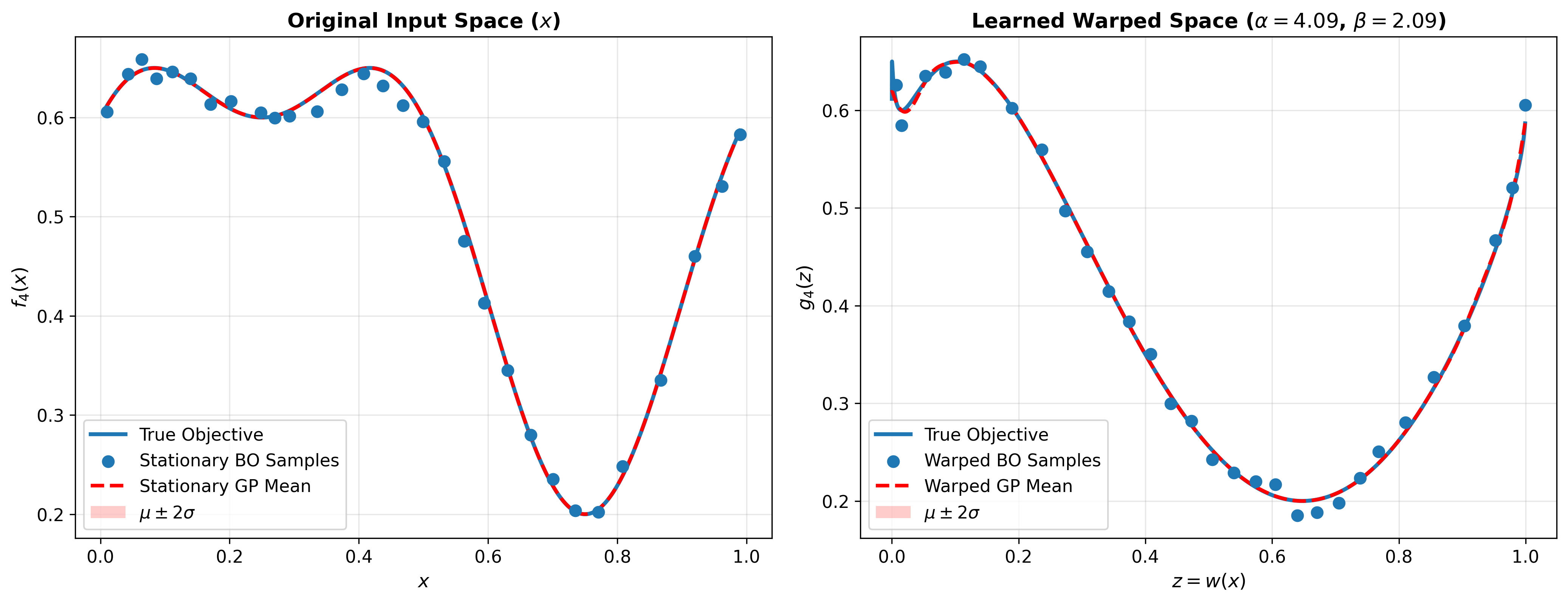}
    \caption{P4 before and after the selected coordinate change.
    The original coordinate (left) already presents the objective at a scale well matched to the base kernel.
    The selected warped coordinate (right) redistributes resolution but supplies no corresponding modeling advantage.}
    \label{fig:supp-p4-warped-space}
\end{figure*}

\begin{figure*}[p]
    \centering
    \includegraphics[width=0.76\textwidth]{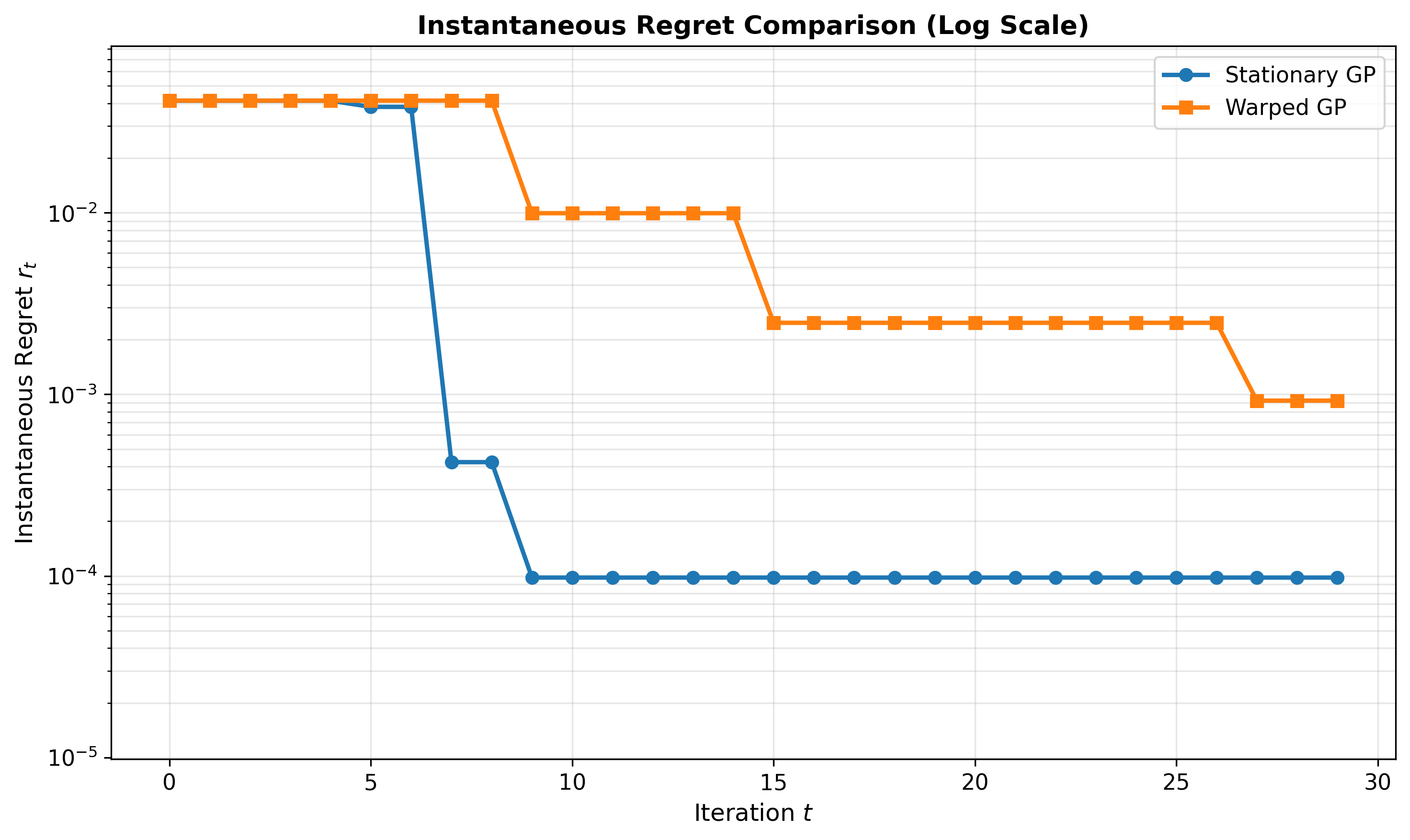}
    \caption{P4 instantaneous regret.
    The stationary model finds low-regret queries sooner and finishes below the warped model, reversing the ordering of P1--P3.
    This negative control shows that finite-library adaptation is not a free improvement.
    Lower is better and the vertical axis is logarithmic.}
    \label{fig:supp-p4-regret}
\end{figure*}

\clearpage

\section{Additional Experimental Design Details}
\label{sec:supp-experimental-design}

\subsection{Shared FLIWBO Implementation}

Raw coordinates are affinely mapped to a task-specific $[\tau,1-\tau]^D$, $\tau>0$, before applying coordinate-wise Beta-CDF transformations
$w_d(x_d)=\BetaCDF(x_d,\alpha_d,\beta_d)$.
The maps and their inverses are therefore smooth on the modeled domain, including at physical-domain endpoints.
Regularized marginal likelihood selects from a fixed one-coordinate grid of $L$ parameter pairs.
In higher dimensions, $S$ coordinate sweeps search within the predeclared $L^D$ Cartesian-product library by scoring $L$ joint warps at a time while holding the other coordinates fixed.
Every scored GP uses the full shared history, so the selector fits $SDL+1$ GPs per round while $N_\varepsilon=L^D$ in the confidence schedule.
Base Mat\'ern-kernel parameters are fixed within each experiment.
The theorem assumes exact UCB maximization, whereas experiments use probabilistic reparameterization as a stochastic acquisition optimizer.
Problem-specific constants and complete configurations are provided with the code.

\subsection{Repeated Optimizer Comparisons}

\subsubsection{Known-function problems.}
Each method receives the same initial design in a given run.
We report best-so-far simple regret to the known or numerically approximated optimum, averaged over 50 runs with 95\% confidence intervals.

The Confidence-Fence Problem separates representation from acquisition.
Its latent objective is
\begin{equation}
\begin{aligned}
 g(z)&=\exp[-((z-0.20)/0.08)^2]\\
     &\quad+0.10\sin(8\pi z)\\
     &\quad+2\exp[-((z-0.80)/0.008)^2],
\end{aligned}
 \label{eq:confidence-fenced-objective}
\end{equation}
and the observed coordinate is mapped by $z=\BetaCDF(x,6.08,2.0933)$.
Sixteen fixed initial points characterize the broad local peak and bracket, without observing, the narrow global peak.
The construction asks whether a useful warp paired with UCB exploration can cross this ``fence.''
EI remaining trapped is an outcome of this setup, not the definition of the task or a general claim about EI, and the adversarial construction should not be read as a representative distribution of black-box objectives.

The Warped Gaussian-Mixture Problem is a five-mode anisotropic mixture with a low-amplitude oscillatory background in latent coordinates and unequal Beta-CDF distortions in the two observed coordinates.
The Warped Hartmann6 Problem similarly composes the standard six-dimensional maximization objective with six distinct coordinate warps.
Exact objective and warp parameters for these problems are provided with the code.

All three problems compare fixed-kernel GP-UCB in observed coordinates, FLIWBO-UCB, Snoek-style continuously warped GP-EI with MCMC, and oracle GP-UCB/EI in latent coordinates \citep{snoek2014input}.
The Warped Gaussian-Mixture Problem and Warped Hartmann6 Problem additionally include FLIWBO-EI.
Oracle methods receive the generating warp and are optimistic references under perfect geometry recovery, not deployable competitors or finite-budget bounds.
The objectives are deterministic; run variation comes from initial designs, except for the fixed initialization of the Confidence-Fence Problem, and stochastic acquisition optimization.

\subsubsection{Fashion-MNIST HPO Problem.}
We optimize validation cross-entropy for a small two-convolutional-layer network trained for five epochs on 12,000 Fashion-MNIST images, using 3,000 validation and 5,000 test images \citep{xiao2017fashion}.
The five variables are learning rate in $[10^{-4},10^{-1}]$, weight decay in $[10^{-7},10^{-2}]$, momentum in $[0,0.99]$, dropout in $[0,0.6]$, and augmentation strength in $[0,1]$.
Test metrics are logged but never drive the search; we report the test error of the incumbent chosen by validation loss.

We compare FLIWBO-UCB/EI, Snoek-style warped GP-EI, and fixed-kernel GP-UCB/EI under both linear and log encodings of learning rate and weight decay.
Profiles are locked across the 20 runs.
Methods share run and evaluation seeds and the same initial unit-cube design under their declared encoding, so linear and log variants do not start from identical physical learning rates or weight decays.
The small network, data subset, short training schedule, and method-specific encodings limit extrapolation to large-scale HPO.

\subsection{MAS Feasibility Protocol}

\paragraph{Setting.}
In collaboration with a participating organization, we applied FLIWBO-UCB to multi-agent workflows for Python program repair, using QuixBugs \citep{lin2017quixbug} as the evaluation suite.
A human engineer from the organization provided a manually designed MAS as a practical baseline.
Each BO query executes a full candidate workflow and incurs model calls, tokens, and wall-clock time, so the study asks whether FLIWBO can operate under these costs and how its selected designs compare to that human reference---not whether warping causes the gain or whether FLIWBO dominates other optimizers.

\paragraph{Encoding and objective.}
A candidate contains at most five agents, each encoded by a model, one of 32 tool subsets, one of 48 prompts, and one of six successor choices.
This gives a 20-dimensional categorical encoding with nominal size
$(3\cdot32\cdot48\cdot6)^5\approx1.62\times10^{22}$, although this overcounts behaviorally distinct executed workflows.
Semantically ordered categories are normalized to $[0.02,0.98]^{20}$ before modeling.
The objective balances repairs and token cost:
\begin{equation}
 f(x)=R(x)-10^{-5}C_{\mathrm{tok}}(x),
 \label{eq:mas-objective}
\end{equation}
where $R(x)$ is the number of programs repaired (40 possible) and $C_{\mathrm{tok}}(x)$ is total token use.

\paragraph{Protocol and limitations.}
We perform two runs with three initial and 158 sequential evaluations each.
The second run weakens the identity-warp prior after a post-hoc diagnostic on the first run's observations, so the runs are neither independent replications nor a controlled ablation.
Each selected final design and the human baseline are re-evaluated 58 times.
The human baseline uses the same models and tools but may use freely written prompts and more than five agents, making it a practical reference rather than a matched point in the search space.
Other optimizers are omitted because reproducing the comparison across methods and seeds would require thousands of complete workflow evaluations.
The study can establish application feasibility and a practical comparison against human design, but not a causal benefit from warping, optimizer superiority, or empirical validation of the asymptotic regret guarantee.

\section{Fashion-MNIST HPO Problem: Endpoint Statistics}
\label{sec:supp-fashion-endpoints}

Table~\ref{tab:supp-fashion-endpoints} reports the final incumbent after 150 BO evaluations for the 20 paired runs of the Fashion-MNIST HPO Problem.
The endpoint is the lowest validation loss observed in each run; the validation and test errors are those of the corresponding incumbent.

\begin{table*}[t]
    \centering
    {\small
    \setlength{\tabcolsep}{3.5pt}
    \begin{tabular}{@{}lccccc@{}}
        \toprule
        Method & Final validation loss & Val. error (\%) & Test error (\%) & Paired $\Delta$ (95\% CI) & $p$ \\
        \midrule
        Fixed GP-UCB (linear)       & $0.3735 \pm 0.0120$ & 13.86 & 14.67 & $+0.0227 \pm 0.0154$ & .009 \\
        Fixed GP-EI (linear)        & $0.3737 \pm 0.0147$ & 13.66 & 14.46 & $+0.0229 \pm 0.0190$ & .029 \\
        Fixed GP-UCB (log)          & $0.3690 \pm 0.0090$ & 13.58 & 14.49 & $+0.0182 \pm 0.0147$ & .026 \\
        Fixed GP-EI (log)           & $0.3509 \pm 0.0091$ & 12.90 & 14.08 & reference              & --   \\
        Snoek-style warped GP-EI    & $0.3310 \pm 0.0083$ & 12.03 & 13.28 & $-0.0199 \pm 0.0144$ & .014 \\
        FLIWBO-UCB                   & $0.3580 \pm 0.0095$ & 13.15 & 14.06 & $+0.0072 \pm 0.0129$ & .290 \\
        FLIWBO-EI                    & $0.3502 \pm 0.0076$ & 12.95 & 13.95 & $-0.0006 \pm 0.0141$ & .930 \\
        \bottomrule
    \end{tabular}
    }
    \caption{Fashion-MNIST HPO Problem endpoint statistics ($n=20$).
    The first numerical column gives the across-run mean final validation loss and its 95\% confidence interval; validation and test errors are percentages.
    Paired differences are method minus fixed GP-EI with manual log encoding, so negative values favor the listed method; parentheses give 95\% confidence intervals and $p$-values are from two-sided paired $t$-tests.
    Linear- and log-encoded methods share run and evaluation seeds but not identical physical initial designs, so cross-encoding comparisons inherit this limitation.}
    \label{tab:supp-fashion-endpoints}
\end{table*}

\end{document}